\documentclass[10pt,twocolumn,letterpaper]{article}

\usepackage[pagenumbers]{cvpr} 
\usepackage{cvpr}              

\def\confName{3DV\xspace}
\def\confYear{2024\xspace}

\usepackage{epsfig}
\usepackage{graphicx}
\usepackage{amsmath}
\usepackage{amssymb}

\usepackage{graphicx}
\usepackage{amsmath}
\usepackage{amssymb}
\usepackage{booktabs}

\usepackage{multirow}
\usepackage{multicol}
\usepackage{amssymb}
\usepackage{booktabs,array}
\usepackage{xcolor}
\usepackage{amsmath}
\usepackage{bm}
\usepackage{subcaption}
\usepackage{tabularx}
\usepackage{tikz}
\usepackage[ruled,vlined]{algorithm2e}

\definecolor{commentcolor}{RGB}{110,154,155}  
\newcommand{\PyComment}[1]{\ttfamily\textcolor{commentcolor}{\# #1}}  
\newcommand{\PyCode}[1]{\ttfamily\textcolor{black}{#1}} 

\usepackage[pagebackref,breaklinks,colorlinks]{hyperref}

\title{Sparse 3D Reconstruction via Object-Centric Ray Sampling}

\author{Llukman Cerkezi \ Paolo Favaro\\
Computer Vision Group, Institute of Computer Science, University of Bern, Switzerland \\
{\tt\small \{llukman.cerkezi, paolo.favaro\}@unibe.ch}
}

\begin{document}

\captionsetup[subfigure]{skip=0.5ex, belowskip=0.5ex, labelformat=simple}

\renewcommand\thesubfigure{}
\renewcommand{\tabcolsep}{0pt}

\twocolumn[{%
\maketitle

\begin{tabular}{ccccccc}
\includegraphics[width=0.28\textwidth]{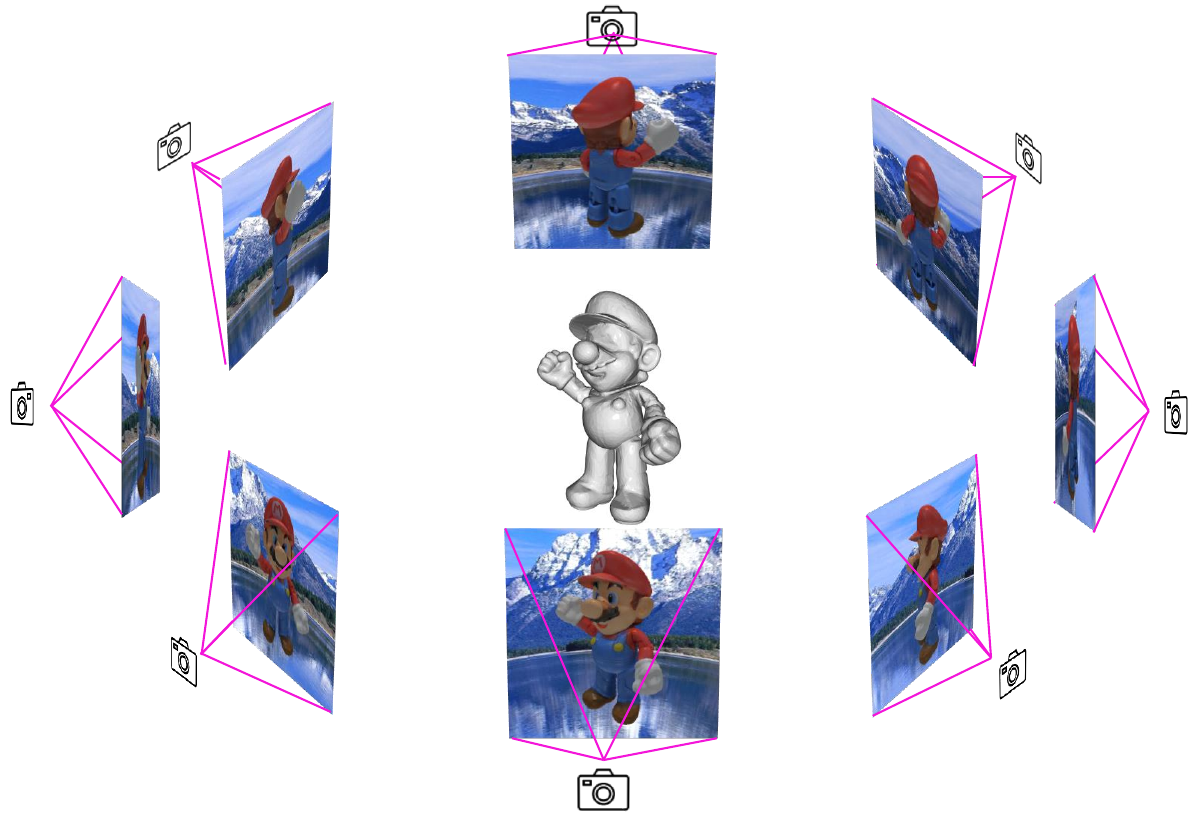} &{\includegraphics[width=0.12\textwidth, trim=7.5cm 3.5cm 9.5cm 4.5cm,clip]{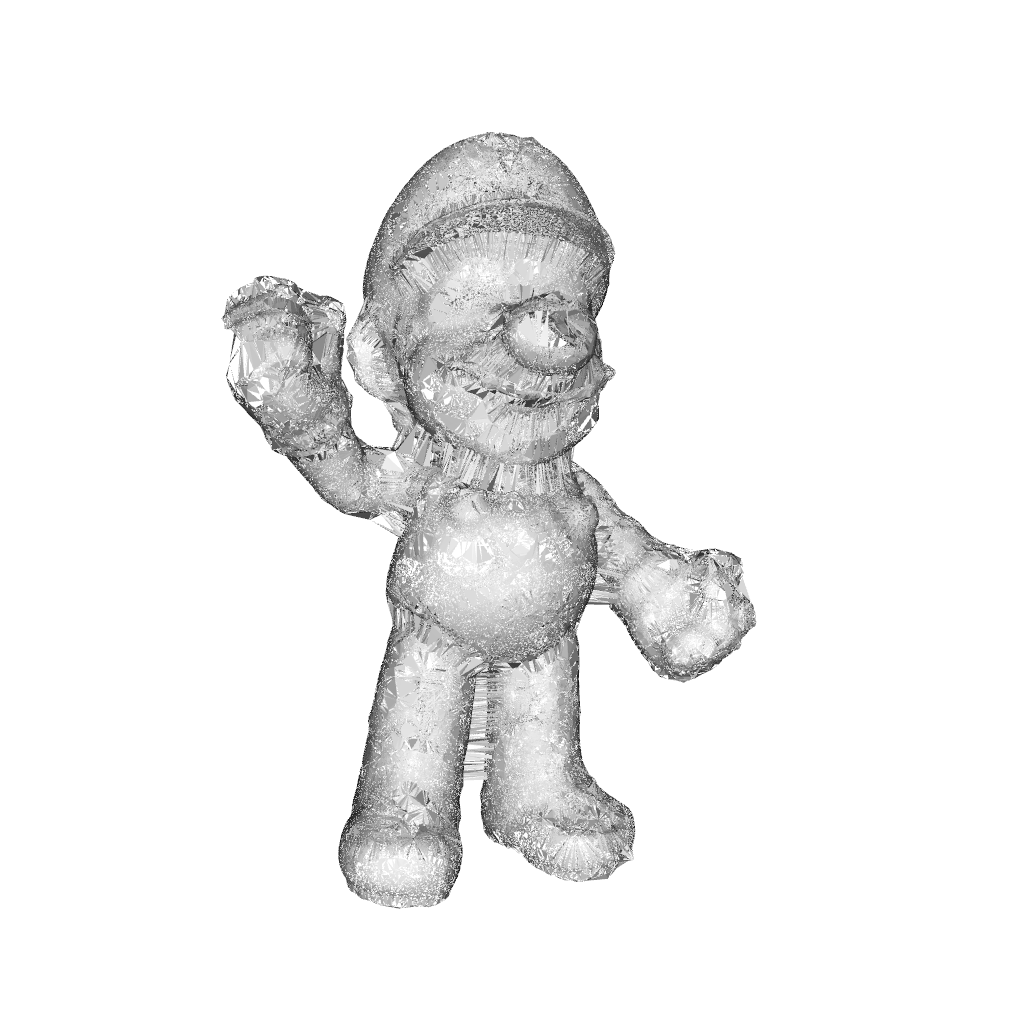}} & {\includegraphics[width=0.12\textwidth, trim=7.5cm 3.5cm 9.5cm 4.5cm,clip]{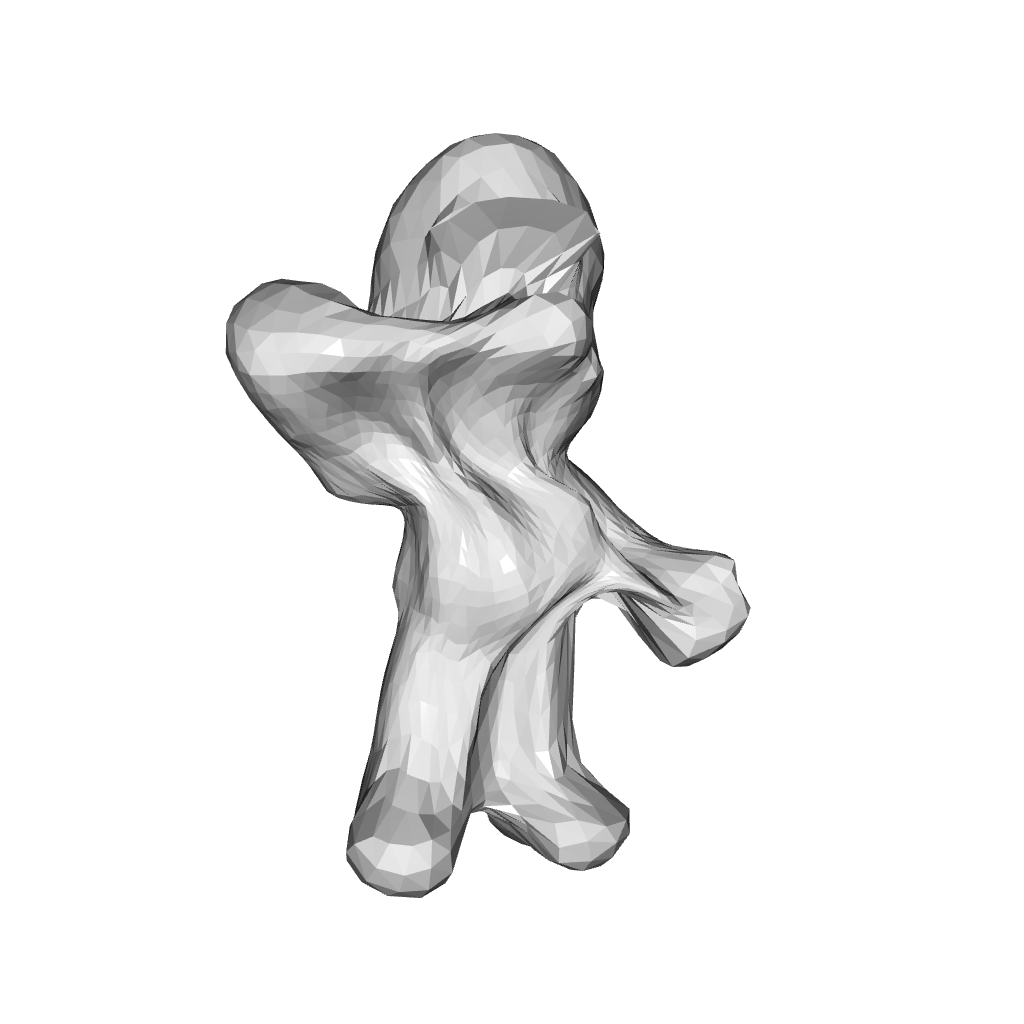}} & {\includegraphics[width=0.12\textwidth, trim=7.5cm 3.5cm 9.5cm 4.5cm,clip]{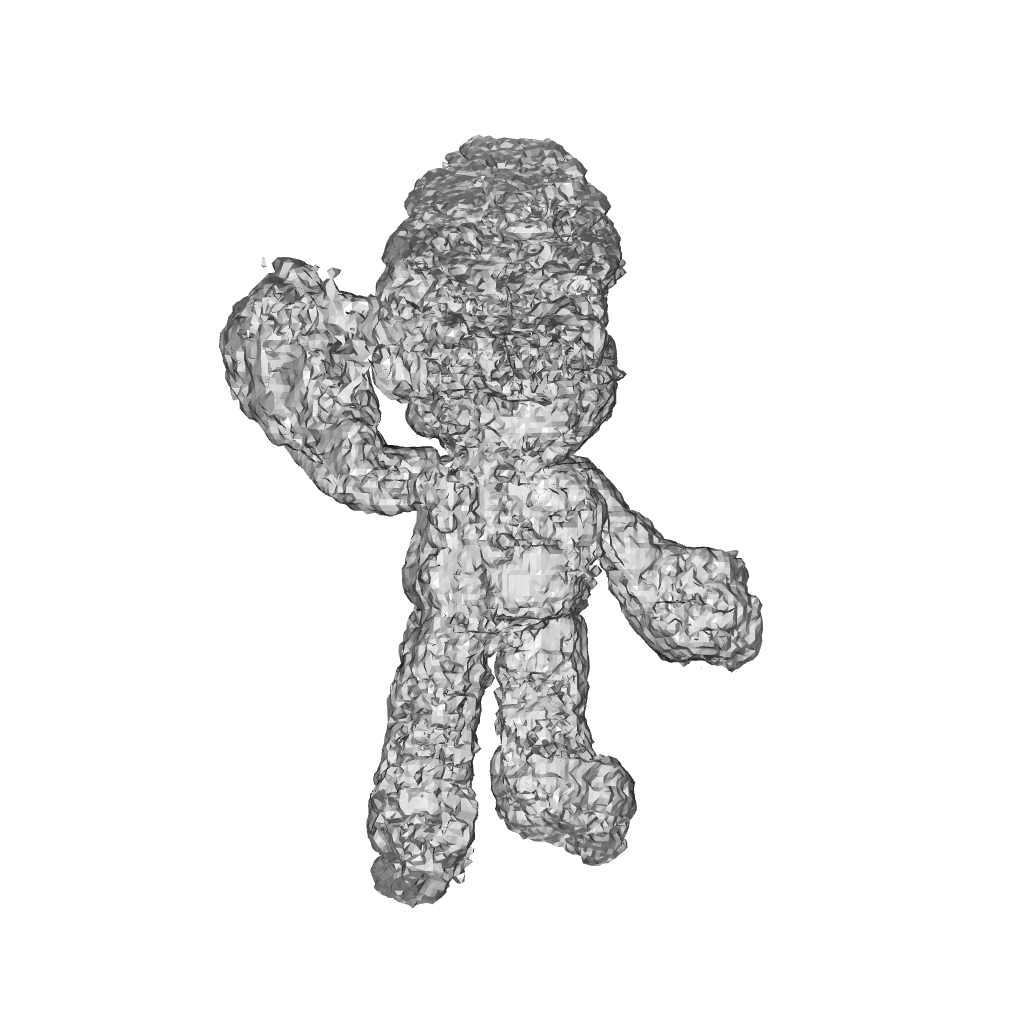}} & {\includegraphics[width=0.12\textwidth, trim=7.5cm 3.5cm 9.5cm 4.5cm,clip]{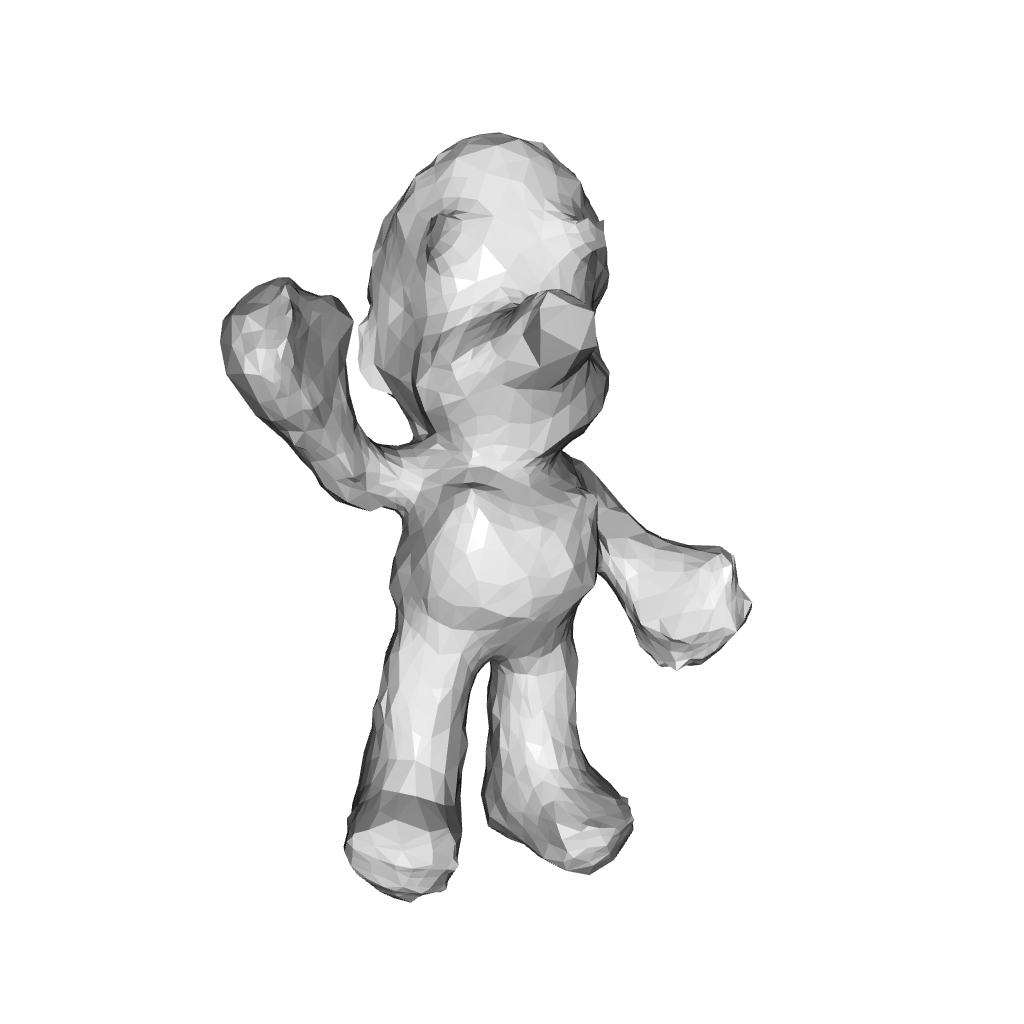}} & {\includegraphics[width=0.12\textwidth, trim=7.5cm 3.5cm 9.5cm 4.5cm,clip]{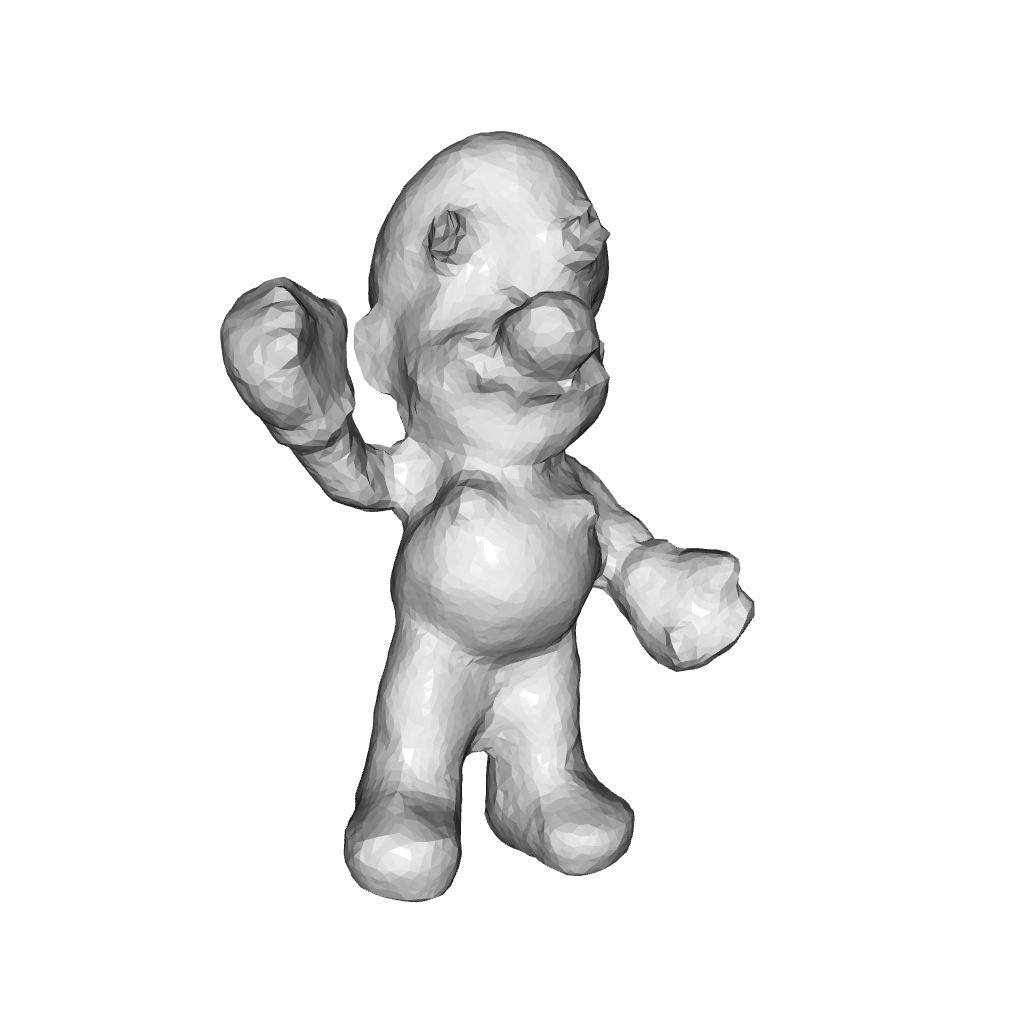}} & {\includegraphics[width=0.12\textwidth,trim=7.5cm 3.5cm 9.5cm 4.5cm,clip]{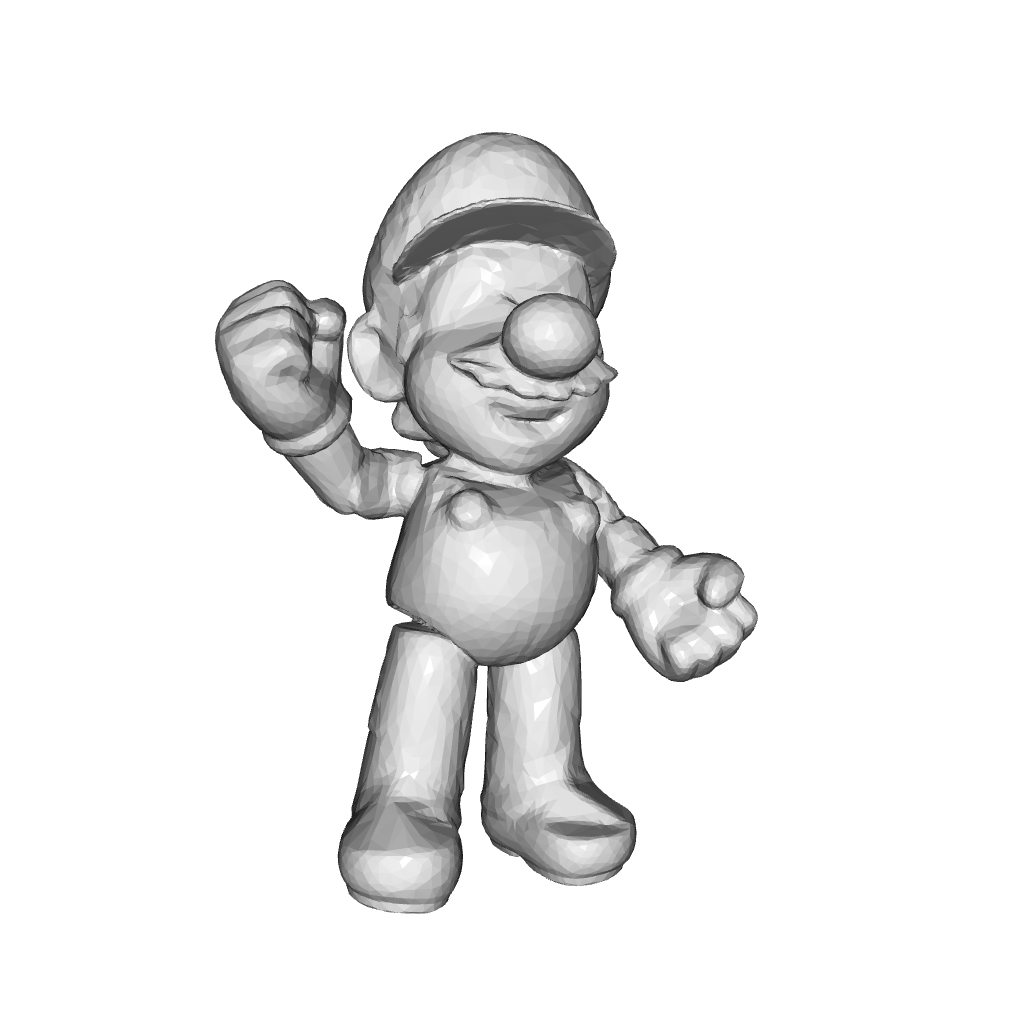}} \\
& COLMAP*-50 & NeRS & RegNeRF & DS & Ours & GT  \\
\end{tabular}

\captionof{figure}{\label{fig:teaser} Left: Sparse view setting of a $360^{\circ}$ camera rig with 8 views. 
Right: 3D reconstructions with existing SotA methods. Due to the sparsity and wide spacing of the camera views, methods such as  
NeRS~\cite{NERS_2021_Neurips} 
and 
RegNeRF~\cite{ regnerf_2021} 
reconstruct surfaces with visible artifacts. COLMAP$^*$  \cite{schoenberger2016mvs, Schonberger_2016_CVPR} returned a valid mesh only with 50 views (so is used only as a reference). Methods such as DS~\cite{diff_stereopsis_2021_arxiv}  
obtain better reconstructions, but with fewer details than with our approach. Most methods make use of masks to segment the object in each view. In contrast, our method can work without this additional supervision and still obtain accurate 3D reconstructions (compare to the GT). 
\vspace{1em}}
}]  
\begin{abstract}
    \vspace{-0.5em}
    We propose a novel method for 3D object reconstruction from a sparse set of views captured from a 360-degree calibrated camera rig. 
    We represent the object surface through a hybrid model that uses both an MLP-based neural representation and a triangle mesh. 
    A key contribution in our work is a novel object-centric sampling scheme of the neural representation, where rays are shared among all views. This efficiently concentrates and reduces the number of samples used to update the neural model at each iteration. This sampling scheme relies on the mesh representation to ensure also that samples are well-distributed along its normals.  
    The rendering is then performed efficiently by a differentiable renderer. We demonstrate that this sampling scheme results in a more effective training of the neural representation, does not require the additional supervision of segmentation masks, yields state of the art 3D reconstructions, and works with sparse views on the Google’s Scanned Objects, Tank and Temples and MVMC Car datasets. Code available at: \url{https://github.com/llukmancerkezi/ROSTER}
\end{abstract}
\vspace{-1em}    
\section{Introduction}
\label{sec:intro}

The task of reconstructing the 3D surface of an object from multiple calibrated views is a well-established problem with a long history of methods exploring a wide range of 3D representations and optimization methods \cite{Furukawa_et_al,furukawa2015multi,forsyth2002computer,szeliski2022computer}. Recent approaches have focused their attention on deep learning models \cite{DVR, sitzmann2019srns, NERF_2020_ECCV, khot2019learning, NERS_2021_Neurips, chen2019learning, liu2020neural, Mildenhall_LLFF}. In particular, methods based on neural rendering such as NeRF and its variants \cite{NERF_2020_ECCV, activeNERF,barron2022mipnerf360,chen2021mvsnerf,yariv2020multiview}, have not only shown impressive view interpolation capabilities, but also the ability to output 3D reconstructions as a byproduct of their training.

\begin{figure}[t]
    \captionsetup{font={normalsize}}
    \centering
    \includegraphics[scale=0.45,trim=0 0 0cm 0,clip]{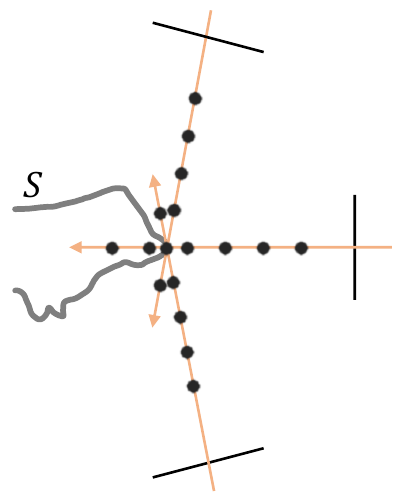}\hspace{0.5cm} 
    \includegraphics[scale=0.45,trim=0cm 0 0 0,clip]{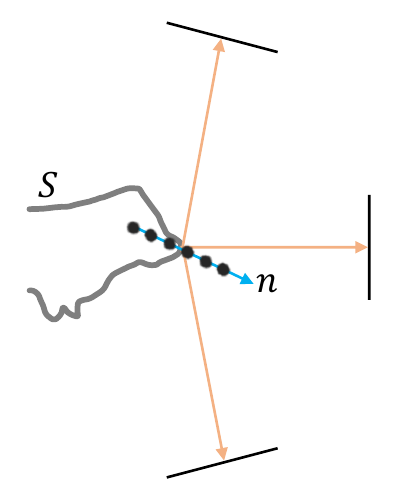} 
    \caption{\textbf{Sampling schemes.} Left: NeRF view-centric sampling scheme. Right: Our object-centric sampling scheme. The view-centric sampling scheme uses separate sets of 3D samples for each camera view. This leads to overfitting when views are sparse. Object-centric sampling instead shares the same 3D samples across multiple views.}
    \label{fig:sampling}
\end{figure}

NeRF's neural rendering drastically simplifies the generation of images given a new camera pose. It altogether avoids the complex modeling of the light interaction with surfaces in the scene. 
A neural renderer learns to output the color of a pixel as a weighted average of 3D point samples from the NeRF model. Current methods choose these samples along the ray defined by the given pixel and the camera center (see Figure~\ref{fig:sampling}~left). Because each camera view defines a separate pencil of rays, the 3D samples rarely overlap. Thus, each view will provide updates for mostly independent sets of parameters of the NeRF model, which can lead to data overfitting. In practice, overfitting means that views used for training will be rendered correctly, but new camera views will give unrealistic images. 
Such overfitting is particularly prominent when training a NeRF on only a sparse set of views with a broad object coverage (\eg, see the $360^{\circ}$ camera rig in Figure~\ref{fig:teaser}).

\begin{figure}
    \captionsetup{font={normalsize}}
    \centering
    \begin{tabular}{cccc}
        \includegraphics[width=.12\textwidth, trim=0cm 0cm 0cm 0cm, clip]{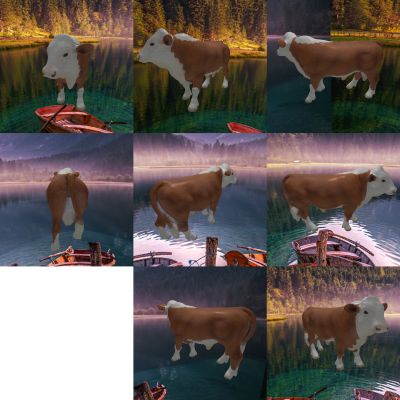} &
        \includegraphics[width=.12\textwidth, trim=1.0cm 1.2cm 0.7cm 2cm, clip]{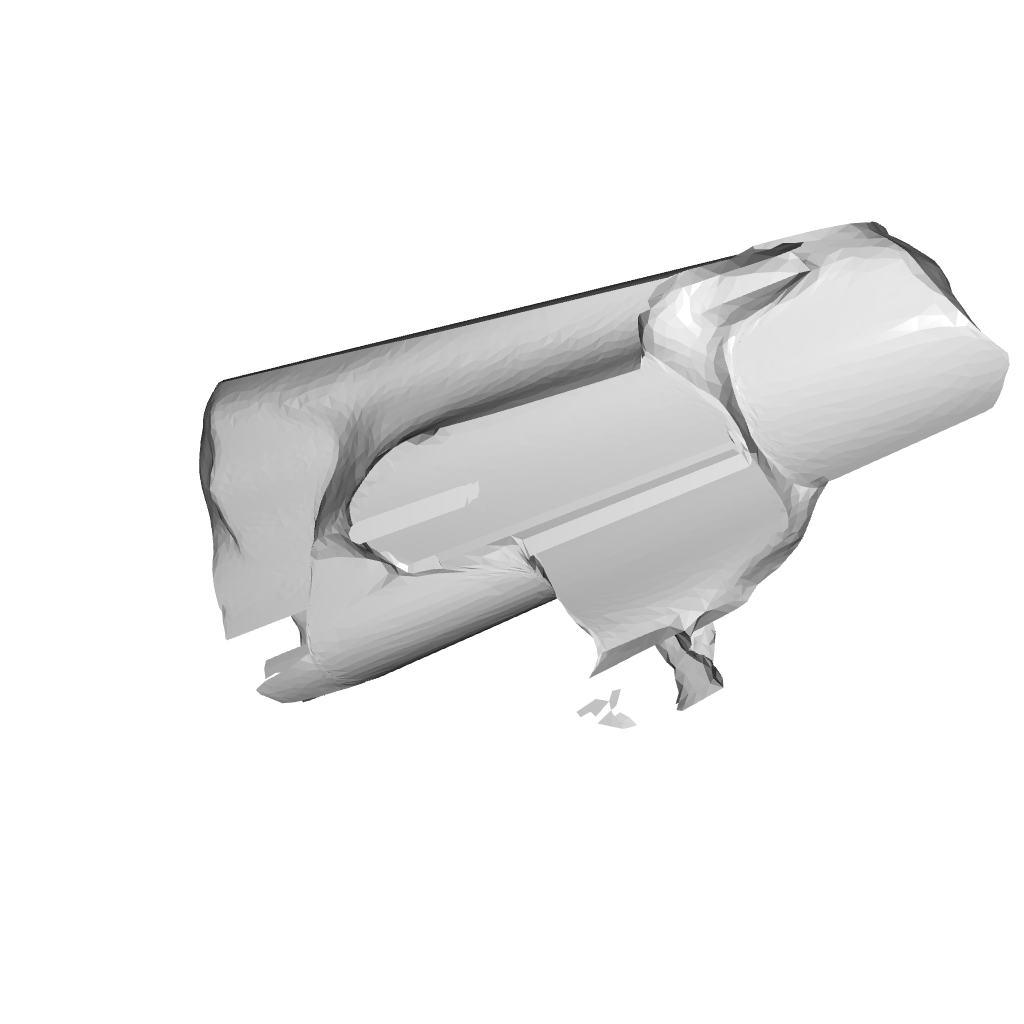} &
        \includegraphics[width=.12\textwidth, trim=1.0cm 1.2cm 0.7cm 2cm, clip]{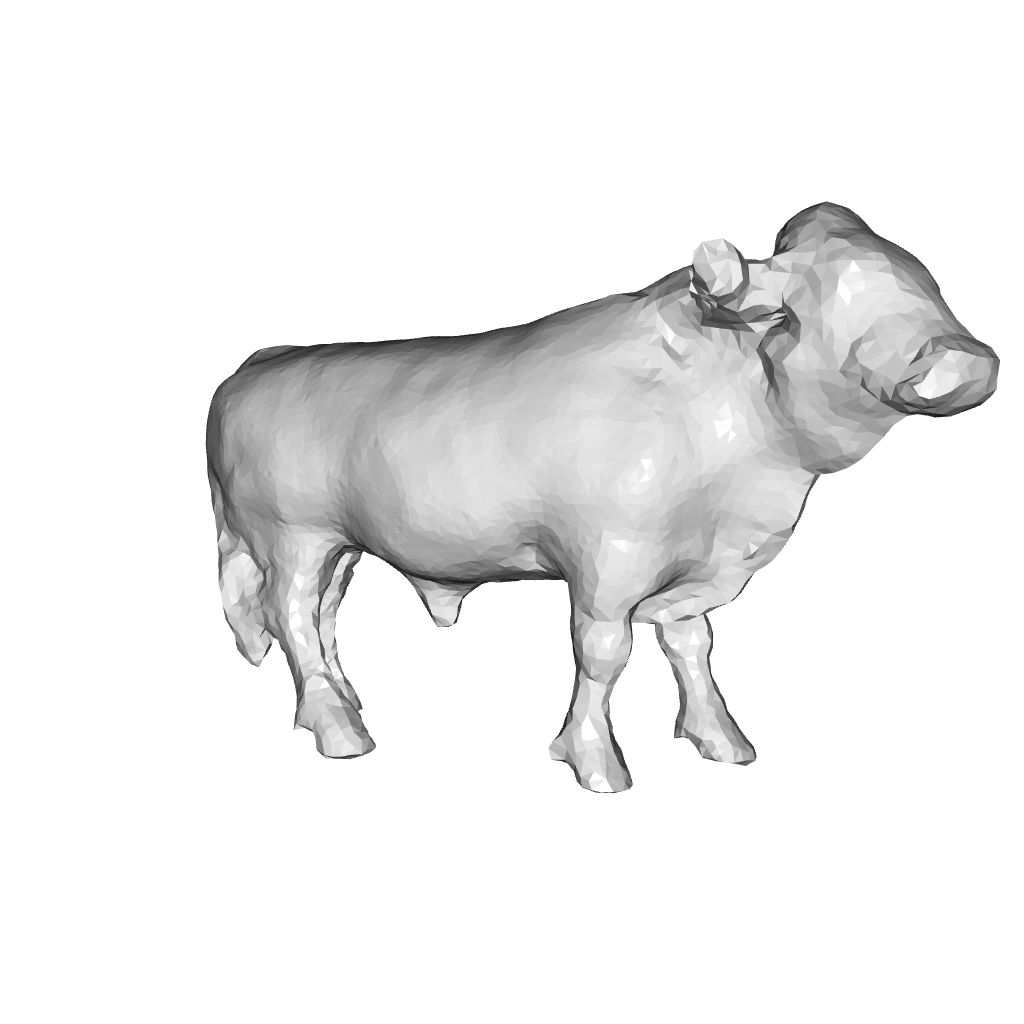} &
        \includegraphics[width=.12\textwidth, trim=1.0cm 1.2cm 0.7cm 2cm, clip]{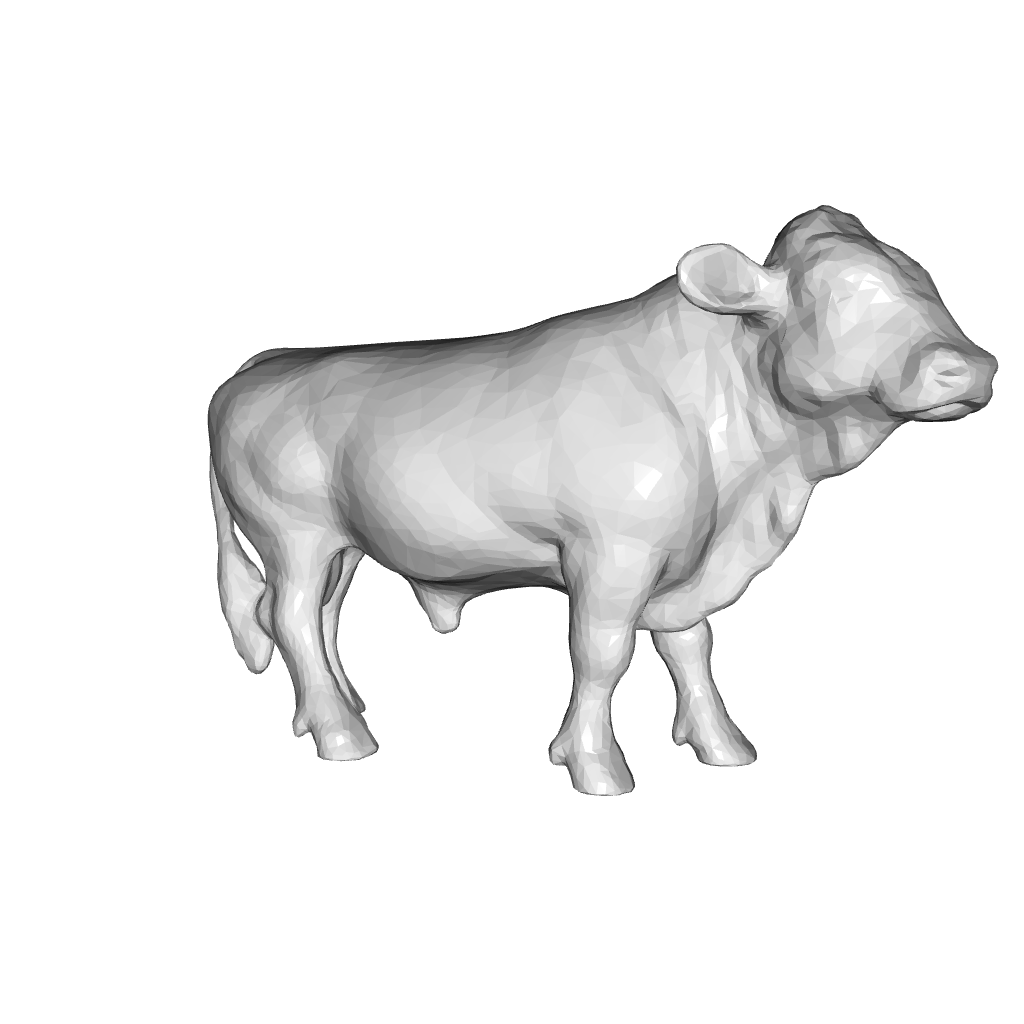} \\
        \includegraphics[width=.12\textwidth, trim=0cm 0cm 0cm 0cm, clip]{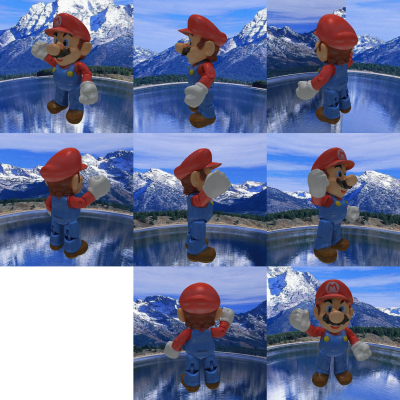} &
        \includegraphics[width=.12\textwidth, trim=1.0cm 1.2cm 2cm 2cm, clip]{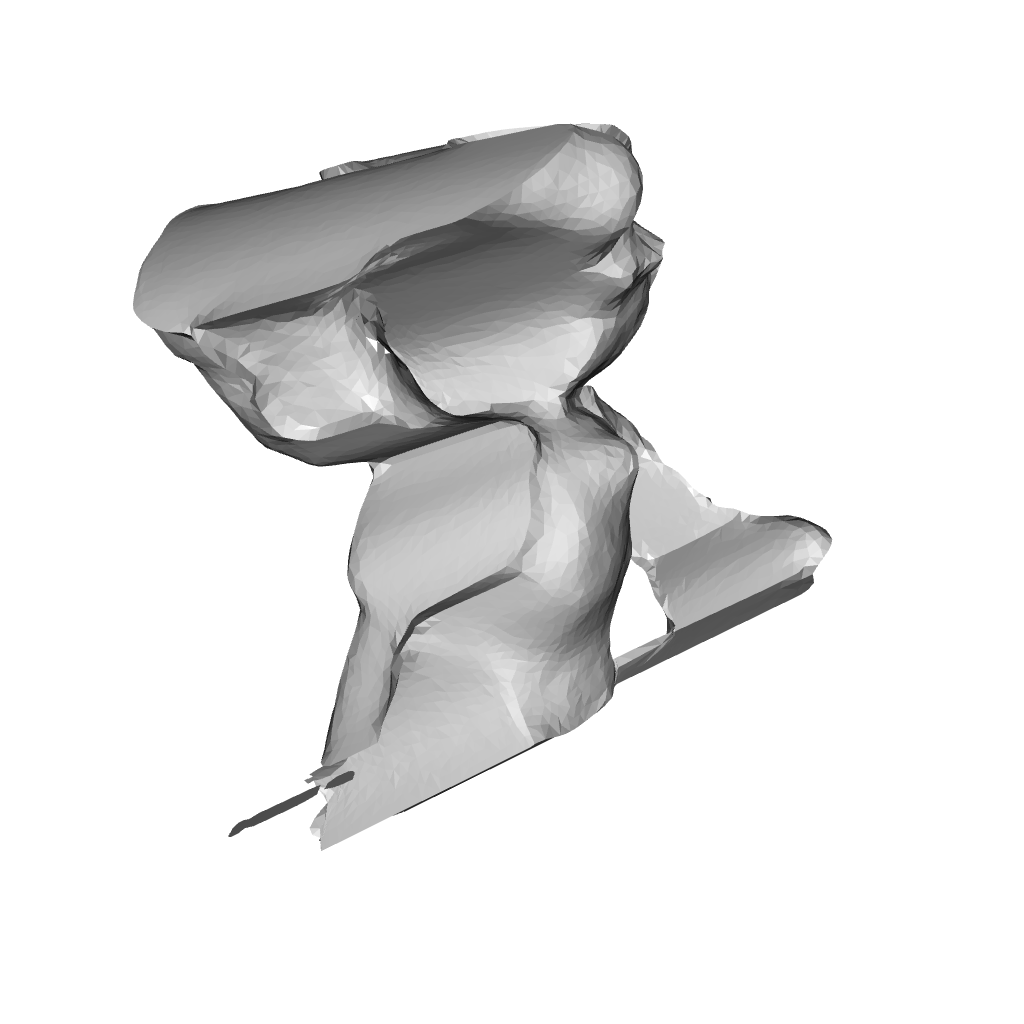} &
        \includegraphics[width=.12\textwidth, trim=1.0cm 1.2cm 2cm 2cm, clip]{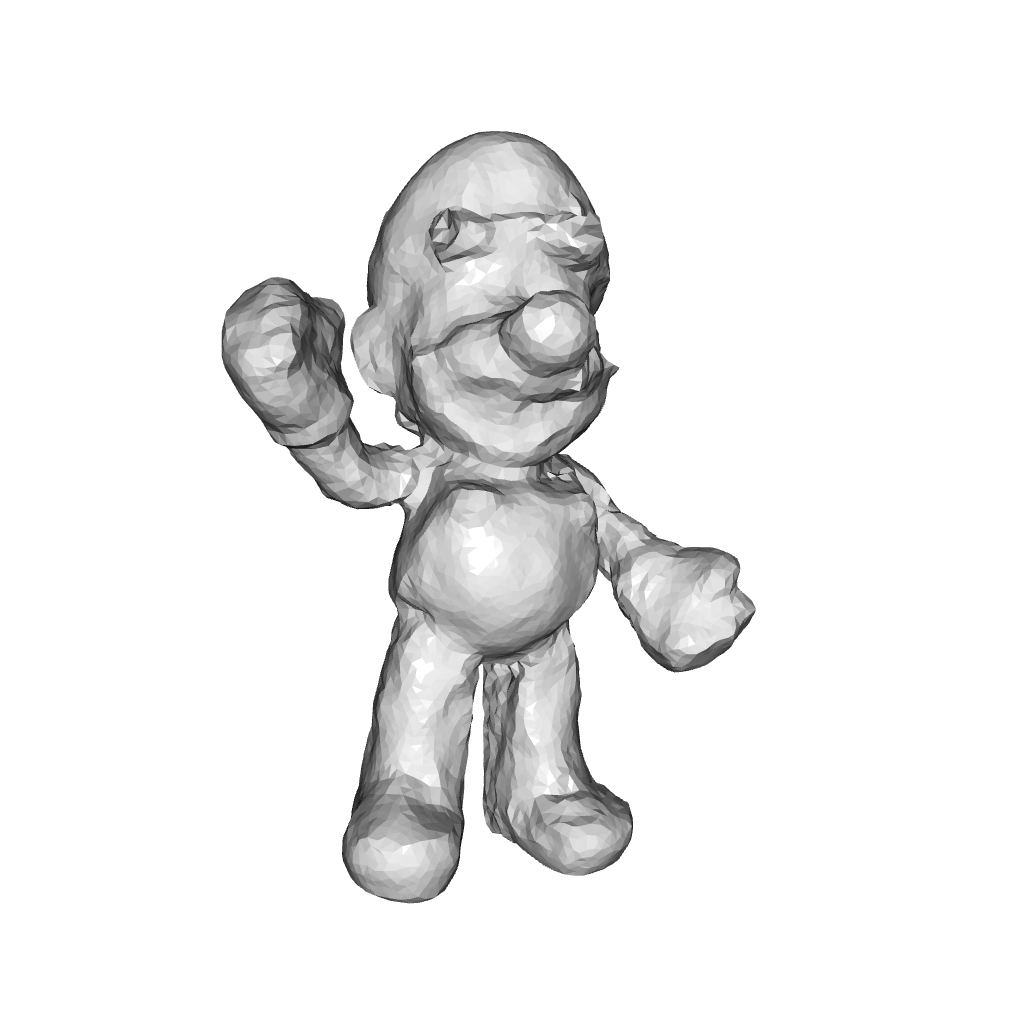} &
        \includegraphics[width=.12\textwidth, trim=1.0cm 1.2cm 2cm 2cm, clip]{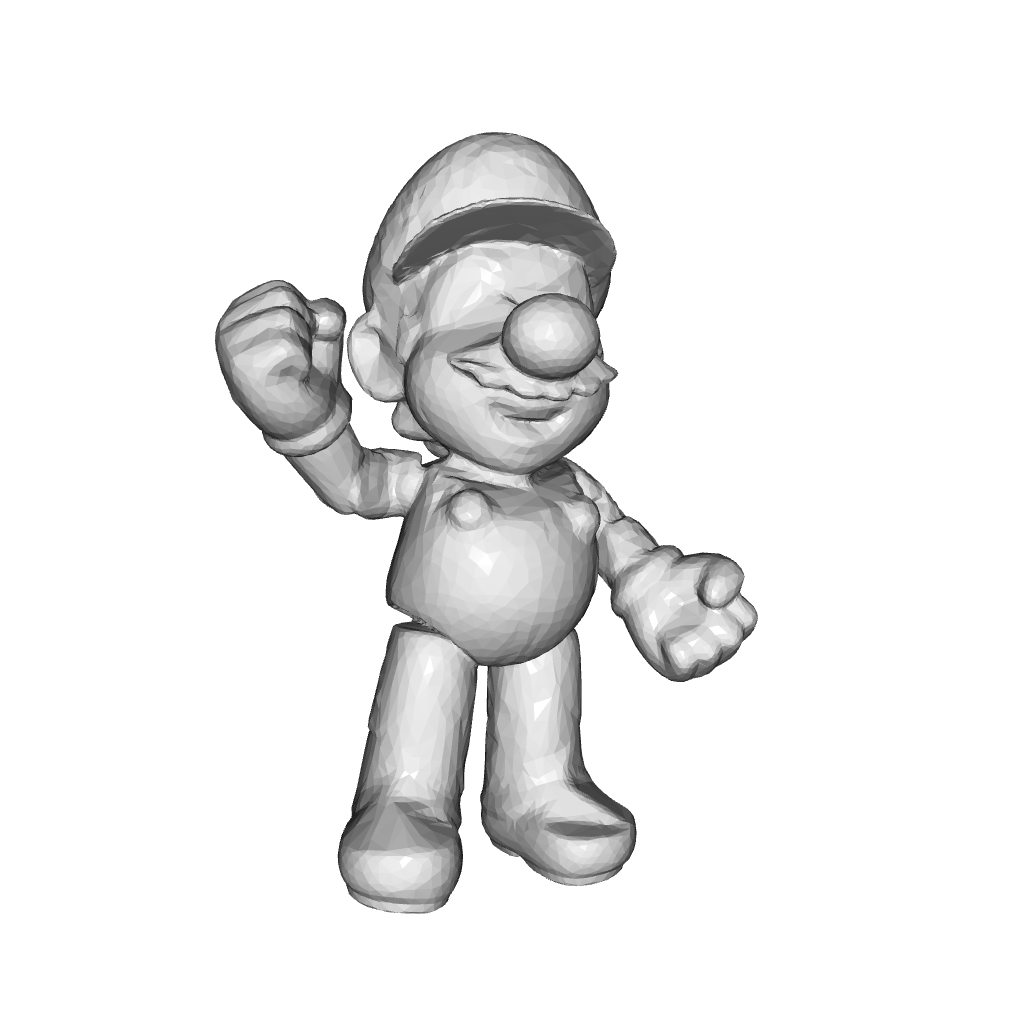} \\
    \end{tabular}
    \caption{\textbf{View vs object-centric sampling (see Figure~\ref{fig:sampling}).} Computational efficiency: The view-centric approach uses $8\times K$ samples per mesh vertex, with $K$ camera views. In contrast, the object-centric approach uses only $8$ samples per vertex regardless of the number of camera views. Object-centric sampling is not only more efficient but also avoids overfitting. For more details, please check Section \ref{sec:method}.}
    \label{fig:object_vs_view_centric_recon}
\end{figure}

In this work, we address overfitting when working with sparse views by proposing an object-centric sampling scheme that is shared across all views (see Figure~\ref{fig:sampling}~right and Figure~\ref{fig:object_vs_view_centric_recon}). 
We design the scheme so that all (visible) views can provide an update for the same 3D points on a given sampling ray. 
To do so, we introduce a hybrid 3D representation, where we simultaneously update a Multi Layer Perceptron (MLP) based implicit surface model (similarly to a NeRF) and an associated triangle mesh. The MLP model defines an implicit 3D representation of the scene, while the mesh is used to define the sampling rays.
These rays are located at each mesh vertex and take \textit{the direction of the normal to the mesh}.
Then, the mesh vertex of the current object surface is updated by querying the MLP model at 3D samples on the corresponding ray.
We use a similar deep learning model to associate color to the mesh.
We then image the triangle mesh in each camera view via a differentiable renderer. 
Because of our representation, the queried 3D samples can be shared across multiple views and thus avoid the overfitting shown by NeRF models in our settings.

Notice that a common practice to handle overfitting in NeRF models trained on sparse views is to constrain the 3D reconstruction through object masks. Masks provide a very strong 3D cue. In fact, a (coarse) reconstruction of an object can even be obtained from the masks alone, a technique known as shape from silhouette \cite{laurentini1994visual}. We show that our method yields accurate 3D reconstructions even without mask constraints. This confirms experimentally the effectiveness of our sampling scheme in avoiding overfitting.

To summarize, our contributions are 
\begin{itemize}
\item A novel object-centric sampling scheme that efficiently shares samples across multiple views and avoids overfitting;
the robustness of our method is such that it does not need additional constraints, such as 2D object masks;
\item A 3D reconstruction method that yields state of the art results with a sparse set of views from a 360-degree camera rig (on the Google’s Scanned Objects \cite{google_dataset}, Tank and Temples \cite{tank_dataset} and MVMC Car datasets \cite{NERS_2021_Neurips}).
\end{itemize}

\section{Prior Work}
\label{sec:literature}

\noindent\textbf{Mesh-based methods.}
With the development of differentiable renderers  \cite{kato2017neural,szabo2019unsupervised,chen2019learning,ravi2020accelerating}, object reconstruction is now possible through gradient descent (or backpropagation in the context of deep learning).
A common approach to predict the shape of an object using differentiable rendering is to use category level image collections \cite{kanazawa_2018_arxiv,goel2020shape,ye2021shelfsupervised,Simoni_2021,tulsiani_implicit_2020_arxiv, monnier2022unicorn}.
Recently, some methods aim to estimate the shape of an object in a classic multi-view stereo setting
and without any prior knowledge of the object category \cite{NERS_2021_Neurips,diff_stereopsis_2021_arxiv,worchel2022,Xu_22,munkberg2021nvdiffrec}.
Several methods also propose different ways to update the surface of the reconstructed object.
The methods proposed by Goel et al.~\cite{diff_stereopsis_2021_arxiv} and Worchel et al.~\cite{worchel2022} update the mesh surface by predicting vertex offsets to the template mesh.
Zhang et al.~\cite{NERS_2021_Neurips} use a neural displacement field over a canonical sphere, but restrict the geometry to model only genus-zero topologies.
Xu et al.~\cite{Xu_22}, after getting a smooth initial shape via \cite{yariv2021volume}, proposes surface-based local MLPs to encode the vertex displacement field for the reconstruction of surface details.
Munkberg et al.~\cite{munkberg2021nvdiffrec} use a hybrid representation as we do in our method. They learn the signed distance field (SDF) of the reconstructed object. The SDF is defined on samples on a fixed tetrahedral grid and then converted to a surface mesh via deep marching tetrahedra \cite{shen2021dmtet}. In contrast, we adapt the samples to the surface of the object as we reconstruct it. 

\noindent\textbf{Implicit representations for volume rendering.}
Recently, Neural Radiance Field (NeRF)-based methods have shown great performance in novel view synthesis tasks \cite{NERF_2020_ECCV, nerf_plusplus, verbin2021refnerf, barron2021mipnerf, lin2021barf, sitzmann2022light, martinbrualla2020nerfw, Plenoxels_2022_cvpr}.
However, these methods require a dense number of training views and camera poses to render realistic views.
Methods that tackle novel view rendering from a small set of training views usually exploit two directions. 
The methods in the first group pre-train their models on large scale calibrated multiview datasets of diverse scenes \cite{SRF_2021, chen2021mvsnerf, wang2021ibrnet, yu2020pixelnerf, Sharf_2021, liu2022neuray, mine2021}.  
In our approach, however, we consider training only on a small set of images.

The methods in the second group add an additional regularization term to their optimization cost to handle the limited number of available views.
Diet-Nerf~\cite{Jain_2021_ICCV} incorporates an additional loss that encourages the similarity between the pre-trained CLIP features between the training images and rendered novel views.
RegNerf~\cite{regnerf_2021} incorporates two additional loss terms: 1) color regularization to avoid color shifts between different views and 2) geometry regularization to enforce smoothness between predicted depths of neighboring pixels. 
InfoNerf~\cite{infonerf} adds a ray entropy loss to minimize the entropy of the volume density along each ray (thus, they encourage the concentration of the volume density on a single location along the ray). This is not suitable for sparse 360-degree camera rigs, where the camera positions lie at the same elevation angle (as in our case) as a ray can be shared by two opposite camera centers. They also add a loss that minimizes the KL-divergence between the normalized density profiles of two neighboring rays. 
DS-Nerf~\cite{kangle2021dsnerf} instead improves the reconstruction quality by adding depth supervision. As they report, its performance is only as good as the
estimates of depth obtained by COLMAP \cite{schoenberger2016mvs, Schonberger_2016_CVPR}.
Common to all of the above methods is that they require some sort of additional training (except InforNerf~\cite{infonerf}), while our method reconstructs objects without any additional pre-training.

\noindent\textbf{Implicit representations for surface rendering.} \cite{neural_rendering_survey} provides an overview of methods that use implicit representations for either volume or surface rendering.
This family of approaches uses a neural SDF or an occupancy field as an implicit surface representation.
DVR \cite{DVR} and IDR \cite{yariv2020multiview} are pioneering SDF methods that use only images for training. They both provide a differentiable rendering formulation using implicit gradients. 
However, both methods require accurate object masks as well as appropriate weight initialization due to the difficulty of propagating gradients.
IRON~\cite{iron_2022} proposes a method to estimate edge derivatives to ease the optimization of neural SDFs.
Some of the methods combine implicit surface models with volumetric ones \cite{wang2021neus,yariv2021volume,Oechsle2021ICCV} and also implicit surface models with explicit ones \cite{ishit_levelset, remelli2020meshsdf, neumesh}.  
One advantage of the methods that combine implicit surface models with volumetric ones is that they do not require mask supervision and are more stable. 
However, they heavily depend on a large number of training images.
SparseNeuS~\cite{sparse_neus} can work in the sparse view setting, but requires pre-training on a multi-view dataset of multiple scenes.
Additionally, it is pretrained only for the narrow view setup, as opposed to the 360-degree one.

\section{Sparse 3D Reconstruction}
\label{sec:method}

\subsection{Problem Formulation}

Our goal is to reconstruct the 3D surface of the object depicted in $N$ images $I = \{ I_1, I_2, \cdots, I_N\}$ given their corresponding calibrated camera views $\Pi = \{ \pi_1, \pi_2, \cdots, \pi_N\}$, where $\pi_i$ denotes the 3D camera pose and intrinsic camera calibration parameters. 
We consider the \emph{sparse setting}, \ie, when $N$ is small (\eg, $8-15$ views). We mostly use camera views distributed uniformly in a $360^{\circ}$ rig (see Figure~\ref{fig:teaser}), but our method can also work for the narrow view setup (see the supplementary material for experiments with this setting).
We pose the 3D reconstruction task as the problem of finding the 3D surface and texture such that the images $I^r = \{I_1^r,\dots,I_N^r\}$ rendered with the given camera views $\Pi$, best match the corresponding set of captured images $I$.

\subsection{3D Representation}
\label{sec:representation}
We describe the surface of the 3D object via a hybrid model that maintains both an implicit (density-based) and explicit (mesh-based) representation. 
The two representations serve different purposes. The explicit one is used to efficiently render views of the object and is directly obtained from the implicit representation. 
The advantage of the implicit representation is that it can smoothly transition through a general family of 3D shapes (\eg, from a sphere to a torus). This is especially useful when the 3D reconstruction is achieved through iterative gradient-based optimization algorithms. Such transition is typically much more difficult to achieve with a lone explicit representation.
More specifically, the implicit representation is based on the Implicit Surface Neural Network (ISNN) $F_\text{shape}: \mathbb{R}^3 \rightarrow \mathbb{R}$, that outputs the object density value $\sigma \doteq F_\text{shape}(X)$ at a 3D point $X\in \mathbb{R}^3$. 
The explicit representation is obtained by converting the implicit representation in $F_\text{shape}$ to a triangle mesh $\mathcal{G} = (\mathcal{V}, \mathcal{F})$ consisting of $M$ vertex locations $\mathcal{V}$ and a face list $\mathcal{F}$. A triangle in $\mathcal{F}$ is the triplet of indices of the vertices in $\mathcal{V}$ that form that triangle. 
The conversion of $F_\text{shape}$ to the explicit representation mesh is based on the selection of a finite set of 3D points, which we call samples and discuss in detail in the next section.

\subsection{Object-Centric Sampling} 
\begin{figure}[t]
\captionsetup{font={normalsize}}
\begin{center}
 \includegraphics[width = \linewidth]{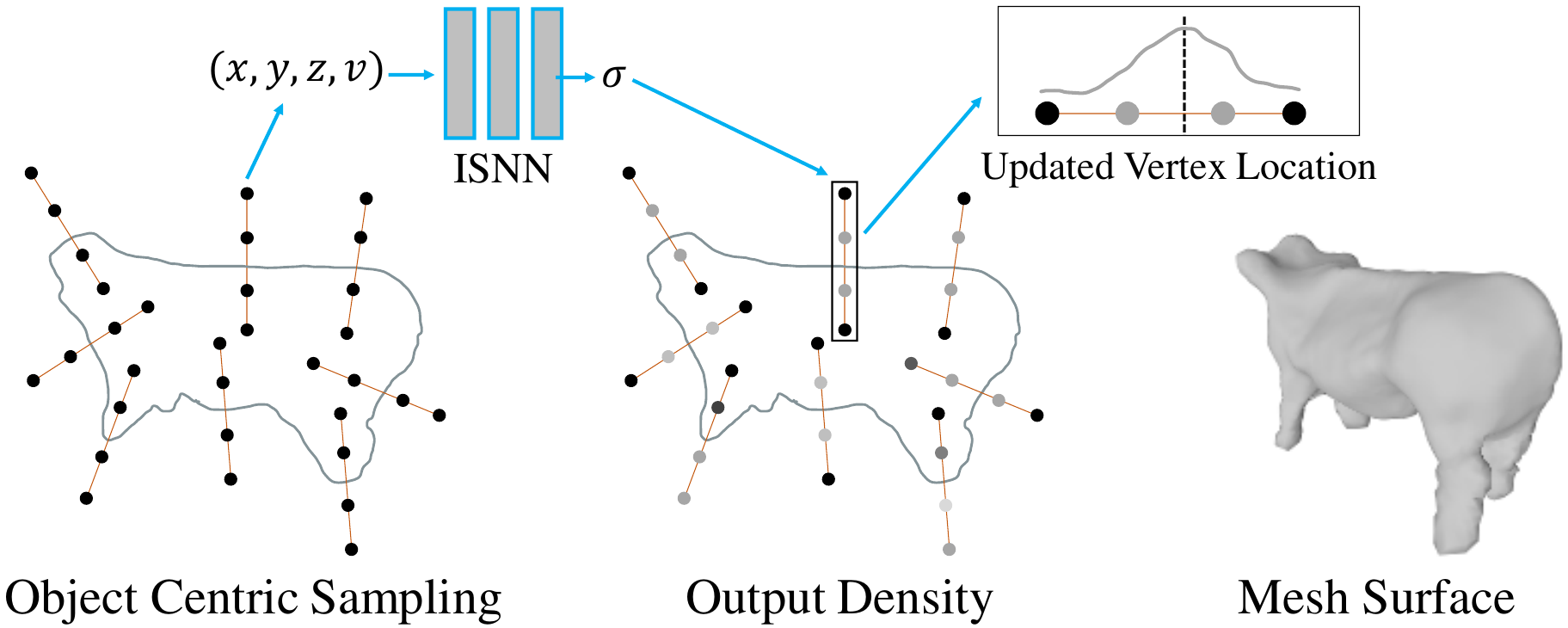}  \caption{Detailed model representation. We feed the object-centric points to ISNN and obtain a density value. Then, we update the vertex location via eq.~\eqref{eq:calc_vert_loc} using the points sampled along the vertex normal. We repeat this operation for all vertices to get the updated mesh surface.}
 \label{Fig:Surface_Learning}
\end{center}
\vspace{-0.28cm}
\end{figure}

In Figure~\ref{Fig:Surface_Learning}, we show a 2D slice of the implicit representation of the \textsc{cow} 3D shape. The implicit representation will have a density $\sigma$ that is close to $1$ at the surface of the object and $0$ elsewhere. In an iterative procedure, we can assume that we already have some existing mesh that is sufficiently close to the current surface of the implicit representation (recall that the implicit representation will be updated through the optimization procedure). To also update the mesh, we use its existing mesh vertices and normals to define segments that are approximately normal to the updated implicit surface and to select samples on the segments in equal number on either side of the surface.

More formally, for each vertex $V_i\in \mathcal{V}$, $i=1,\dots,M$, in the current \emph{out-of-date} mesh, we define a \emph{sampling ray} $R_i$, such that $R_i\propto N_i$, where $N_i$ is the surface normal at the vertex $V_i$. Along the ray $R_i$ we draw $K$ 3D samples $X_{i,1},\dots,X_{i,K}$ (in Figure~\ref{Fig:Surface_Learning} we show $K=4$). We define outward and inward point samples by drawing $K$ equally spaced 3D points from the segments $[V_i, V_i + t_i^\text{out}N_i]$ and $[V_i - t_i^\text{in}N_i, V_i]$, where $t_i^\text{in}, t_i^\text{out}>0$.  
The range factors $t_i^\text{in}$ and  $t_i^\text{out}$ are defined independently so that samples on either one of the two segments stay always either inside or outside the mesh, with a maximum possible range. This choice allows to deal with the reconstruction of thin structures of the mesh (\eg, the leg of a horse). For each 3D point we obtain corresponding densities $\sigma_{i,1}, \cdots, \sigma_{i,K}$ from the ISNN via $\sigma_{i,j} = F_\text{shape}(X_{i,j})$. We then compute normalized weights via the softmax function as $w_{i,j} \propto \exp(\sigma_{i,j})$, such that $\sum_{j=1}^K w_{i,j} = 1$. 
Finally, we define the updated mesh vertex $V_i$ as the following weighted sum
\begin{align}
    \hat V_i = \sum_{k=1}^{K} w_{i,k} X_{i,k}.
    \label{eq:calc_vert_loc}
\end{align}
See Figures~\ref{Fig:Template_Learning} and \ref{Fig:Surface_Learning} for an illustration of these steps.

\noindent\textbf{Remark. }
In view-centric methods, such as NeRF, the sampling rays are defined via the camera directions. 
The view-centric approach presents two drawbacks: Firstly, the number of points grows linearly with the number of cameras. Secondly, when using view-centric (VC) sampling, the surface can only evolve within the subspace determined by the camera poses, resulting in elongated shapes (as observed in Figure~\ref{fig:object_vs_view_centric_recon}). This limitation becomes particularly challenging in scenarios with sparse camera views.

\noindent\textbf{Adding Texture.} 
Instead of obtaining color directly from the ISNN as in NeRF models \cite{NERF_2020_ECCV, nerf_plusplus, barron2021mipnerf, sitzmann2022light, martinbrualla2020nerfw}, we introduce a separate model, the Texture Neural Network (TNN) $F_\text{texture}:\mathbb{R}^P\mapsto \mathbb{R}^3$, where $P$ is the size of the 3D position embedding.
Given the updated 3D vertex location $\hat V_i$, we compute its positional embedding $\gamma(\hat V_i)$, where $\gamma(\cdot)$ denotes the positional encoding operator, and then obtain the color $C_i \doteq F_\text{texture} (\gamma(\hat V_i))$. 

\begin{figure}[t]
\captionsetup{font={normalsize}}
\begin{center}
 \includegraphics[width=\linewidth]{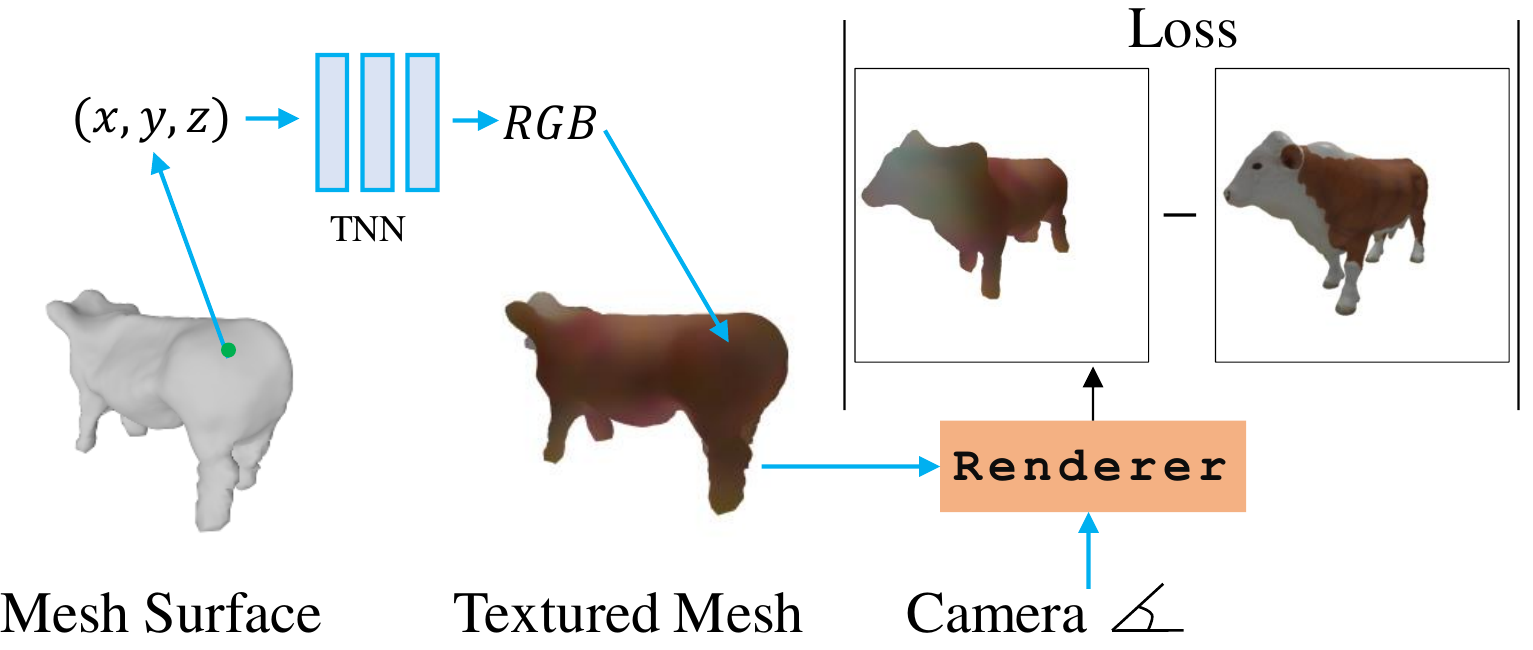}  \caption{
 We assign a color to each vertex of the mesh by querying the TNN model at that vertex. Then, we feed the textured mesh and a camera viewpoint as input to a differentiable renderer to synthesize a view of the scene. The reconstruction task is based on minimizing the difference between the synthesized view and a captured image (with the same  viewpoint) in both $L_1$ and perceptual norms.}
 \label{fig:Texture_Learning}
\end{center}
\vspace{-0.1cm}
\end{figure}

\noindent\textbf{Image Rendering.} 
The above procedure yields an updated triangle mesh $\hat{ \mathcal{G}} = (\hat{\mathcal{V}}, \mathcal{F})$, where $\hat{\mathcal{V}}=\{\hat V_1,\dots,\hat V_M\}$, with corresponding vertex colors $C=\{C_1,\dots, C_M\}$.
We render the image viewed by the $j$-th camera with calibration $\pi_j$, $j\in \{1,\dots,N\}$, by feeding $\hat{\mathcal{G}}$ and the vertex colors $C$ to a differentiable renderer \cite{szabo2019unsupervised,kato2017neural,chen2019learning,ravi2020accelerating}. 
This yields the rendered image $I_j^{r} = \mathtt{Renderer}(\hat{\mathcal{G}}, C, \pi_j )$ (see Figure~\ref{fig:Texture_Learning}). 

\noindent\textbf{Reconstruction Loss.} 
$F_\text{shape}$ and  $F_\text{texture}$ are parametrized as Multi Layer Perceptron (MLP) networks (more details of their architectures are in section~\ref{sec:implementation}). We train their parameters by minimizing the following loss on the images $I$
\begin{equation}
    L = L_\text{images} (I,I^r) + L_\text{perceptual} (I,I^r) + \lambda L_\text{laplacian} (\hat{\mathcal{G}})
    \label{eq:loss}
\end{equation}
where $L_\text{images}(I,I^r) = \sum_{i=1}^{N} |I_{i} - I_{i}^{r}|_1$ is the $L_1$ loss between the rendered images and captured images. 
$L_\text{perceptual}$ is the same loss as $L_\text{images}$, but where instead of the $L_1$ loss we use the perceptual loss \cite{zhang2018unreasonable} and $L_\text{Laplacian}$ is the Laplacian loss of the mesh $\hat{\mathcal{G}}$ \cite{mesh_laplacian_reg}, which we use to regularize the reconstructed 3D vertices $\hat{\mathcal{V}}$ through the parameter $\lambda>0$.
We optimize the loss $L$ using the AdamW optimizer \cite{adamW_optimizer}.

\subsection{Technical Details}
\label{sec:sampling}

We employ several ideas to make the optimization robust and accurate. 

\begin{figure}[t]
\begin{center}
 \includegraphics[width = .95\linewidth]{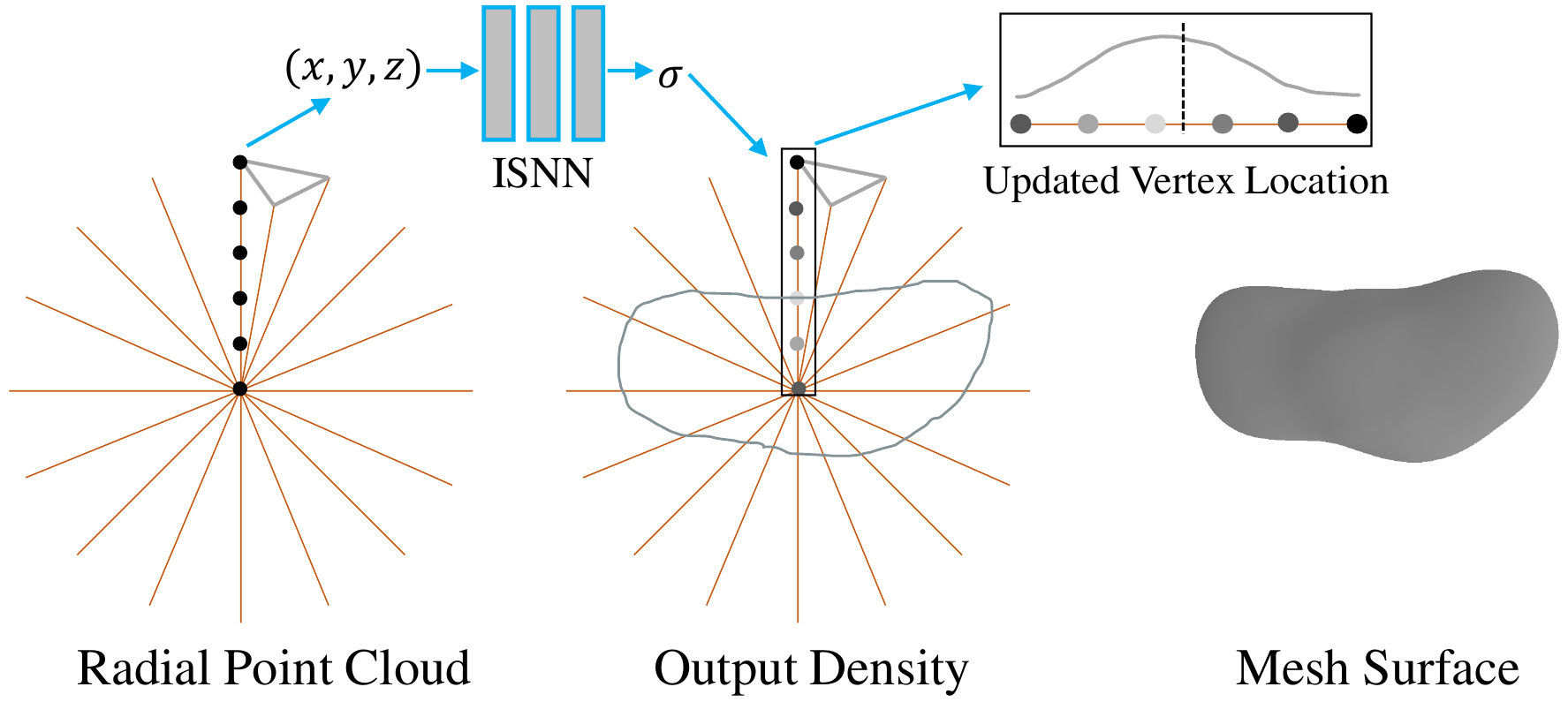}  \caption{Mesh initialization. The points in the radial point cloud are fed to ISNN to obtain a density value. Then, for each ray we update the vertex location using eq.~\eqref{eq:calc_vert_loc}. By repeating this operation for all rays, we obtain the mesh surface.}
 \label{Fig:Template_Learning}
\end{center}
\end{figure}

\noindent\textbf{Mesh Initialization.}
We use a robust initialization procedure to obtain a first approximate surface mesh. 
We start from a predefined sphere mesh with radius $\rho$. 
This sphere defines vertices and triangles of the mesh $\mathcal{G}$. Then, for each vertex $V_i \in \mathcal{V}$ we cast a ray $R_i$ in the radial direction 
from the origin of the sphere towards the vertex $V_i$ and draw $K$ equally spaced points $X_{i,1}, \cdots, X_{i,K}$ along the ray $R_i$, such that $|X_{i,K}|_2 = \rho$ (see  Figure~\ref{Fig:Template_Learning}). These 3D points are never changed throughout this initial model training. The mesh vertices are then computed as in eq.~\eqref{eq:calc_vert_loc}. 
Because each linear combination considers only samples $X_{i,1}, \cdots, X_{i,K}$ along a ray, the initial representation $F_\text{shape}^\text{init}$ can only move the updated vertices radially. 
Although the reconstructed mesh can model only genus zero objects and only 
describe a radial structure, it gives us a very reliable initial mesh. 

\noindent\textbf{Surface Normals.}
Surface normals are computed by averaging the normals of the faces within the second order neighbourhood around $V_i$.

\noindent\textbf{Re-meshing.}
Every $100$ iterations during the first $2500$ iterations, and then every $250$ iterations afterwards, we apply a re-meshing step that regularizes and removes self-intersections from the mesh $\mathcal{G}$. This is a separate step that is not part of the optimization of the loss eq.~\eqref{eq:loss} (\ie, there is no backpropagation through these operations). Through re-meshing, the mesh can have genus different from zero and its triangles are adjusted so that they have similar sizes. 
We use the implementation available in PyMesh - Geometry Processing Library~\cite{pymesh}.
For more details, see the supplementary material. 
Notice that the total time for the above calculations during training is almost negligible as we do not apply these operations at every iteration and they are highly optimized.

\noindent\textbf{Inside/Outside 3D Points.} For the identification of which 3D samples are inside/outside the 3D surface we use the generalized \emph{winding number} technique, which is also available in the PyMesh - Geometry Processing Library~\cite{pymesh}.

\noindent\textbf{Texture Refinement.} During training, we observe that the TNN model does not learn to predict sharp textures. 
Therefore, we run a final phase during which the mesh is kept constant and we fine-tune the TNN separately.
Following IDR~\cite{yariv2020multiview}, we feed the vertex location, the vertex normal and the camera viewing direction to the TNN so that it can describe a more general family of reflectances.
These quantities are concatenated to $\gamma(\hat V_i)$ and then fed as input to $F_\text{texture}$.  
As described earlier on, in this final phase, we optimize eq.~\eqref{eq:loss} only with respect to the parameters of $F_\text{texture}$.

\noindent\textbf{Handling the Background.}
So far, we have not discussed the presence of a background in the scene and have focused instead entirely on the surface of the object. Technically,  unless a mask for each view is provided, there is no explicit distinction between the object and the background. Masks give a strong 3D cue about the reconstructed surface, so much so that they can do most of the heavy-lifting in the 3D reconstruction. Thus, to further demonstrate the strength of our sampling and optimization scheme, we introduce a way to avoid the use of user pre-defined segmentation masks. We extend our model with an approximate background mesh representation. For simplicity, we initialize the background on a fixed mesh (a cuboid) that is sufficiently separated from the volume of camera frustrums' intersections. When we reconstruct the background, we only optimize the texture assigned to each vertex of the background. Note that we add a separate TNN to estimate the texture of the background and the texture is view-independent at all stages.
See also the supplementary material for further details.

\renewcommand{\tabcolsep}{3pt}
\begin{table}[t]
\begin{center}
\caption{Reported 3D metrics on the GSO Dataset. *Note that we obtained COLMAP-reconstructed point clouds using 50 views.  
\textbf{(**All scores have been multiplied by $10^{4}$).} Notice the robustness of our method even when not using the mask constraints.
\label{table:GSO_3D_Metric}
}
\small
\begin{tabular}{l c c c c c}
\toprule
 Method &  Mask  & CH-L2* $\downarrow$  &CH-L1 $\downarrow$  & Normal $\uparrow$ & F@10 $\uparrow$  \\  \hline
 NeRS   &yes & 18.58 & 0.052 & 0.54 & 98.10  \\
 RegNeRF  & yes&  60.19 & 0.107 &0.30 & 91.44  \\
 Munkberg  
 &yes & 13.32 & 0.047 & 0.56 & 98.65  \\  
 DS & yes & 13.21 & 0.042 & 0.71 & 98.58  \\
 NeuS  & no&  1217.00 & 0.495 & 0.37 & 32.40  \\
 NeuS  & yes&  13.85 & 0.049 & 0.70 & 98.79  \\
 COLMAP* & yes & 34.35 & 0.049 & - & 99.11  \\
 Our wo\slash BCG & yes &\textbf{ 8.69} & 0.034 & \textbf{0.75} & \textbf{99.24}   \\
 Our w\slash BCG &no & 11.08 &  0.038& 0.75 & 98.85 \\
 \hline
\end{tabular}
\end{center}
\vspace{-0.5cm}
\end{table}
\renewcommand{\tabcolsep}{3.pt}
\begin{table}[t]
\begin{center}
\caption{GSO Dataset: Quantitative evaluation of generated views on the test set. Notice the robustness of our method even when not using the mask constraints. \label{table:GSO_Texture_Metric}} 
\begin{tabular}{ l c c c c c  } 
 \toprule
 Method &  Mask & PSNR $\uparrow$  & MSE $\downarrow$ & SSIM $\uparrow$ & LPIPS  $\downarrow$ \\ \hline  
  NeRS  & yes & 20.108  & 0.0185 &  0.874 & 0.126 \\
  RegNeRF &yes & 20.217 & 0.013 & 0.882 & 0.143   \\
 Munkberg  et al. & yes & 26.838 & 0.002 & 0.955 & 0.067  \\
 DS & yes & 24.649 & 0.004 & 0.944 & 0.081  \\
 Our wo\slash BCG & yes & \textbf{29.029} &  \textbf{0.001}  & \textbf{0.967} & \textbf{0.028} \\
 Our w\slash BCG  & no &27.370& 0.002 & 0.964 &  0.038\\
\hline
\end{tabular}
\end{center}
\vspace{-0.3cm}
\end{table}
\section{Experiments}
\label{sec:experiments}

In this section, we present implementation details and results obtained on the standard datasets.
For more comprehensive ablation studies as well as for more visual results, we refer to the supplementary material.

\subsection{Implementation Details}
\label{sec:implementation}
We parameterize $F_\text{shape}^\text{init}$, $F_\text{shape}$ and $F_\text{texture}$ as MLPs with 5, 5, and 3 layers respectively and a hidden dimension of 256 for all.
The initialization mesh uses 2500 vertices, while the mesh for the object reconstruction uses a maximum of 10K vertices. 
For the background mesh we use a vertex resolution of 10K.
When training the texture network TNN in the final step, we upsample the mesh resolution to 250K vertices. 
The scale $t_{in}$ and $t_{out}$ for the detailed model reconstruction are both set to $0.15$. 
The number of samples along the rays for $F_\text{shape}^\text{init}$ and $F_\text{shape}$ is $16$ and $8$ respectively.
The learning rate for the training of the initialization, shape, and texture models is $10^{-5}$,  $5\times10^{-5}$, and $10^{-3}$.
The Laplacian regularization may change across datasets. 
For objects with non-Lambertian surfaces, \eg, specular surfaces, we use a higher Laplacian regularization, as in this case the $F_\text{shape}$ network can overfit and generate spiky surfaces due to the lack of multiview consistency across the views. 
We will release the code of all components of our work to facilitate further research and allow others to reproduce our results.

\renewcommand{\tabcolsep}{2pt}
\begin{table}[t]
\captionsetup{font={normalsize}}
\begin{center}
\caption{MVMC Car Dataset: Quantitative evaluation of generated views on the test set. Our wo\slash BCG* and Our w\slash BCG are trained with the same pre-processing as in NeRS. \label{table:MVMC_Texture_Metric}} 
\begin{tabular}{  l  cc c c c  } 
 \toprule
 Method &  Mask & PSNR  $\uparrow$  & MSE $\downarrow$ & SSIM $\uparrow$ & LPIPS $\downarrow$ \\  \hline
 NeRS  & yes& 18.381  & 0.015  & 0.852 & 0.080 \\
 RegNeRF  &yes & 15.776 & 0.028 &  0.751 &  0.259  \\
 Munkberg  et al.  & yes  & 15.145  & 0.031 & 0.761 &  0.239\\
 DS & yes & 18.608 & 0.015 & 0.835 & 0.139  \\

 Our w\slash BCG* & no & 18.301 & 0.015 &   0.855 & 0.132 \\
 Our wo\slash BCG*  & yes & 20.030 & 0.010 & 0.867  & 0.095 \\
 Our w\slash BCG& no & 18.450  &  0.014 & 0.865 & 0.142 \\
 Our wo\slash BCG& yes & \textbf{21.563}  & \textbf{ 0.007}  & \textbf{0.883}  & \textbf{0.091}  \\
  \hline
\end{tabular}
\end{center}
\vspace{-0.5cm}
\end{table}

\subsection{Datasets}
\label{sec:datasets}

\noindent\textbf{Google’s Scanned Objects (GSO) \cite{google_dataset}.} We test our algorithm on 14 different objects. 
For training, we use 8 views, and for validation 100 views.
Camera poses are uniformly spread out around the object where the elevation angle is uniformly sampled in $[0^{\circ}, 15^{\circ}]$.
The background image is generated by warping a panorama image onto a sphere.\\ 
\noindent\textbf{MVMC Car dataset \cite{NERS_2021_Neurips}.} 
We run our algorithm on 5 different cars
from the MVMC dataset.
We use the optimized camera poses provided in the dataset. 
Although they are optimized, we find that some of them are not correct. Thus, we eliminated those views for both training and testing.
We follow a leave-one-out cross-validation setup, where we leave one view for validation and the rest is used for training. 
We repeat this 5 times for each car. 
Note that this dataset is more challenging than the GSO Dataset \cite{google_dataset} as the camera locations are not spread out uniformly around the object. Most of the views are placed mainly on two opposite sides of the cars.
Furthermore, the surface of cars is not-Lambertian, and there are many light sources present in the scene too.  

\noindent \textbf{Tank and Temple dataset \cite{tank_dataset}.}
We evaluate our method on images from 2 objects, \textit{Truck} and \textit{Ignatius}. We use 15 images for training and the rest as the test set.  
We obtain the image masks of each object by rendering its corresponding laser-scanned ground-truth 3D point cloud. 
The camera poses are computed via COLMAP’s SfM pipeline \cite{Schonberger_2016_CVPR}. 
\renewcommand{\tabcolsep}{0.5pt}
\begin{figure*}
    \captionsetup{font={normalsize}}
	\begin{center}
	\begin{tabular}{cccccccc}

        \includegraphics[width = .12\textwidth, trim=0.cm 0cm 0cm 0cm,clip]{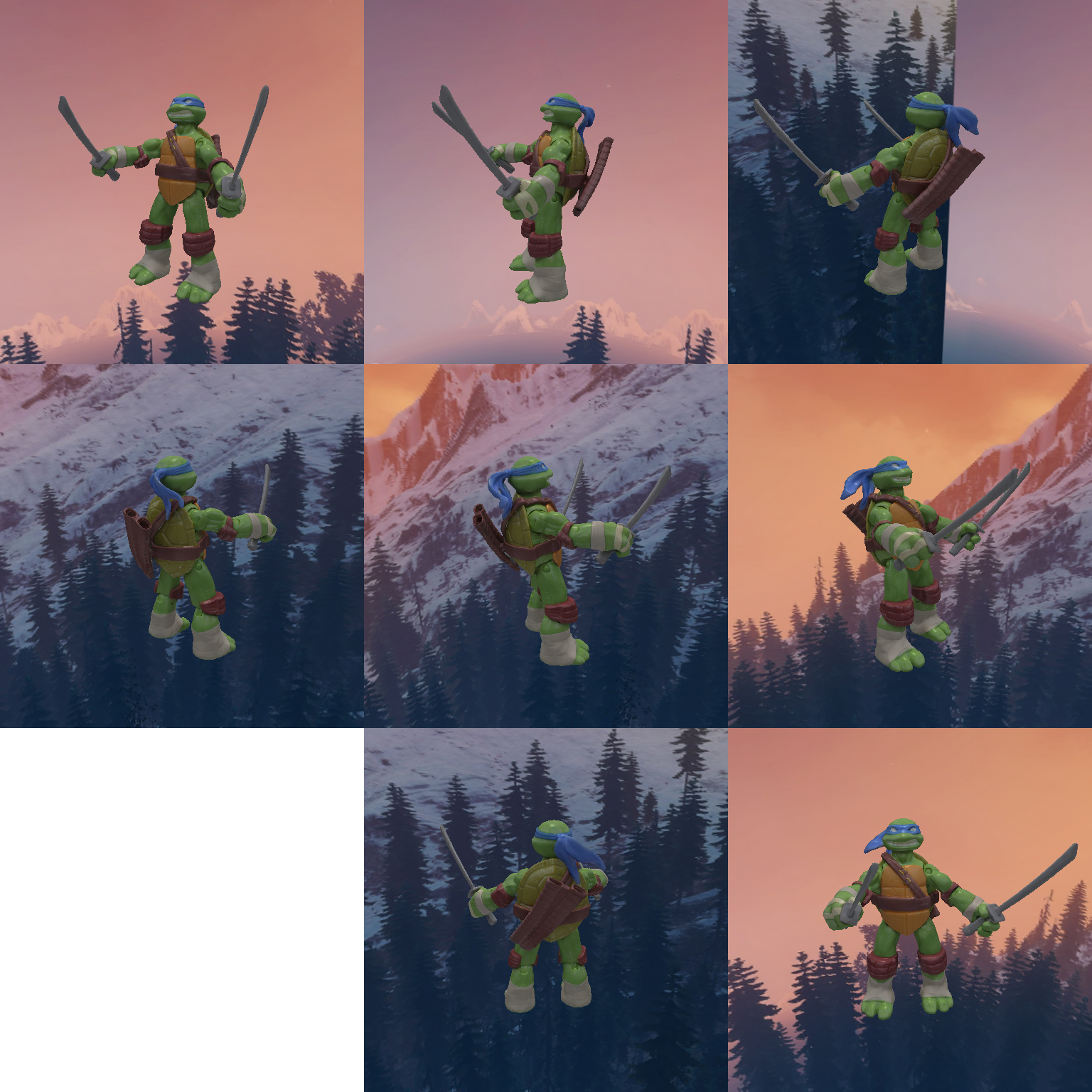}&
		\includegraphics[width = .12\textwidth,trim=1.0cm 1.2cm 2.cm 2cm,clip]{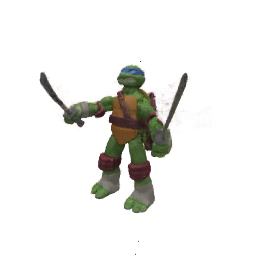}&
		\includegraphics[width = .12\textwidth,trim=1.0cm 1.2cm 2.cm 2cm,clip]{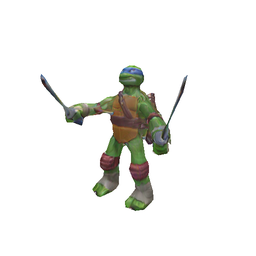}&
		\includegraphics[width = .12\textwidth,trim=1.0cm 1.2cm 2.cm 2cm,clip]{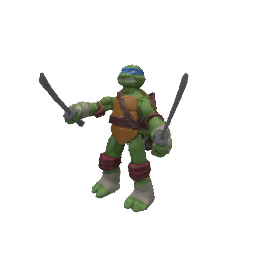}&
        \includegraphics[width = .12\textwidth,trim=1.0cm 1.2cm 2.cm 2cm,clip]{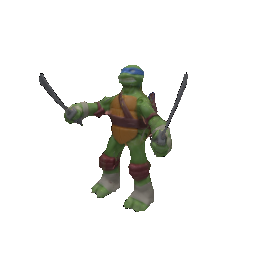}&
        \includegraphics[width = .12\textwidth,trim=1.0cm 1.2cm 2.cm 2cm,clip]{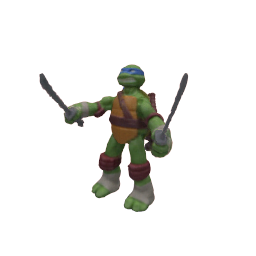}&
		\includegraphics[width = .12\textwidth,trim=1.0cm 1.2cm 2.cm 2cm,clip]{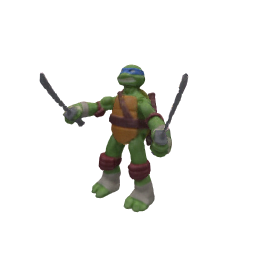}&
		\includegraphics[width = .12\textwidth,trim=1.0cm 1.2cm 2.cm 2cm,clip]{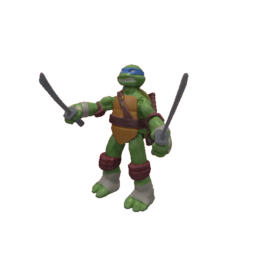}\\
        \includegraphics[width = .12\textwidth, trim=0.cm 0cm 0cm 0cm,clip]{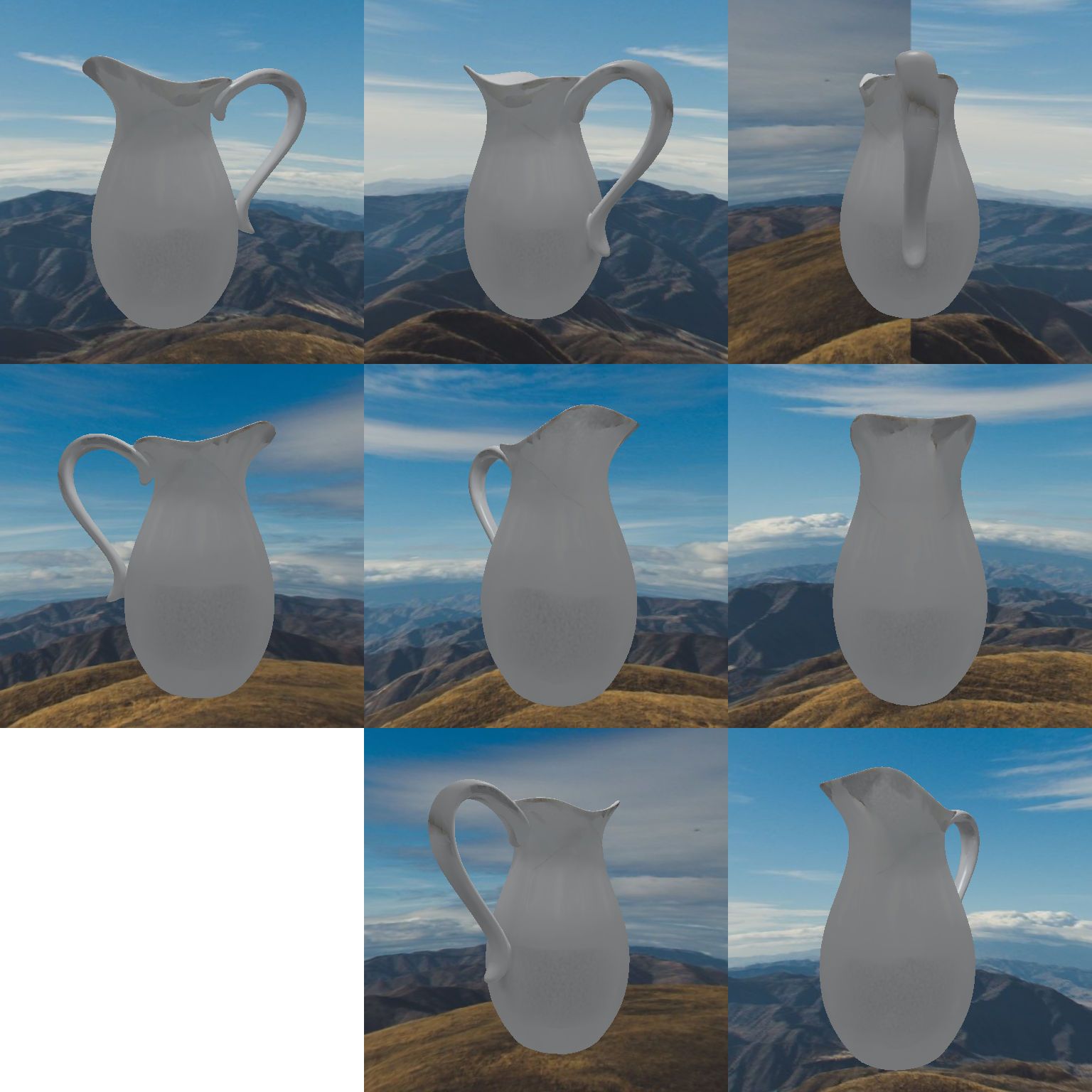}&
		\includegraphics[width = .12\textwidth, trim=1cm 0.5cm 1cm 0.5cm,clip]{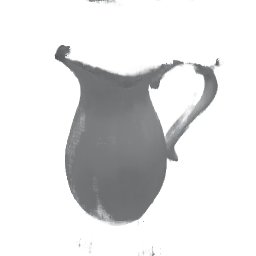}&
		\includegraphics[width = .12\textwidth, trim=0.4cm 0.cm 0.4cm 0.1cm,clip]{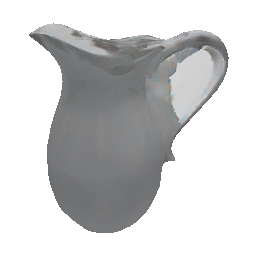}&
		\includegraphics[width = .12\textwidth, trim=1cm 0.5cm 1cm 0.5cm,clip]{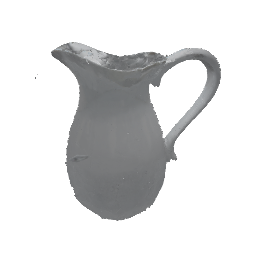} &
        \includegraphics[width = .12\textwidth,trim=1cm 0.5cm 1cm 0.5cm,clip]{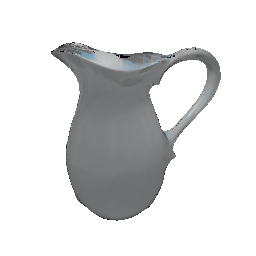} &
        \includegraphics[width = .12\textwidth,trim=1cm 0.5cm 1cm 0.5cm,clip]{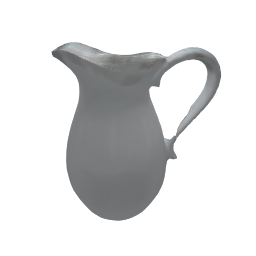}&
		\includegraphics[width = .12\textwidth,trim=1cm 0.3cm 1cm 1cm,clip]{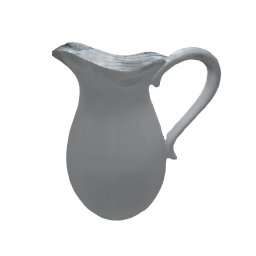}&
		\includegraphics[width = .12\textwidth,trim=1cm 0.5cm 1cm 0.5cm,clip]{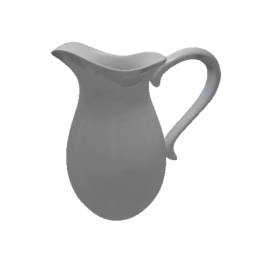}\\
        \includegraphics[width = .12\textwidth, trim=0.cm 0cm 0cm 0cm,clip]{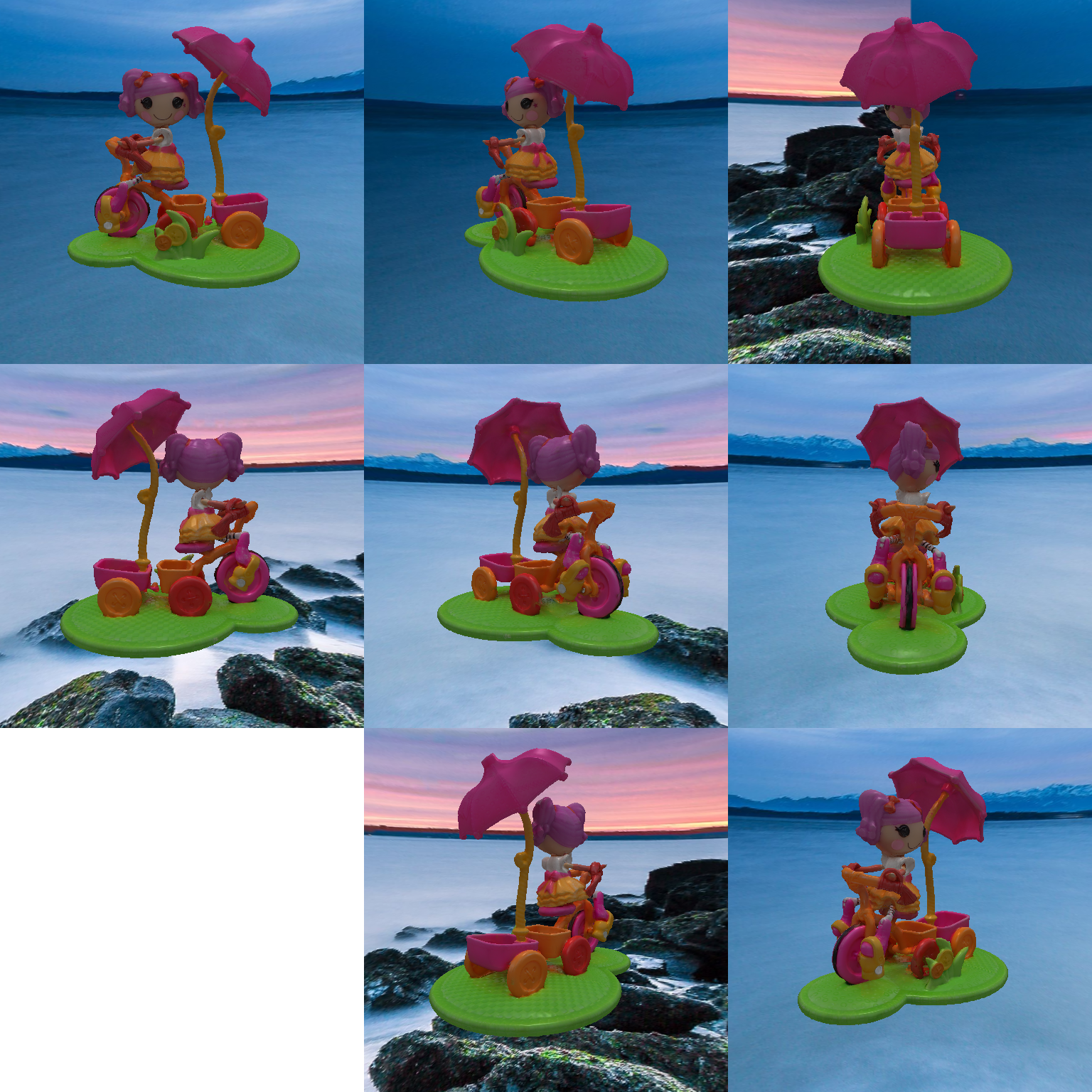}&
		\includegraphics[width = .12\textwidth, trim=1.7cm 1.7cm 1.7cm 0.3cm,clip]{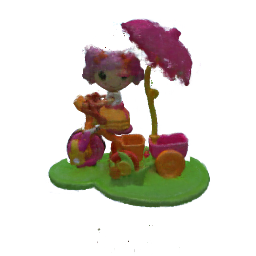}&
		\includegraphics[width = .12\textwidth,trim=1.3cm 0.7cm 1.3cm 0.3cm,clip]{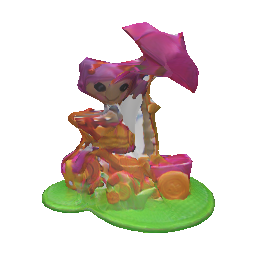}&
		\includegraphics[width = .12\textwidth,trim=1.7cm 1.7cm 1.7cm 0.3cm,clip]{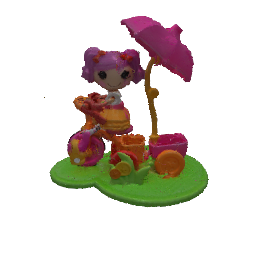}&
        \includegraphics[width = .12\textwidth,trim=1.7cm 1.7cm 1.7cm 0.3cm,clip]{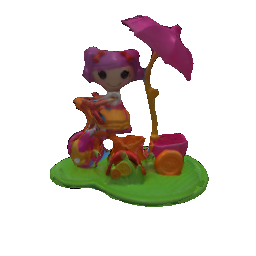}&
        \includegraphics[width = .12\textwidth,trim=1.7cm 1.7cm 1.7cm 0.3cm,clip]{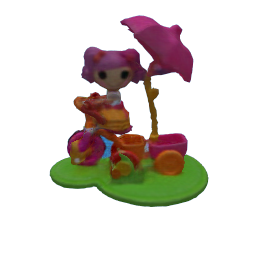}&
		\includegraphics[width = .12\textwidth,trim=1.7cm 1.7cm 1.7cm 0.3cm,clip]{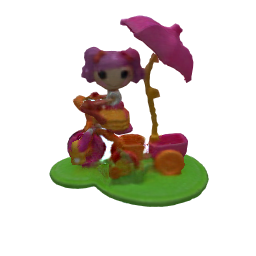}&
		\includegraphics[width = .12\textwidth,trim=1.7cm 1.7cm 1.7cm 0.3cm,clip]{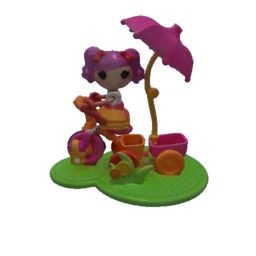}\\
		Input Images&RegNeRF & NeRS & Munkberg et al.~\cite{munkberg2021nvdiffrec} & DS & our w/BCG* & our wo/BCG & GT

    \end{tabular}		

    \end{center}
	\caption{Qualitative Results on the GSO Dataset. Note that in our w/BCG* column we remove the background rendering in the w/BCG column to simplify the visual comparisons.\label{fig:qualitative_results_google}}
\end{figure*}

\begin{figure*}
    \captionsetup{font={normalsize}}
	\begin{center}
	\begin{tabular}{cccccccc}
		\includegraphics[width = .12\textwidth, trim=0cm 0cm 0cm 0cm,clip]{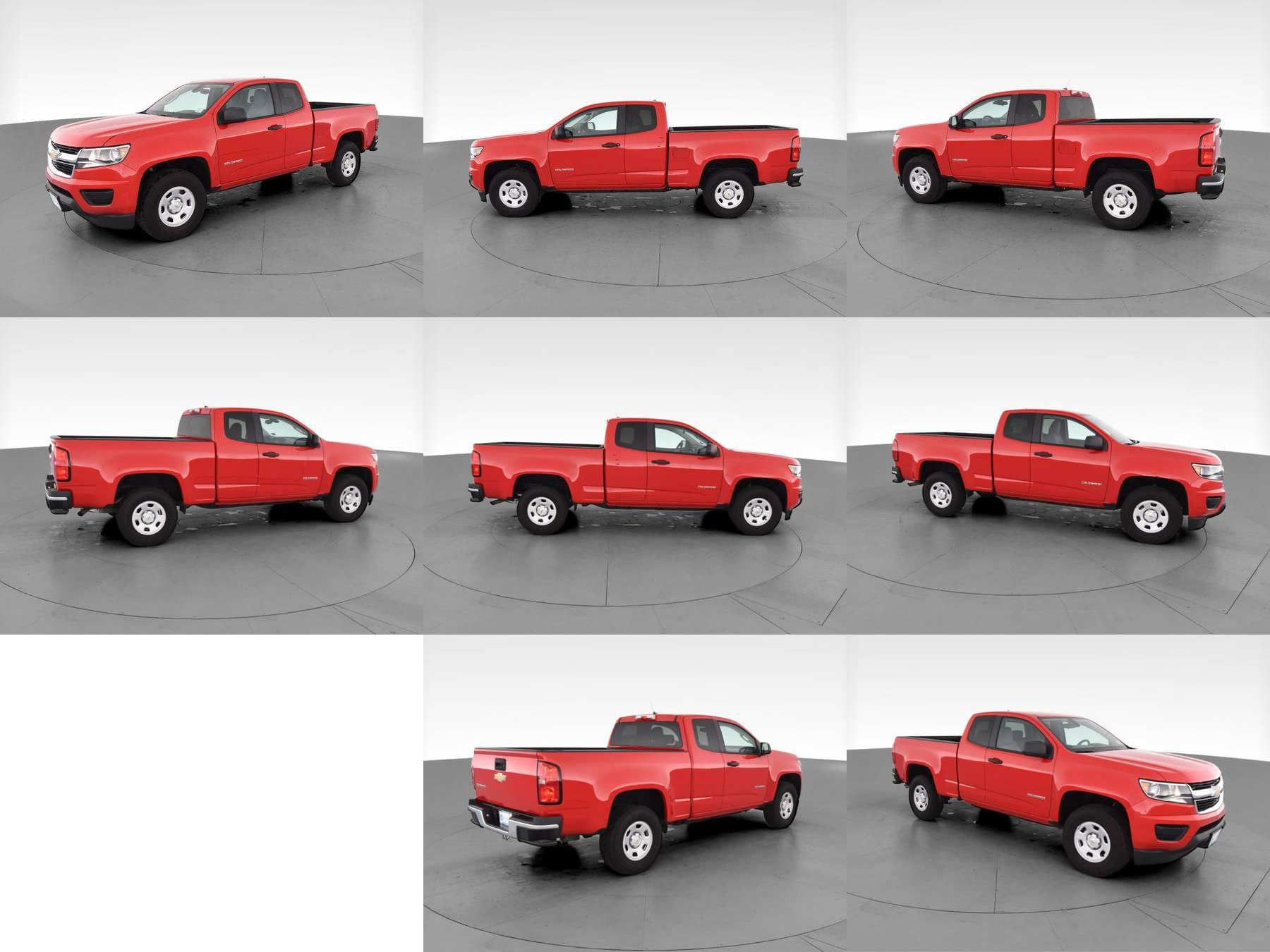} &
        \includegraphics[width = .12\textwidth, trim=0.9cm 1.5cm 0.9cm 1.5cm,clip]{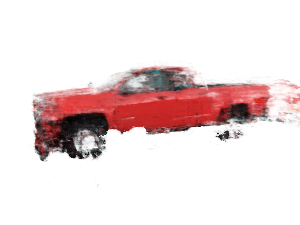}&
		\includegraphics[width = .12\textwidth, trim=0.5cm 3cm 0.5cm 3.cm,clip]{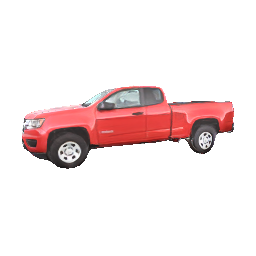}&
		\includegraphics[width = .12\textwidth, trim=0.9cm 1.5cm 0.9cm 1.5cm,clip]{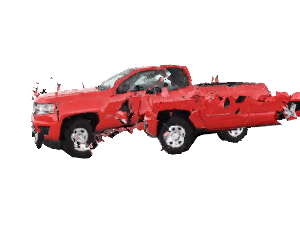}&
        \includegraphics[width = .12\textwidth, trim=0.9cm 1.5cm 0.9cm 1.5cm,clip]{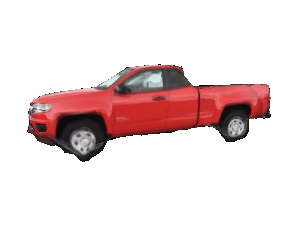}
        &
        \includegraphics[width = .12\textwidth, trim=0.9cm 1.5cm 0.9cm 1.5cm,clip]{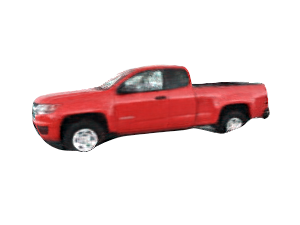}&
		\includegraphics[width = .12\textwidth, trim=0.9cm 1.5cm 0.9cm 1.5cm,clip]{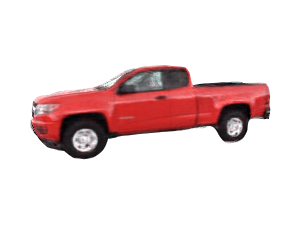} &
		\includegraphics[width = .12\textwidth, trim=0.9cm 1.5cm 0.9cm 1.5cm,clip]{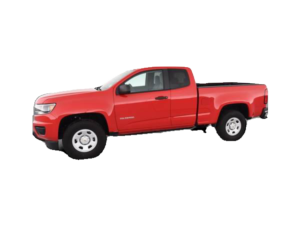} \\
        \includegraphics[width = .12\textwidth, trim=0cm 0cm 0cm 0cm,clip]{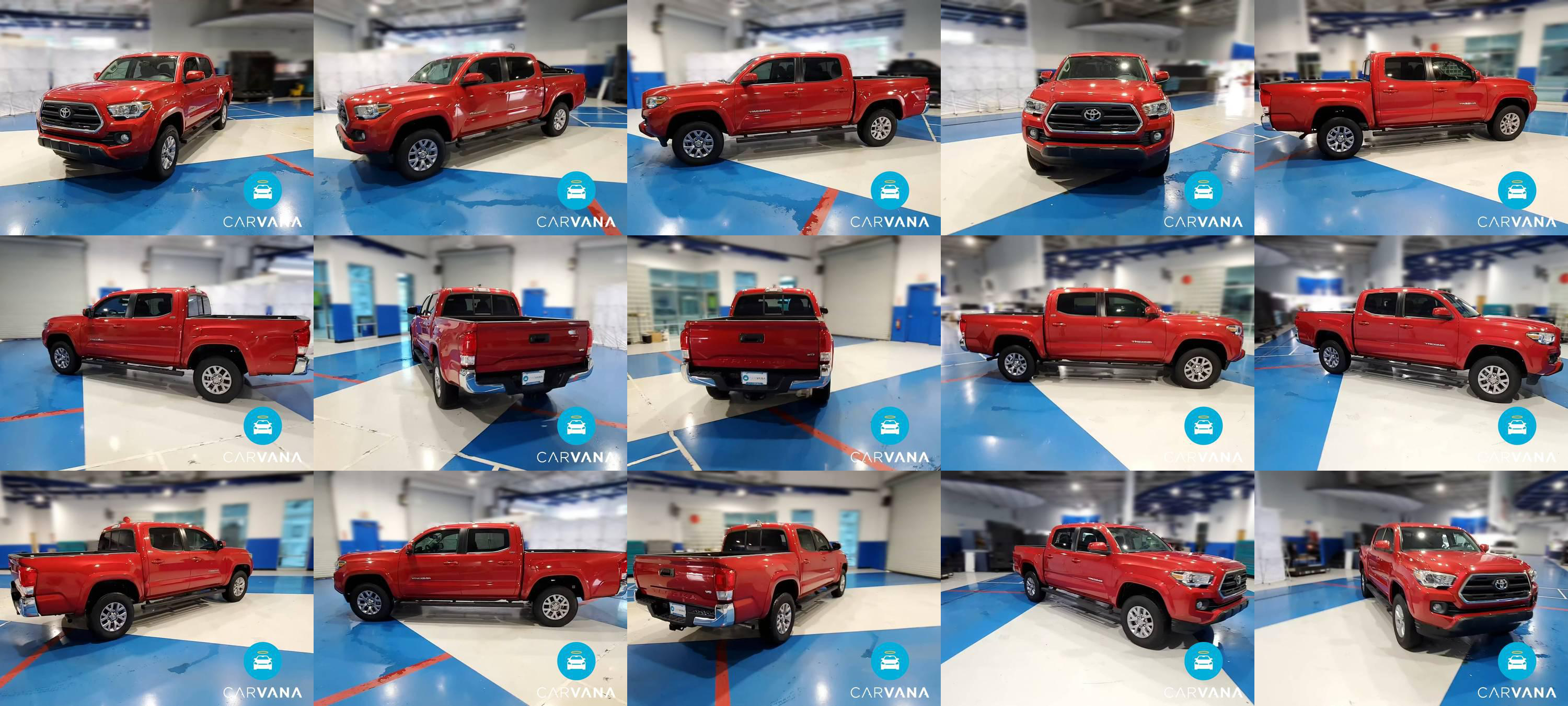}&
		\includegraphics[width = .12\textwidth, trim=1.5cm 1cm 0.2cm 1.5cm,clip]{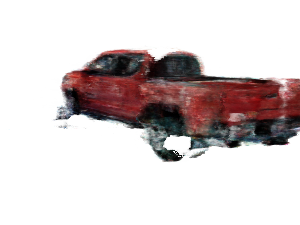}&
		\includegraphics[width = .12\textwidth, trim=0.cm 2cm 0.5cm 2.cm,clip]{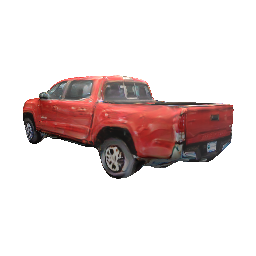}&
		\includegraphics[width = .12\textwidth, trim=1.5cm 1cm 0.2cm 1.5cm,clip]{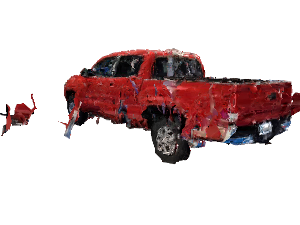}&
        \includegraphics[width = .12\textwidth, trim=1.5cm 1cm 0.2cm 1.5cm,clip]{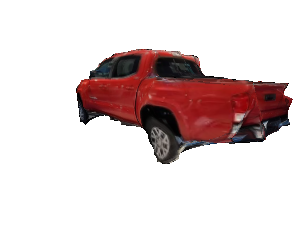}
         &
        \includegraphics[width = .12\textwidth, trim=1.5cm 1cm 0.2cm 1.5cm,clip]{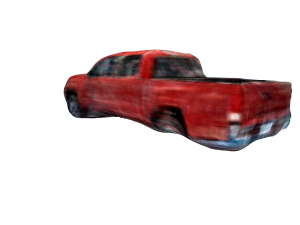}&
		\includegraphics[width = .12\textwidth, trim=1.5cm 1cm 0.2cm 1.5cm,clip]{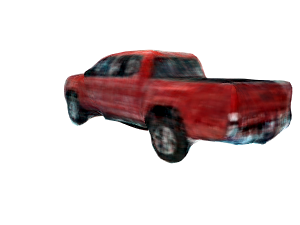} &
		\includegraphics[width = .12\textwidth, trim=1.5cm 1cm 0.2cm 1.5cm,clip]{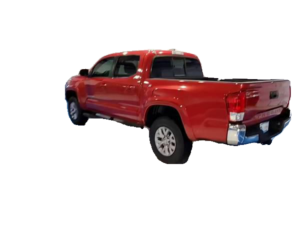} \\
        \includegraphics[width = .12\textwidth, trim=0cm 0cm 0cm 0cm,clip]{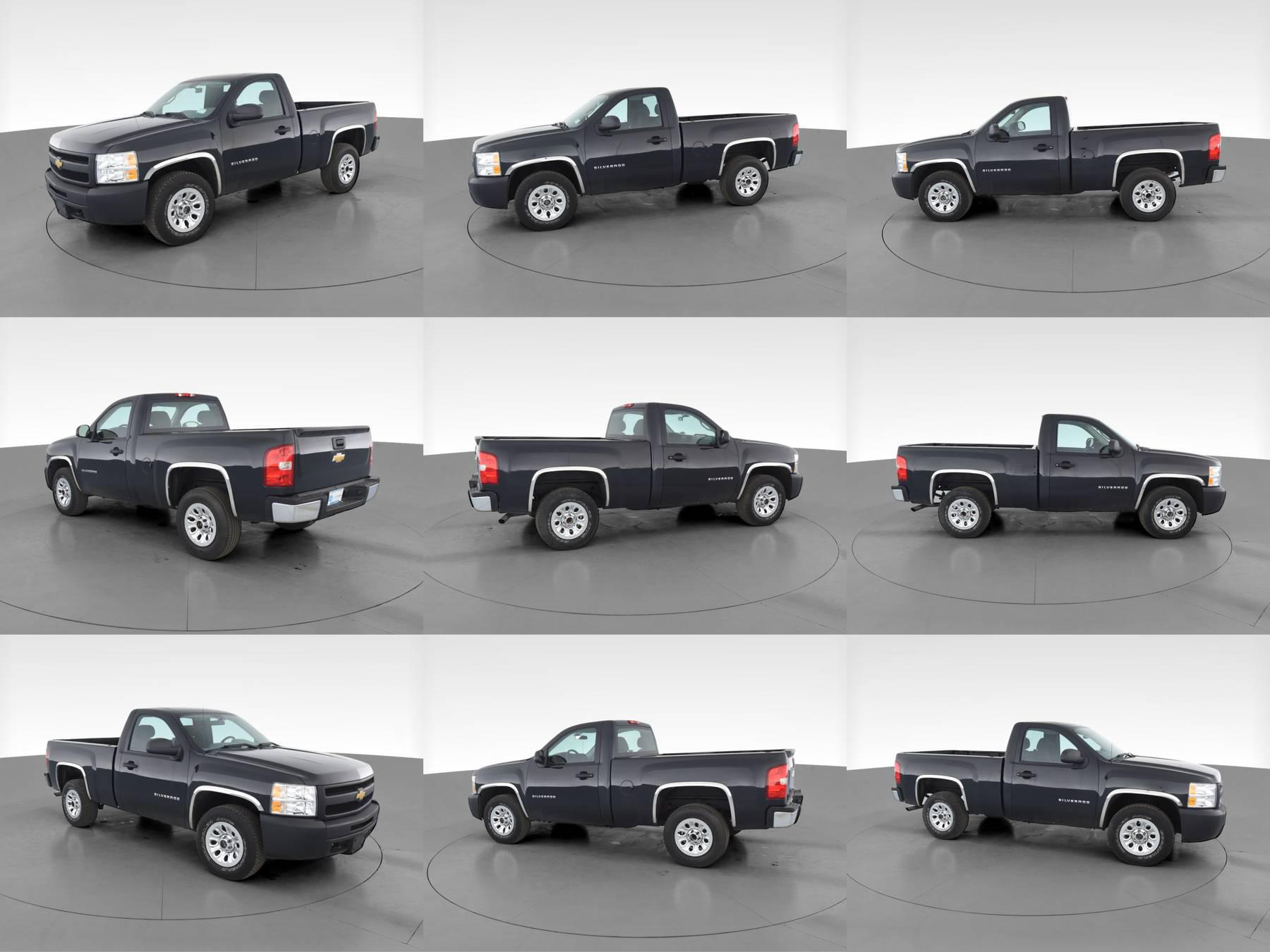}&
		\includegraphics[width = .12\textwidth, trim=0.9cm 1.5cm 0.9cm 1.5cm,clip]{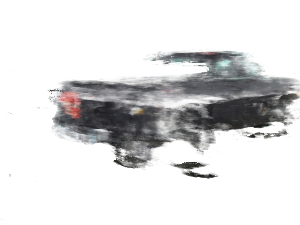}&
		\includegraphics[width = .12\textwidth, trim=0.5cm 2cm 0.5cm 2.cm,clip]{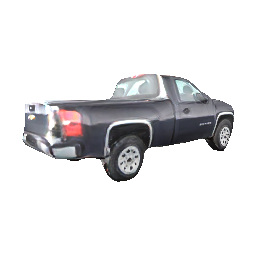}&
		\includegraphics[width = .12\textwidth, trim=0.9cm 1.5cm 0.9cm 1.5cm,clip]{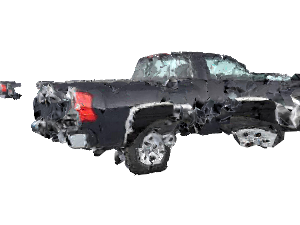}&
        \includegraphics[width = .12\textwidth, trim=0.9cm 1.5cm 0.9cm 1.5cm,clip]{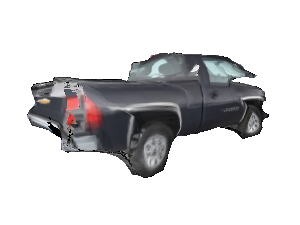}&
        \includegraphics[width = .12\textwidth, trim=0.9cm 1.5cm 0.9cm 1.5cm,clip]{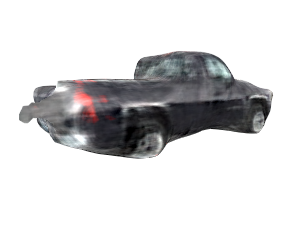}
&
		\includegraphics[width = .12\textwidth, trim=0.9cm 1.5cm 0.9cm 1.5cm,clip]{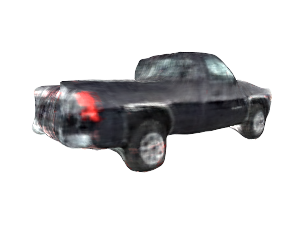} &
		\includegraphics[width = .12\textwidth, trim=0.9cm 1.5cm 0.9cm 1.5cm,clip]{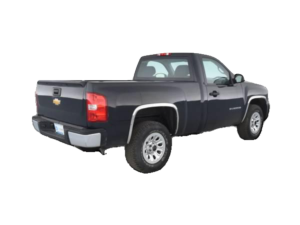} \\
		Input &RegNeRF & NeRS* & Munkberg et al. \cite{munkberg2021nvdiffrec} &DS & our w/BCG*  & our wo/BCG & GT
    \end{tabular}	
    \end{center}

	\caption{Qualitative Results on the MVMC Car dataset. NeRS*: the original NeRS implementation crops the images before training and thus changes their aspect ratio during training. Thus, the rendered images have an aspect ratio of 1, while the original ones do not. For more information, see section~\ref{sec:evaluation}. Note that we remove the background rendering in the w/BCG* column to simplify the visual comparisons.  
	}
	\label{fig:qualitative_results_mvmc_car}
    
\end{figure*}

\subsection{Evaluation}
\label{sec:evaluation}
\noindent\textbf{Metrics.} For the datasets with a 3D ground truth, we compare the reconstructed meshes with the ground-truth meshes or point clouds.
More specifically, we report the L2-Chamfer and L1-Chamfer distances, normal consistency, and F1 score, following \cite{meshrcnn}.
We also report texture metrics to evaluate the quality of the texture on unseen views.
More specifically, we employ Mean-Square
Error (MSE), Peak Signal-to-Noise Ratio (PSNR), and Structural Similarity Index Measure (SSIM), and Learned Perceptual Image Patch Similarity (LPIPS) \cite{zhang2018unreasonable}.
\\
\noindent \textbf{Baselines.}
We compare our method with the following methods: (1) RegNerf~\cite{regnerf_2021}, a volume rendering method, (2) Munkberg et al.~\cite{munkberg2021nvdiffrec}, a hybrid-based method, (3) DS~\cite{diff_stereopsis_2021_arxiv}, 
a mesh-based method, (4) COLMAP~\cite{schoenberger2016mvs, Schonberger_2016_CVPR}, a multi-view stereo method, (5) NeUS~\cite{wang2021neus}, neural surface reconstruction method, and (6) NeRS~\cite{NERS_2021_Neurips}  a neural reflectance surface method. 
Further details about these baseline methods can be found in the supplementary material.

\renewcommand{\tabcolsep}{0pt}
\begin{figure*}[t]
	\begin{center}
	\begin{tabular}{ccccccccc}
        \includegraphics[width=.11\linewidth,trim=0cm 0cm 0cm 0cm,clip]{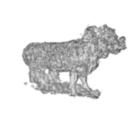} &
		\includegraphics[width=.11\linewidth,trim=0cm 0cm 0cm 0cm,clip]{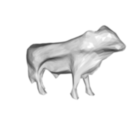} &
		\includegraphics[width=.11\linewidth,trim=0cm 0cm 0cm 0cm,clip]{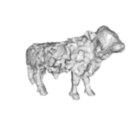} &
		\includegraphics[width=.11\linewidth,trim=0cm 0cm 0cm 0cm,clip]{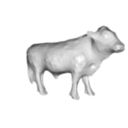} &
        \includegraphics[width=.11\linewidth,trim=0cm 0cm 0cm 0cm,clip]{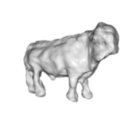} &
        \includegraphics[width=.11\linewidth,trim=0cm 0cm 0cm 0cm,clip]{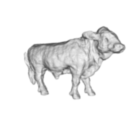} &
        \includegraphics[width=.11\linewidth,trim=0cm 0cm 0cm 0cm,clip]{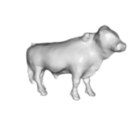} &
		\includegraphics[width=.11\linewidth,trim=0cm 0cm 0cm 0cm,clip]{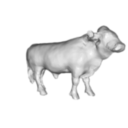} &
		\includegraphics[width=.11\linewidth,trim=0cm 0cm 0cm 0cm,clip]{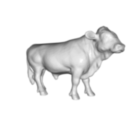}\\
        \scriptsize{RegNeRF} & \scriptsize{NeRS} & \scriptsize{Munkberg et al.} & \scriptsize{DS} & \scriptsize{NeuS w/BCG} & \scriptsize{NeuS wo/BCG} & \scriptsize{Our w/BCG} & \scriptsize{Our wo/BCG} & \scriptsize{GT}
    \end{tabular}
    \end{center}
	\caption{Reconstructed meshes for the \textit{COW} object in the GSO Dataset.\label{fig:qualitative_results_cow}} 
\end{figure*}

\renewcommand{\tabcolsep}{2pt}
\begin{table}[t]
\centering
\caption{Tank and Temples Dataset: Quantitative evaluation of generated views on the test set.} \label{table:TankTemple_Texture_Metric}
\small
\begin{tabular}{@{}c c c c c c c@{}}
\toprule
Scene & Method & Mask & PSNR $\uparrow$ & MSE $\downarrow$ & SSIM $\uparrow$ & LPIPS $\downarrow$ \\
\hline
\multirow{5}{*}{Truck} & RegNeRF & yes & 18.078 & 0.018 & 0.657 & 0.254 \\
& Munkberg et al. & yes & \textbf{18.398} & 0.018 & 0.673 & \textbf{0.245} \\
& Our wo\slash BCG & yes & 18.315 & \textbf{0.017} & \textbf{0.701} & 0.252 \\
& Our w\slash BCG & no & 13.168 & 0.0508 & 0.666 & 0.343 \\
\hline
\multirow{5}{*}{Ignatius} & RegNeRF & yes & 21.123 & 0.105 & 0.866 & 0.108 \\
& Munkberg et al. & yes & 22.720 & \textbf{0.007} & 0.873 & \textbf{0.073} \\
& Our wo\slash BCG & yes & \textbf{23.022} & \textbf{0.007} & \textbf{0.896} & 0.080 \\
& Our w\slash BCG & no & 17.845 & 0.021 & 0.875 & 0.153 \\
\hline
\end{tabular}
\end{table}

\renewcommand{\tabcolsep}{3pt}
\begin{table}
\begin{center}
\caption{Tank and Temples Dataset: 3D metrics. *Note that we obtained the COLMAP*-reconstructed point clouds using 50 views. \label{table:TankTemple_3D_Metric}}
\small
\begin{tabularx}{\columnwidth}{l c >{\centering\arraybackslash}X  >
{\centering\arraybackslash}X  >
{\centering\arraybackslash}X >{\centering\arraybackslash}X}
 \toprule
 Scene&Method & Mask & Chamfer-L2 $\downarrow$ & Chamfer-L1 $\downarrow$   & F@$10$  $\uparrow$ \\  \hline
 \multirow{5}{*}{Truck}
 &RegNeRF-clean &yes  &  0.059 & 0.342 & 42.56   \\
 &Munkberg  et al. 
  & yes & 0.072 & 0.355 & 50.11   \\
& DS &  yes & 0.110  & 0.479 &  31.73 \\
  & COLMAP* &  yes & \textbf{0.056} & \textbf{0.298} &   \textbf{57.74} \\
  & NeuS &  no &  3.342 & 2.417 & 6.50  \\
  & NeuS &  yes &  0.629 & 1.253 & 11.93   \\
  & Our wo\slash BCG  & yes  & 0.094 &0.406 &48.14 \\
 &Our w\slash BCG   & no & 0.225 & 0.613 & 45.78 \\\hline
  \multirow{5}{*}{Ignatius}
 &RegNeRF-clean &yes  & 0.106 & 0.423 &43.45   \\
 &Munkberg  et al. 
  &  yes & 0.022 & 0.189 & 81.75    \\
  & DS &  yes & 0.024  & 0.207 &  77.68  \\
  & COLMAP* &  yes & 0.013 & 0.166 &   86.90 \\
  & NeuS &  no &  0.155 &  0.572 &   31.95  \\
  & NeuS &  yes & 0.061  &  0.3878 &  37.72   \\
  & Our wo\slash BCG  & yes   & \textbf{0.018}  & \textbf{0.147}&\textbf{ 87.12} \\
 &Our w\slash BCG   & no  & 0.139 &0.480 & 55.32  \\\hline
\end{tabularx}
\end{center}
\vspace{-0.5cm}
\end{table}
\subsection{Results}

We run our algorithm under two settings: with background (\textit{w\slash BCG}) and by removing the background (\textit{wo\slash BCG}), the latter of which is akin to using a mask.
We present both qualitative (Figure~\ref{fig:qualitative_results_google} and Figure~\ref{fig:qualitative_results_cow}) and quantitative (Table~\ref{table:GSO_3D_Metric} and Table~\ref{table:GSO_Texture_Metric}) result on the GSO dataset.
We observe that our proposed method for both \textit{w\slash BCG} and \textit{wo\slash BCG} is able to recover the original shape with high accuracy. 
DS, Munkberg et al.~\cite{munkberg2021nvdiffrec} and NeuS (with mask supervision) show the closest performance to ours.
We observe that NeuS without mask supervision struggles to accurately reconstruct the original shape for seven out of the fourteen objects.
NeRS is also able to recover the shape, but cannot recover genus 1 objects. 
RegNeRF shows blur artifacts as a result of the inherent ambiguity of sparse input data and also may miss some parts of the original object, \eg, the leg of the cow object. 
RegNeRF does not always recover the thin parts of the object, \eg, legs of the horse, and thus the reconstructed geometry is not fully accurate.
When we run COLMAP with 8 views we find that for most objects the reconstructed point cloud is mostly empty and the object is not recognizable (see also the supplementary material for visual results). This is not surprising since we have only 8 views covering 360-degrees of the object. Furthermore, in this setting, any surface is visible at most from 3 views and the objects does not have a rich texture. Because of this reason, we run COLMAP with 50 views and present the results in all tables 
just for reference.

In Figure~\ref{fig:qualitative_results_mvmc_car} and Table~\ref{table:MVMC_Texture_Metric} we present qualitative and quantitative results for car objects on MVMC Car dataset. 
We qualitatively observe that our method wo/BCG shows better view renderings than  other methods.
Our method fails to recover the texture around transparent surfaces, \eg, the car window. 
The model in the w/BCG case is able to recover the main shape of the car, but it misses some  parts, \eg, the tires of the car that are attached to the ground due to overlaps with the background mesh. 
Additionally, the tires contain poor texture, \eg, mostly they are black, so this can be easily captured by the background texture network and thus introduce an ambiguity in the reconstruction. 
We note that the performance for NeRS is consistent across different cars.
This is not surprising as they use the mask and their initial template is also car-like. 
We do not run COLMAP on this dataset as the dataset has a limited number of views.
DS has weaker performance on this dataset compared to the GSO. The main reason for poor texture quality is that texture is obtained by 3D back-projections of the mesh to the input views. Thus, incorrect geometry leads to poor texture quality.
We observe that the quality of the reconstructions from RegNeRF and Munkberg et al.~\cite{munkberg2021nvdiffrec} lack realism.
There are two main reasons for this. 
Firstly, the camera locations are not uniformly spread out around the object. 
Most of them are located on two sides of the cars. 
In this case, the methods struggle to recover the original shape. 
Secondly, because of the many light sources present in the scene and non-Lambertian surfaces, the multi-view consistency across the views is not satisfied.
As can be observed, our method is more robust to the above issues.
The main reason for that is that during the training of ISNN, the output color of TNN does not depend on the camera view and thus it is less prone to overfitting.

In Table~\ref{table:TankTemple_3D_Metric} and Table~\ref{table:TankTemple_Texture_Metric} we present quantitative results for two objects in the Tank and Temple datasets. 
Note that the performance of all methods is drastically decreased especially in the recovered 3D shape compared to the GSO dataset.
This is expected as the ground-truth point clouds are hollow (without the bottom), \eg \textit{Truck}, and the reported numbers only approximate the quality of the shape. 
Our \textit{wo\slash BCG} has a higher Chamfer distance compared to the others although it looks visually better. 
This is because the corresponding ground-truth shape does not only include the target object, but also some other components from the background, as, \eg, in the \textit{Truck} scene.  
For visual results and more details, see the supplementary material.

\section{Conclusion}
We have introduced a novel multi-view stereo method that works with sparse views from a 360 rig. The method can handle this extreme setting by using a novel object-centric sampling scheme and a corresponding hybrid surface representation. The sampling scheme allows to concentrate the updates due to multiple camera views to the same components of the surface representation and to structure the updates so that they result in useful surface changes (along its normals, rather than its tangent space).
We have demonstrated the robustness of this method by working without the common mask supervision constraint, by using
datasets with diverse 3D objects (GSO Dataset), on scenes with complex illumination sources and with non-Lambertian surfaces (MVMC Car).

\noindent\textbf{Acknowledgements.} This work was supported by grant 188690 of the Swiss National Science Foundation
{
    \small
    \bibliographystyle{ieeenat_fullname}
    \bibliography{main}
}

\appendix
\setcounter{section}{0} 

\section{Comparing with Baselines}
For all baselines, we run them using their default hyperparameters unless explicitly specified.

\noindent\textbf{RegNeRF \cite{regnerf_2021}\footnote{\url{https://github.com/google-research/google-research/tree/master/regnerf}}.} The released implementation does not include the appearance regularization loss. 
We recover the geometry from the predicted density by running the marching cubes \cite{marching_cubes} algorithm.
We note that the extracted mesh contains noisy parts at times. 
For a fair comparison, we clean out these parts manually and report the results for the manually cleaned mesh.
We employ the iterative-closest-point (ICP) \cite{ICP} algorithm to align the predicted mesh to the ground truth. 
We train the models by removing the background.

\noindent\textbf{Munkberg et al. \cite{munkberg2021nvdiffrec}\footnote{\url{https://github.com/NVlabs/nvdiffrec}}.} We run the method without any modification.

\noindent\textbf{NeRS \cite{NERS_2021_Neurips}\footnote{\url{https://github.com/jasonyzhang/ners}}.} The original implementation of the code crops the objects around their mask and thus changes the original aspect ratio of the original images. 
For a fair comparison, we run our method for both original images and modified images. 

\noindent\textbf{DS \cite{diff_stereopsis_2021_arxiv}\footnote{\url{https://github.com/shubham-goel/ds}}.} We optimize only the mesh and the texture as the camera poses are given. 

\noindent\textbf{COLMAP~\cite{schoenberger2016mvs, Schonberger_2016_CVPR}\footnote{\url{https://colmap.github.io/}}} We run the COLMAP by removing the background as our aim is to estimate the 3D of the object in the scene.

\noindent\textbf{NeuS \cite{wang2021neus}\footnote{\url{https://github.com/Totoro97/NeuS}}}. We run the NeuS under two settings: with mask supervision and without mask supervision. 
As our aim is to recover the 3D of an object we define the regions of interest by pointcloud of COLMAP output for both settings as it is explained in the original repository.

\section{Ablations}


\section{Ablations}

\begin{figure}
	\begin{center}
	\begin{tabular}{c|c|c|c}
        \includegraphics[width = .12\textwidth, trim=0.cm 0cm 0cm 0cm,clip]{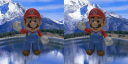}&
        \includegraphics[width = .12\textwidth, trim=0.cm 0cm 0cm 0cm,clip]{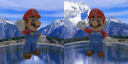}&
        \includegraphics[width = .12\textwidth, trim=0.cm 0cm 0cm 0cm,clip]{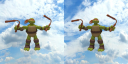}&
        \includegraphics[width = .12\textwidth, trim=0.cm 0cm 0cm 0cm,clip]{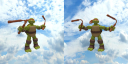} \\
        \includegraphics[width = .12\textwidth, trim=0.cm 4cm 0cm 4cm,clip]{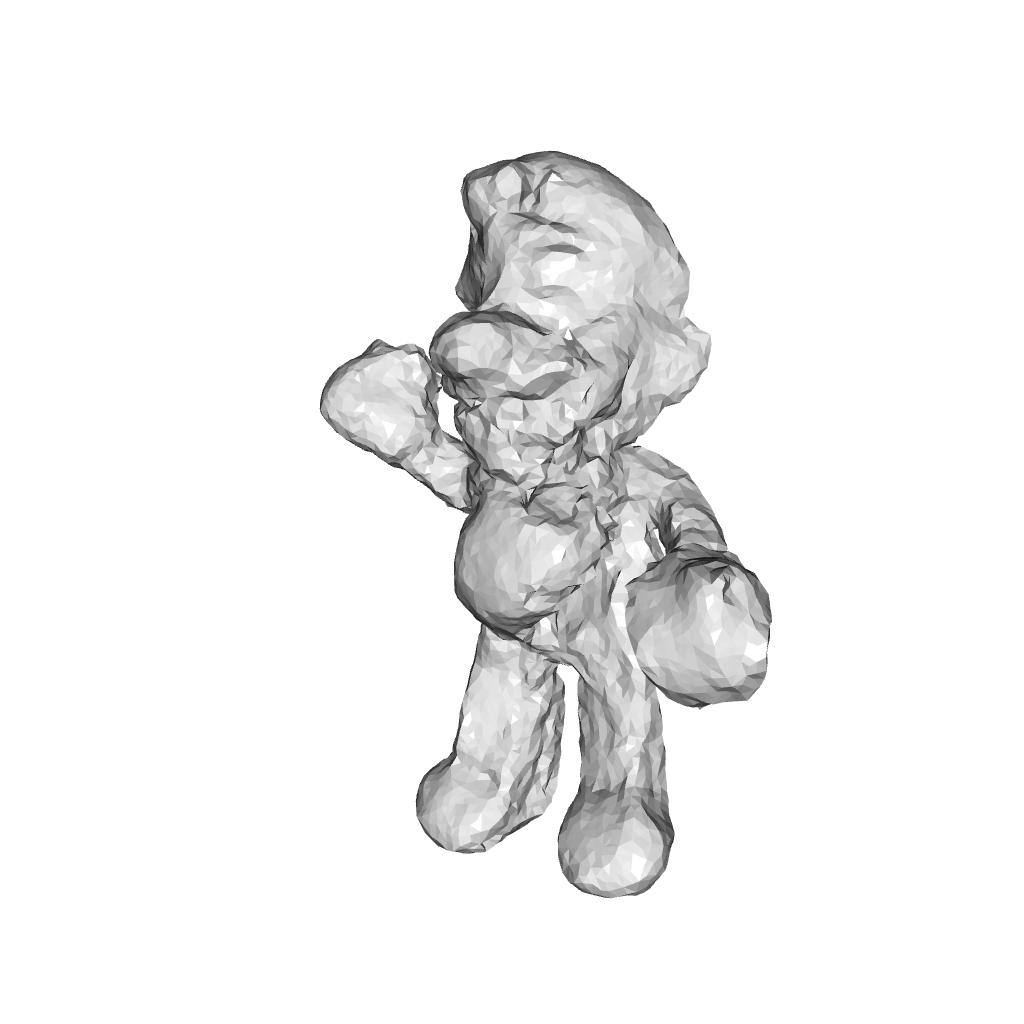}&
        \includegraphics[width = .12\textwidth, trim=0.cm 4cm 0cm 4cm,clip]{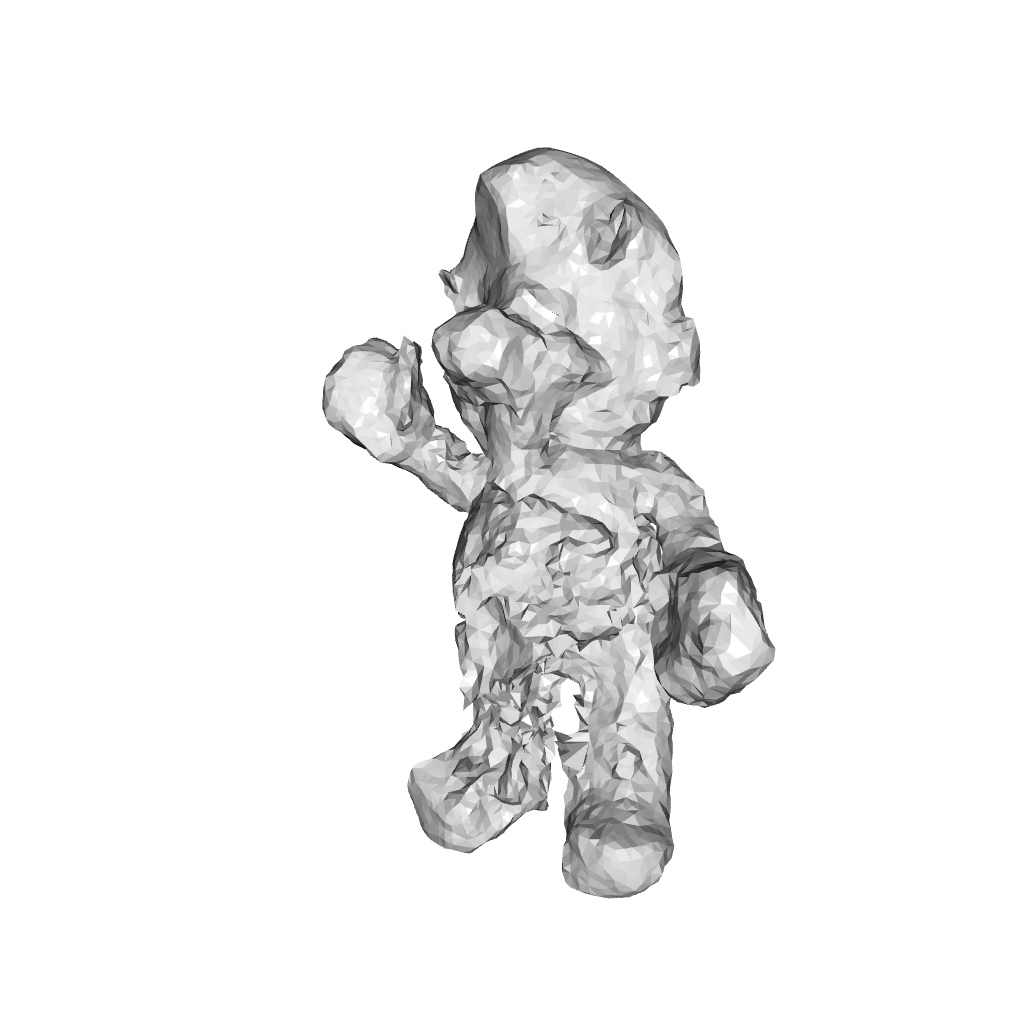}&
        \includegraphics[width = .12\textwidth, trim=0.cm 4cm 0cm 4cm,clip]{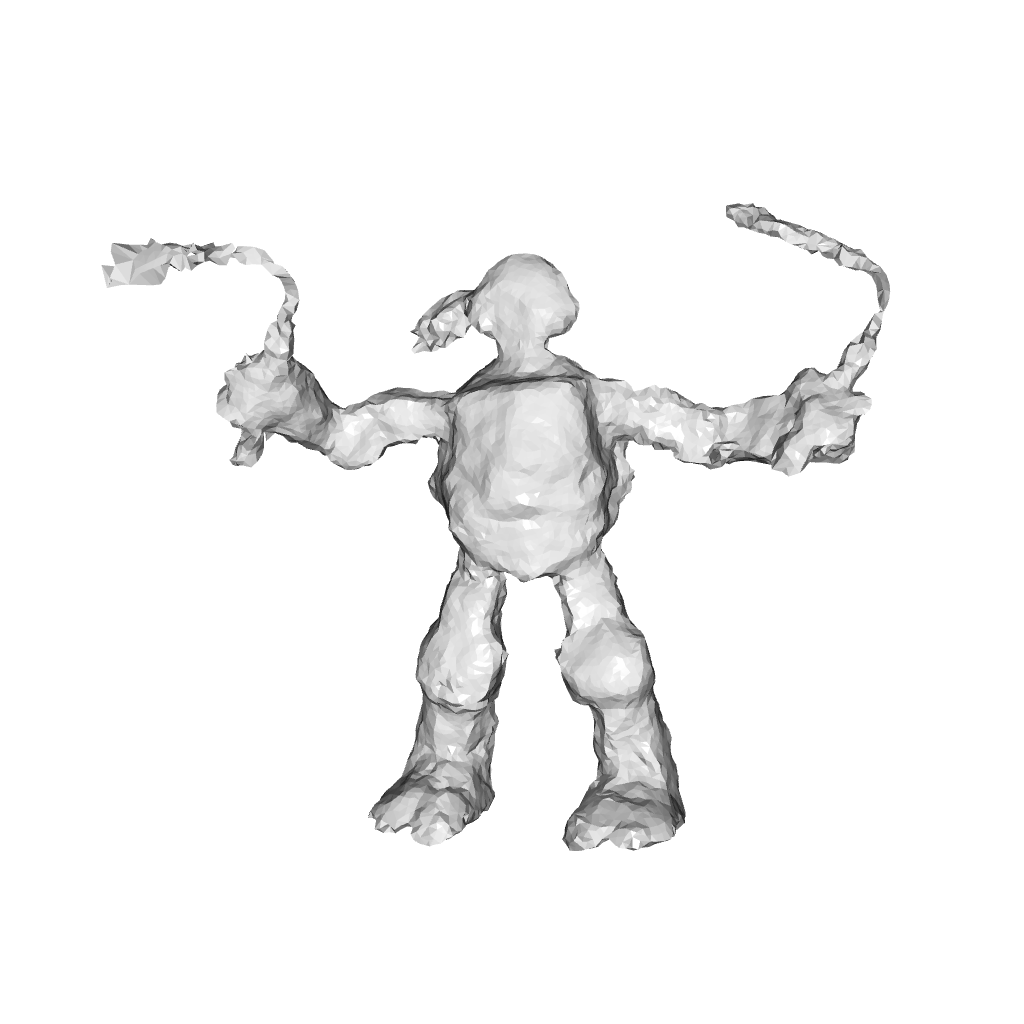}&
        \includegraphics[width = .12\textwidth, trim=-2.5cm 4cm 0cm 4cm,clip]{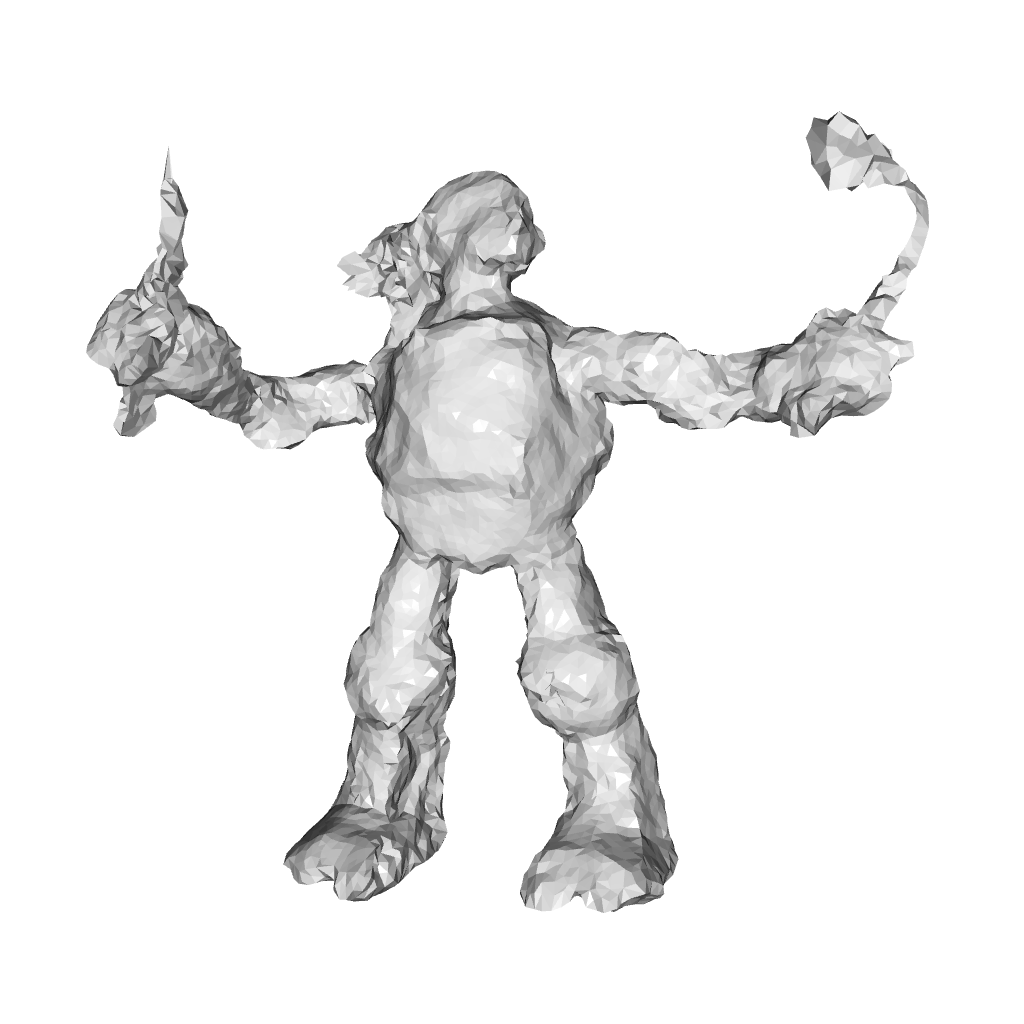} \\
        
        \includegraphics[width = .1\textwidth, trim=0.cm 4cm 0cm 4cm,clip]{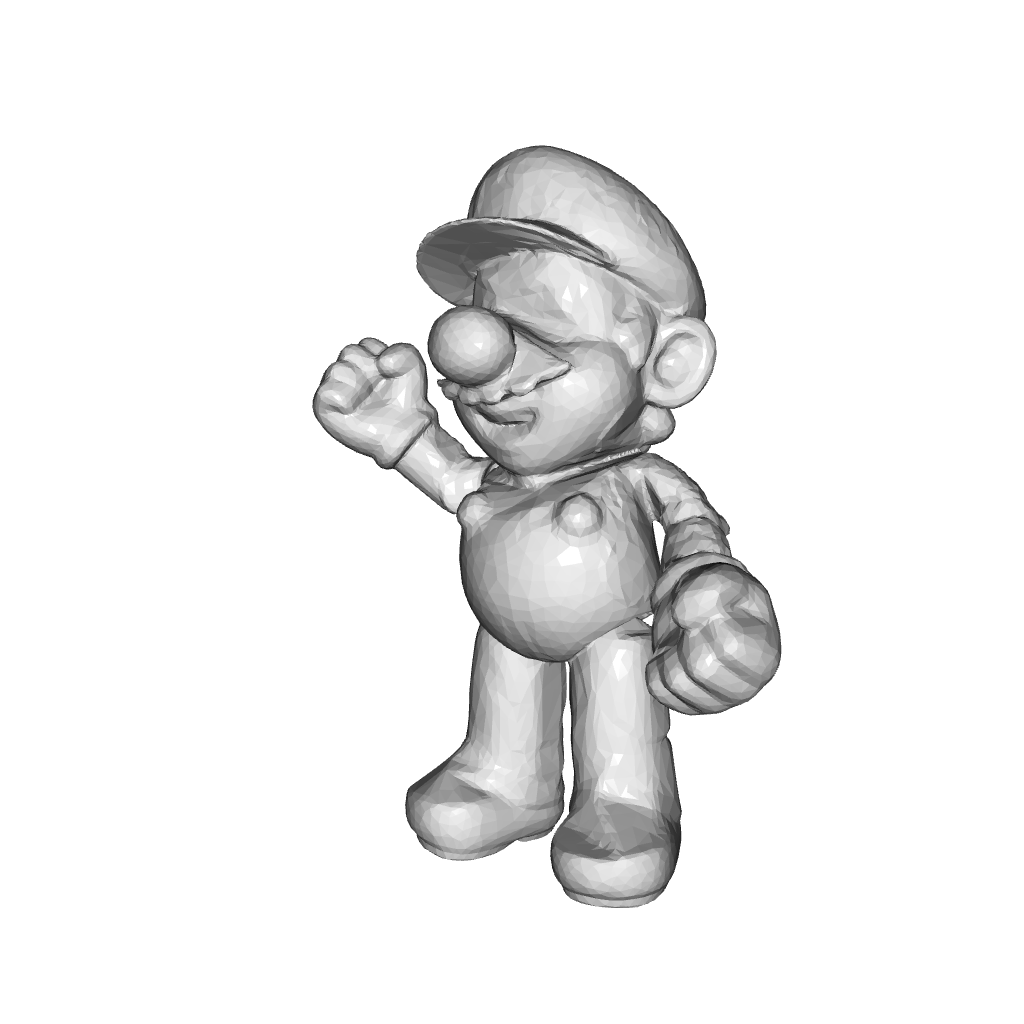}&
        \includegraphics[width = .1\textwidth, trim=0.cm 4cm 0cm 4cm,clip]{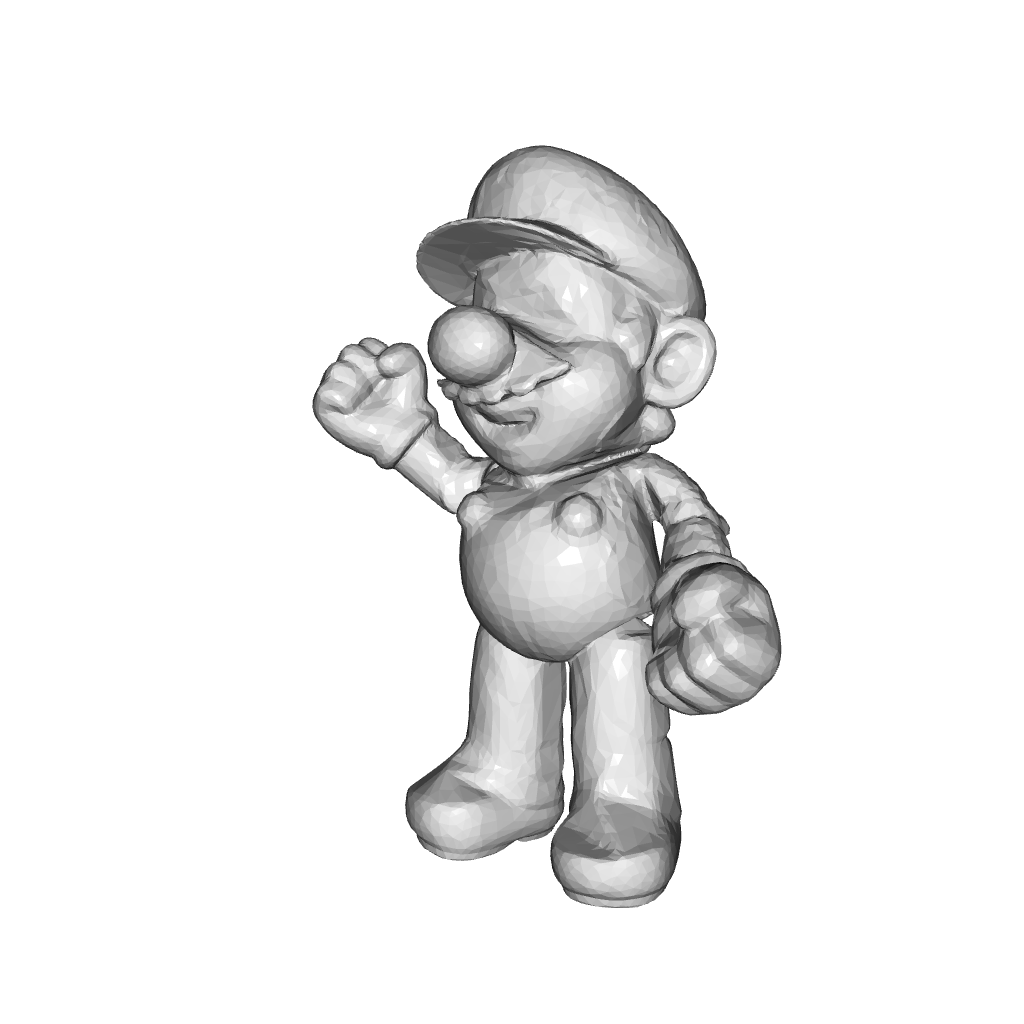}&
        \includegraphics[width = .1\textwidth, trim=0.cm 4cm 0cm 4cm,clip]{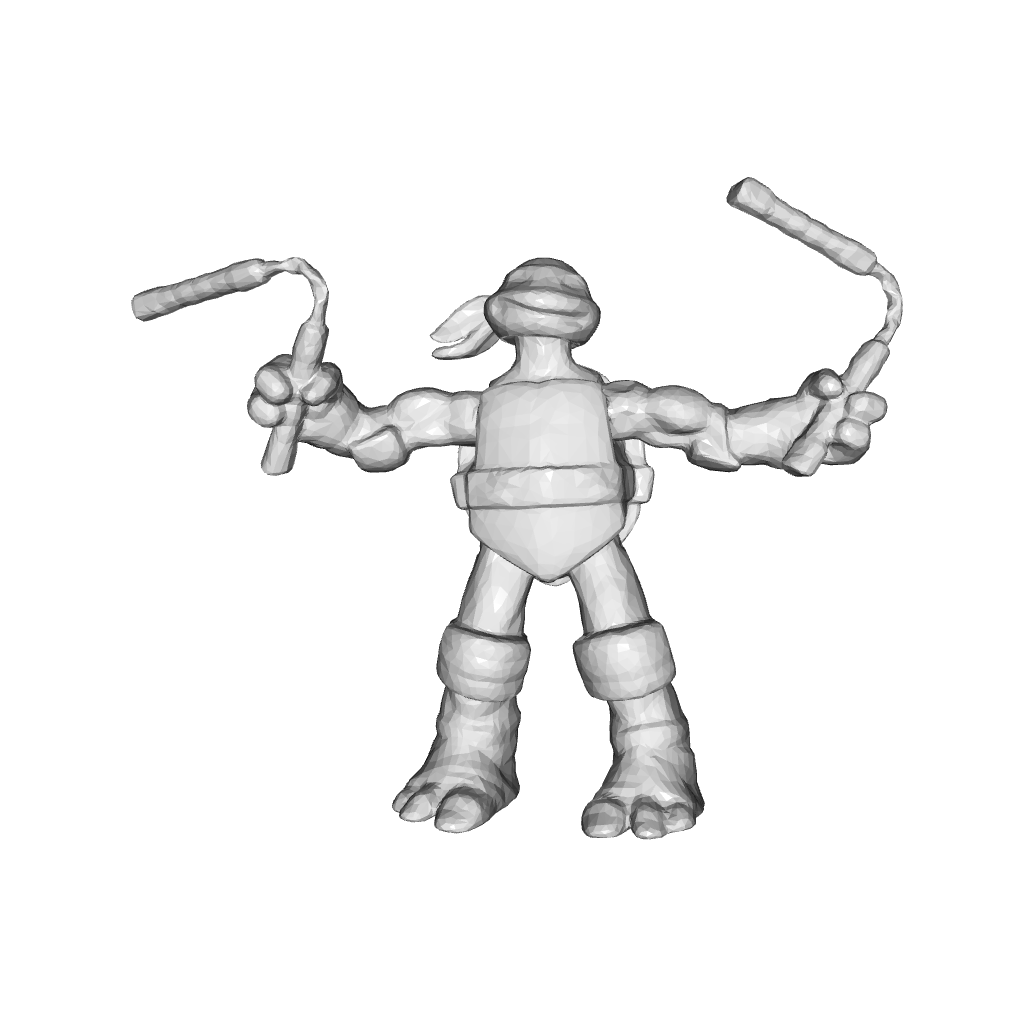}&
        \includegraphics[width = .1\textwidth, trim=0.cm 4cm 0cm 4cm,clip]{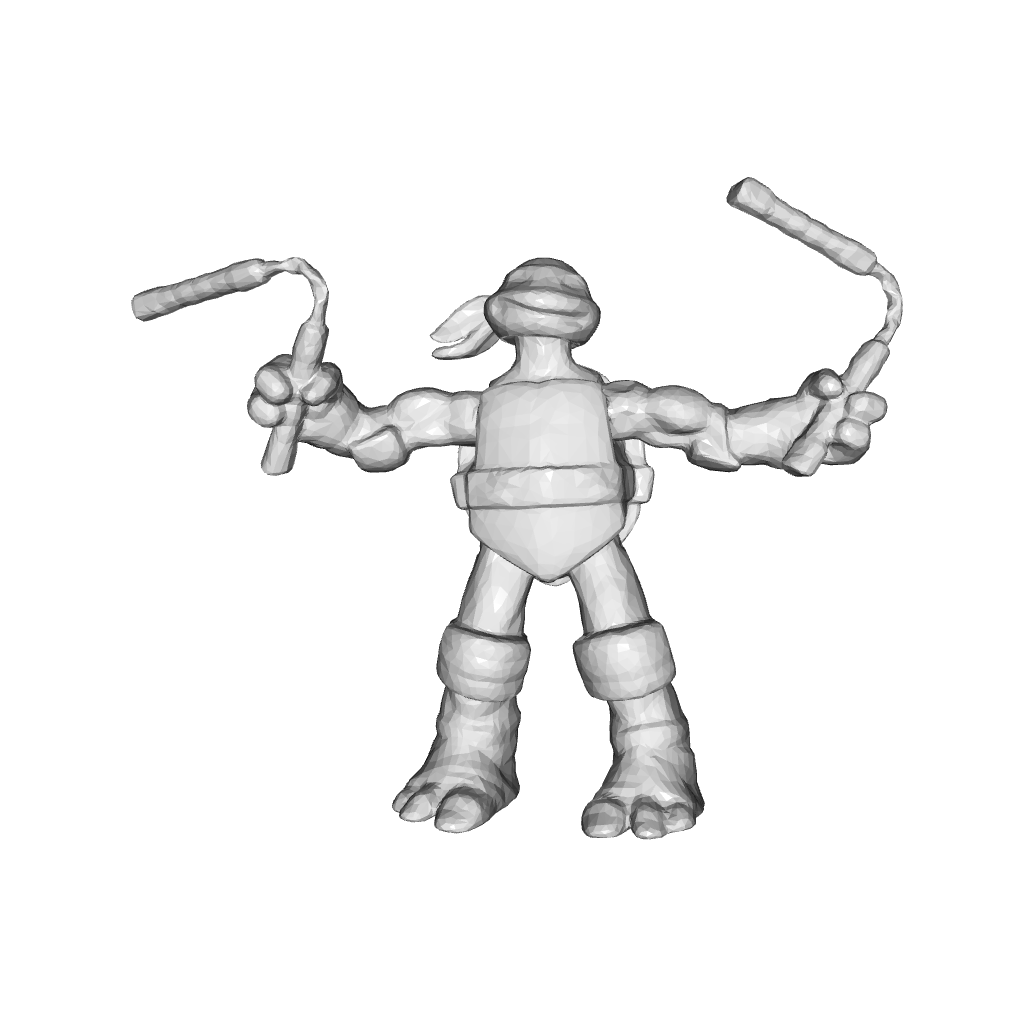} \\
		30 degree&90 degree  & 30 degree  &  90 degree

    \end{tabular}		
    \end{center}
	\caption{Reconstructions with narrow camera views. Input views (top), reconstructed meshes (middle) and GT meshes (bottom). \label{fig:narrow_setup}}
\end{figure}

\noindent\textbf{Narrow view reconstruction.} As described in the main paper, our method can also work in the narrow view configuration. In Figure~\ref{fig:narrow_setup} we present reconstructions given \textit{only two views} each with a relative pose difference of 30 degree and 90 degree. The algorithm successfully reconstructs complete shapes, albeit with noticeable surface deformations.
\renewcommand{\tabcolsep}{0.5pt}

\begin{figure*}[t]
	\begin{center}
	\begin{tabular}{ccccccc}
        \includegraphics[width = .14\linewidth,trim=0cm 0cm 0cm 0cm,clip]{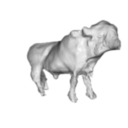}&
        \includegraphics[width = .14\linewidth,trim=0cm 0cm 0cm 0cm,clip]{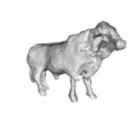}&
        \includegraphics[width = .14\linewidth,trim=0cm 0cm 0cm 0cm,clip]{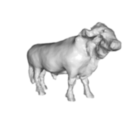}&
        \includegraphics[width = .14\linewidth,trim=0cm 0cm 0cm 0cm,clip]{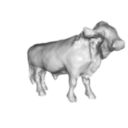}&
        \includegraphics[width = .14\linewidth,trim=0cm 0cm 0cm 0cm,clip]{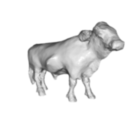}&
        \includegraphics[width = .14\linewidth,trim=0cm 0cm 0cm 0cm,clip]{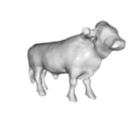}&
        \includegraphics[width = .14\linewidth,trim=0cm 0cm 0cm 0cm,clip]{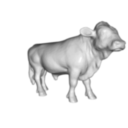}\\
        \includegraphics[width = .14\linewidth,trim=0cm 0cm 0cm 0cm,clip]{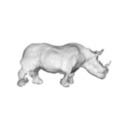}&
        \includegraphics[width = .14\linewidth,trim=0cm 0cm 0cm 0cm,clip]{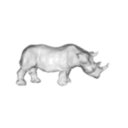}&
        \includegraphics[width = .14\linewidth,trim=0cm 0cm 0cm 0cm,clip]{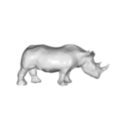}&
        \includegraphics[width = .14\linewidth,trim=0cm 0cm 0cm 0cm,clip]{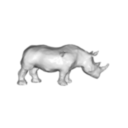}&
        \includegraphics[width = .14\linewidth,trim=0cm 0cm 0cm 0cm,clip]{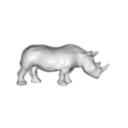}&
        \includegraphics[width = .14\linewidth,trim=0cm 0cm 0cm 0cm,clip]{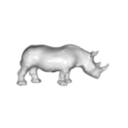}&
        \includegraphics[width = .14\linewidth,trim=0cm 0cm 0cm 0cm,clip]{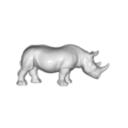}\\
        \includegraphics[width = .14\linewidth,trim=0cm 0cm 0cm 0cm,clip]{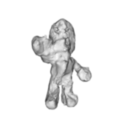}&
        \includegraphics[width = .14\linewidth,trim=0cm 0cm 0cm 0cm,clip]{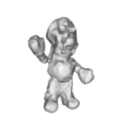}&
        \includegraphics[width = .14\linewidth,trim=0cm 0cm 0cm 0cm,clip]{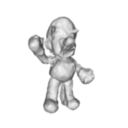}&
        \includegraphics[width = .14\linewidth,trim=0cm 0cm 0cm 0cm,clip]{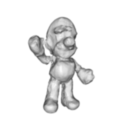}&
        \includegraphics[width = .14\linewidth,trim=0cm 0cm 0cm 0cm,clip]{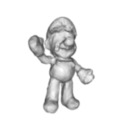}&
        \includegraphics[width = .14\linewidth,trim=0cm 0cm 0cm 0cm,clip]{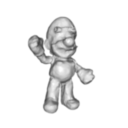}&
        \includegraphics[width = .14\linewidth,trim=0cm 0cm 0cm 0cm,clip]{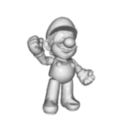}\\
        \includegraphics[width = .14\linewidth,trim=0cm 0cm 0cm 0cm,clip]{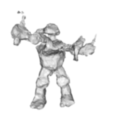}&
        \includegraphics[width = .14\linewidth,trim=0cm 0cm 0cm 0cm,clip]{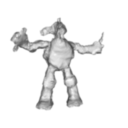}&
        \includegraphics[width = .14\linewidth,trim=0cm 0cm 0cm 0cm,clip]{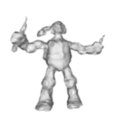}&
        \includegraphics[width = .14\linewidth,trim=0cm 0cm 0cm 0cm,clip]{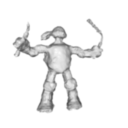}&
        \includegraphics[width = .14\linewidth,trim=0cm 0cm 0cm 0cm,clip]{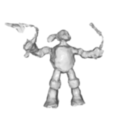}&
        \includegraphics[width = .14\linewidth,trim=0cm 0cm 0cm 0cm,clip]{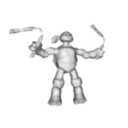}&
        \includegraphics[width = .14\linewidth,trim=0cm 0cm 0cm 0cm,clip]{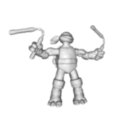}\\
        \includegraphics[width = .14\linewidth,trim=0cm 0cm 0cm 0cm,clip]{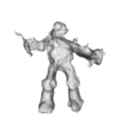}&
        \includegraphics[width = .14\linewidth,trim=0cm 0cm 0cm 0cm,clip]{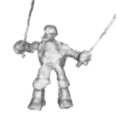}&
        \includegraphics[width = .14\linewidth,trim=0cm 0cm 0cm 0cm,clip]{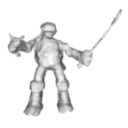}&
        \includegraphics[width = .14\linewidth,trim=0cm 0cm 0cm 0cm,clip]{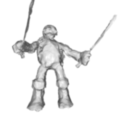}&
        \includegraphics[width = .14\linewidth,trim=0cm 0cm 0cm 0cm,clip]{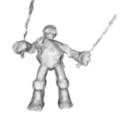}&
        \includegraphics[width = .14\linewidth,trim=0cm 0cm 0cm 0cm,clip]{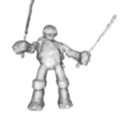}&
        \includegraphics[width = .14\linewidth,trim=0cm 0cm 0cm 0cm,clip]{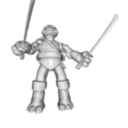}\\
        \includegraphics[width = .14\linewidth,trim=0cm 0cm 0cm 0cm,,clip]{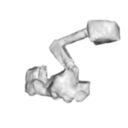}&
        \includegraphics[width = .14\linewidth,trim=0cm 0cm 0cm 0cm,,clip]{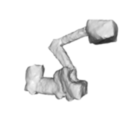}&
        \includegraphics[width = .14\linewidth,trim=0cm 0cm 0cm 0cm,,clip]{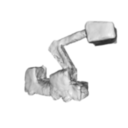}&
        \includegraphics[width = .14\linewidth,trim=0cm 0cm 0cm 0cm,,clip]{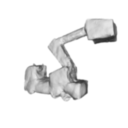}&
        \includegraphics[width = .14\linewidth,trim=0cm 0cm 0cm 0cm,,clip]{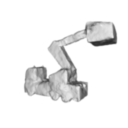}&
        \includegraphics[width = .14\linewidth,trim=0cm 0cm 0cm 0cm,,clip]{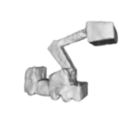}&
        \includegraphics[width = .14\linewidth,trim=0cm 0cm 0cm 0cm,clip]{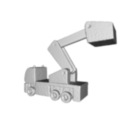}\\
        \includegraphics[width = .14\linewidth,trim=0cm 0cm 0cm 0cm,clip]{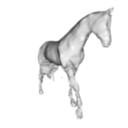}&
        \includegraphics[width = .14\linewidth,trim=0cm 0cm 0cm 0cm,clip]{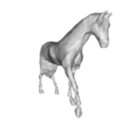}&
        \includegraphics[width = .14\linewidth,trim=0cm 0cm 0cm 0cm,clip]{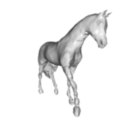}&
        \includegraphics[width = .14\linewidth,trim=0cm 0cm 0cm 0cm,clip]{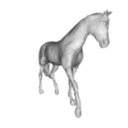}&
        \includegraphics[width = .14\linewidth,trim=0cm 0cm 0cm 0cm,clip]{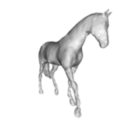}&
        \includegraphics[width = .14\linewidth,trim=0cm 0cm 0cm 0cm,clip]{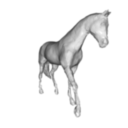}&
        \includegraphics[width = .14\linewidth,trim=0cm 0cm 0cm 0cm,clip]{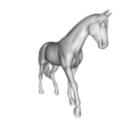}\\
        \includegraphics[width = .14\linewidth,trim=0cm 0cm 0cm 0cm,clip]{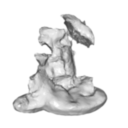}&
        \includegraphics[width = .14\linewidth,trim=0cm 0cm 0cm 0cm,clip]{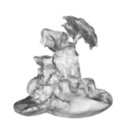}&
        \includegraphics[width = .14\linewidth,trim=0cm 0cm 0cm 0cm,clip]{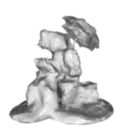}&
        \includegraphics[width = .14\linewidth,trim=0cm 0cm 0cm 0cm,clip]{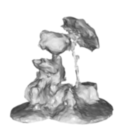}&
        \includegraphics[width = .14\linewidth,trim=0cm 0cm 0cm 0cm,clip]{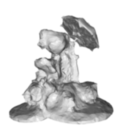}&
        \includegraphics[width = .14\linewidth,trim=0cm 0cm 0cm 0cm,clip]{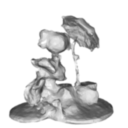}&
        \includegraphics[width = .14\linewidth,trim=0cm 0cm 0cm 0cm,clip]{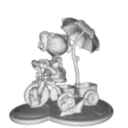}\\

        3 views &4 views & 5 views & 6 views & 7 views & 8 views & GT \\
    \end{tabular}		

    \end{center}
	\caption{Reconstructed object with different number of input views for objects in the GSO dataset.
	}
	\label{fig:mesh_vs_nbviews}
\end{figure*}

\noindent\textbf{Performance vs. the number of views.} In this experiment, we measure the performance of our algorithm for a different number of training views in the GSO dataset (see Table~\ref{tab:views} and Figure~\ref{fig:mesh_vs_nbviews}).
By observing the Figure~\ref{fig:mesh_vs_nbviews} we see that our algorithm can produce approximately correct shapes even with 5 camera views. Note that for 3 camera views, the coarse shape estimate is reasonable, but the shapes either miss some parts (\eg, the back leg of the horse, the sword of the ninja) or some parts are not distinguishable enough (\eg, face of Mario) due to the high ambiguity of potential valid 3D models.

\begin{table}[t]
\begin{center}
\caption{\label{demo-table}3D metrics on the GSO dataset (without background) for a different number of training views.  Note that the result with 6 views is better than other baselines  (see Table~1 in paper) and the result with 5 views is better than NerS [57]. *All scores have been multiplied by $10^{4}$.}
\label{tab:views}
\small
\begin{tabular}{@{}l c c c c c@{}}
\toprule

 \# of views   & CH-L2* $\downarrow$  &CH-L1 $\downarrow$  & Normal $\uparrow$ & F@10 $\uparrow$  \\  \hline

 3 &  59.88  & 0.090 & 0.53 & 93.89   \\
 4 &  22.04 & 0.063 & 0.61 & 98.26   \\
 5 &  16.09 & 0.053 & 0.67 & 98.85   \\
 6 & 12.55 & 0.048 & 0.68 & 99.17   \\
 7 &  11.88 & 0.048 & 0.68 & 99.26   \\
  8 & 8.69 & 0.034 & 0.75& 99.24    \\
 \hline
\end{tabular}
\end{center}
\end{table}

\begin{figure}[t]
	\begin{center}
	\begin{tabular}{cc}

		\includegraphics[width = .45\linewidth,trim=0cm 0.75cm 0cm 0cm,clip]{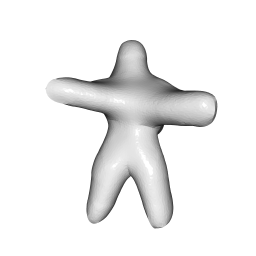} &
        \includegraphics[width = .45\linewidth,trim=0cm 1cm 0.25cm 1cm,clip]{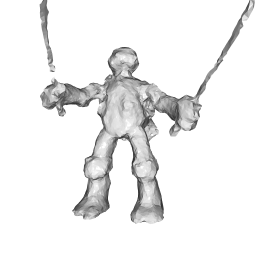} \\
        2 samples & 4 samples  \\
		\includegraphics[width = .45\linewidth,trim=0cm 1cm 0.25cm 1cm,clip]{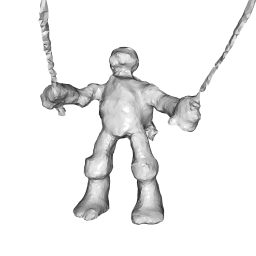} &
		\includegraphics[width = .45\linewidth, trim=0cm 1cm 0.25cm 1cm,clip]{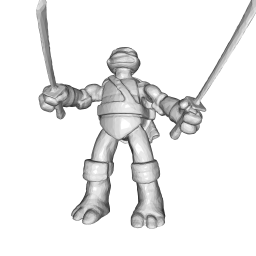}\\
         8 samples & GT  \\
    \end{tabular}		

    \end{center}
	\caption{Number of samples along the ray.
	}
	\label{fig:number_of_samples}
 
\end{figure}

\noindent\textbf{Number of samples along the ray.} In this ablation study, we explore the influence of the number of ray samples. To achieve this, we applied our algorithm to the \textit{Ninja} object (with background) using $2$, $4$, and $8$ samples along the ray. Figure~\ref{fig:number_of_samples} showcases the reconstructions. 
Notably, our algorithm manages to reconstruct almost all the shape even with only $4$ samples along the ray. 
Nonetheless, there are still noticeable artifacts, such as the division of the sword into two parts.

\begin{figure*}[t]
	\begin{tabular}{ccccccc}

        \includegraphics[width = .14\textwidth,trim=0.5cm 1cm 0.5cm 1.5cm,clip]{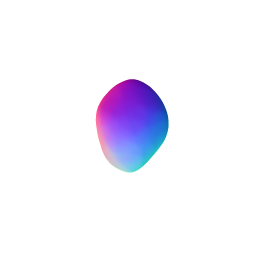}&
		\includegraphics[width = .14\textwidth,trim=0.5cm 1cm 0.5cm 1.5cm,clip]{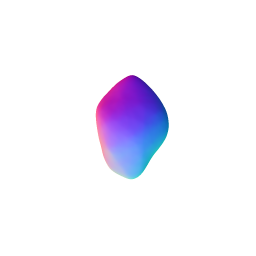} &
		\includegraphics[width = .14\textwidth, trim=0.5cm 1cm 0.5cm 1.5cm,clip]{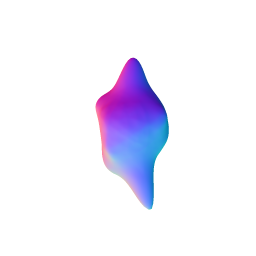}&
        \includegraphics[width = .14\textwidth, trim=0.5cm 1cm 0.5cm 1.5cm,clip]{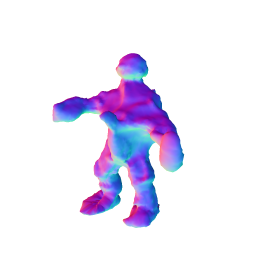}&
		\includegraphics[width = .14\textwidth, trim=0.5cm 1cm 0.5cm 1.5cm,clip]{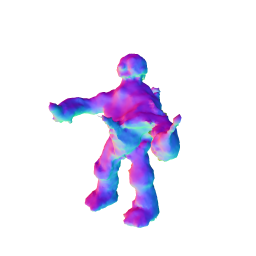}&
		\includegraphics[width = .14\textwidth,trim=0.5cm 1cm 0.5cm 1.5cm,clip]{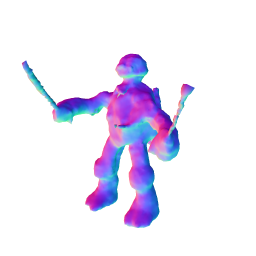} &
        \includegraphics[width = .14\textwidth, trim=0.5cm 1cm 0.5cm 1.5cm,clip]{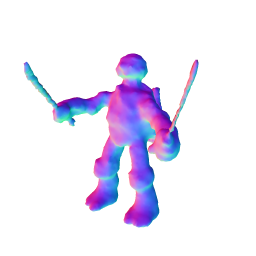}\\
        \includegraphics[width = .14\textwidth, trim=0.cm 1.5cm 0.5cm 2cm,clip]{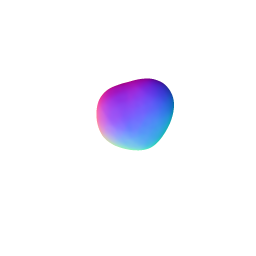}&
		\includegraphics[width = .14\textwidth, trim=0.cm 1.5cm 0.5cm 2cm,clip]{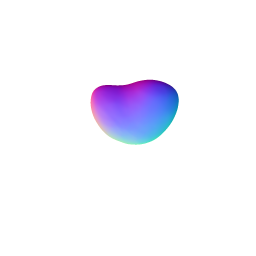} &
		\includegraphics[width = .14\textwidth, trim=0.cm 1.5cm 0.5cm 2cm,clip]{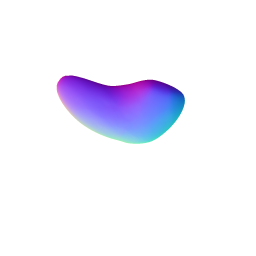}&
        \includegraphics[width = .14\textwidth, trim=0.cm 1.5cm 0.5cm 2cm,clip]{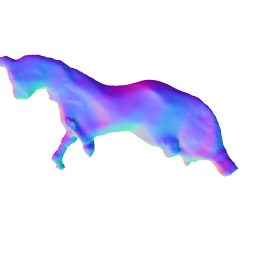}&
		\includegraphics[width = .14\textwidth, trim=0.cm 1.5cm 0.5cm 2cm,clip]{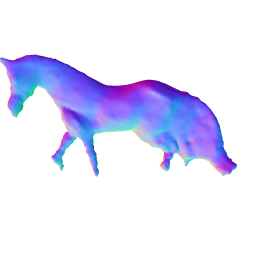}&
		\includegraphics[width = .14\textwidth,trim=0.cm 1.5cm 0.5cm 2cm,clip]{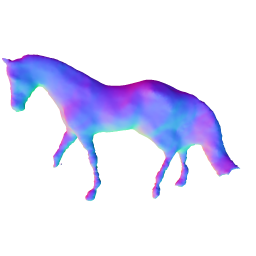} &
        \includegraphics[width = .14\textwidth, trim=0.cm 1.5cm 0.5cm 2cm,clip]{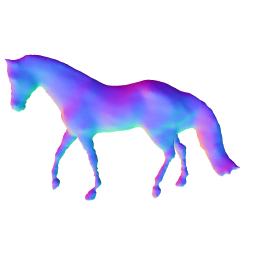} \\
        300 & 500 & 2000 & 500 & 1000 & 2000  & 5000

    \end{tabular}		

	\caption{Evolution of the shape over iteration time for coarse and detailed model training on the \textit{Ninja Turtles Leonardo} (top) and \textit{Breyer Horse} (bottom) from the GSO dataset with 8 views and with the background. We show the mesh normals in false colors. 
	}
	\label{fig:evalution_shape}
\end{figure*}

\noindent\textbf{Using a sphere as a coarse shape initialization.}
In Figure~\ref{fig:evalution_shape} we show the shape evolution during the training. 
As we mention in the paper, the coarse model reconstruction provides us with the robust initialization for the detailed model stage. 
However, one can start directly from the sphere mesh by skipping the first stage and still obtain decent shapes. 
To verify this, we start directly from the detailed model stage by skipping the first stage in the GSO dataset. 
In Table~\ref{table:from_sphere} we present our quantitative results. Although the performance dropped by a little amount the performance is still satisfactory.

\begin{figure}[t]
	\begin{center}
	\begin{tabular}{ccccc}

        \includegraphics[width = .2\linewidth,trim=0cm 0cm 0cm 0cm,clip]{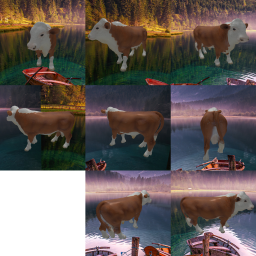}&
        \includegraphics[width = .2\linewidth,trim=0cm 0cm 0cm 0cm,clip]{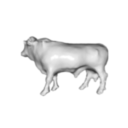}&
		\includegraphics[width = .2\linewidth,trim=0cm 0cm 0cm 0cm,clip]{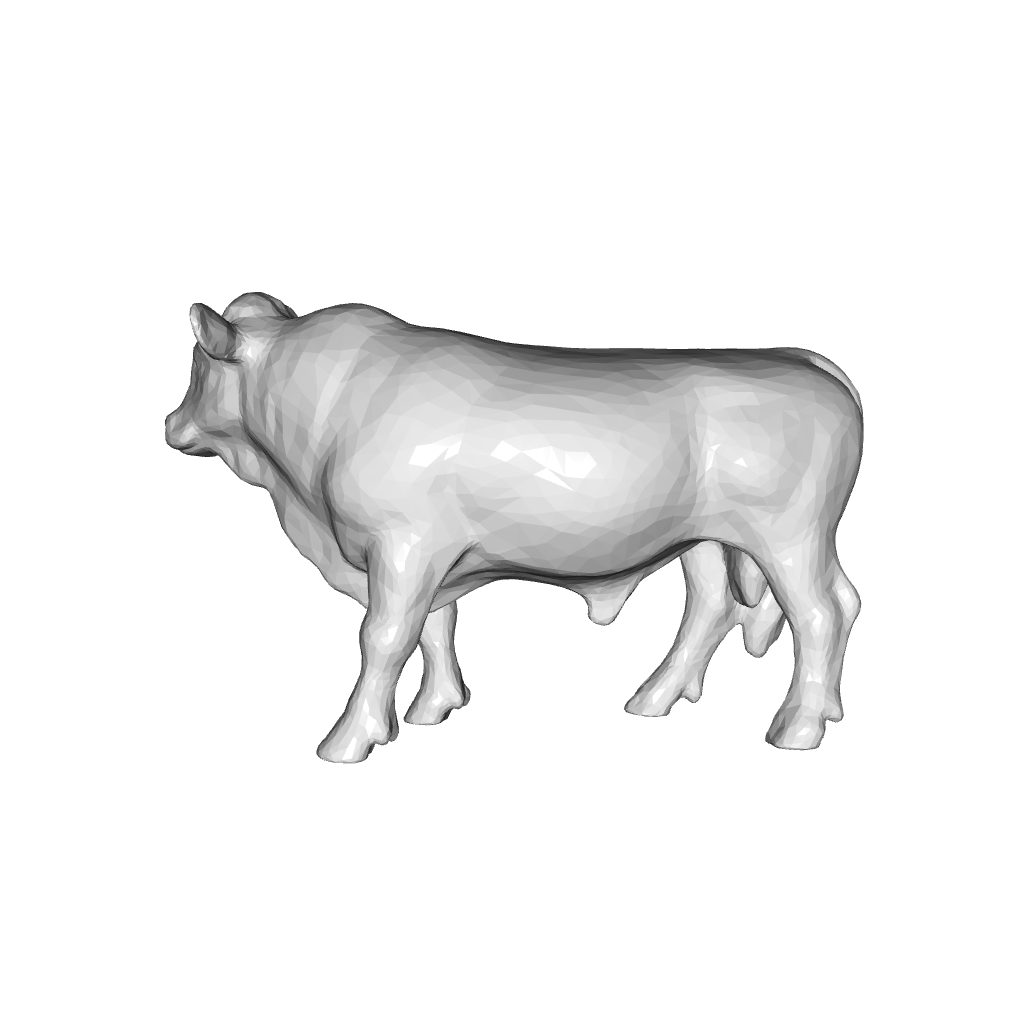} &
        \includegraphics[width = .2\linewidth,trim=0cm 0cm 0cm 0cm,clip]{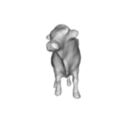}&
		\includegraphics[width = .2\linewidth,trim=0cm 0cm 0cm 0cm,clip]{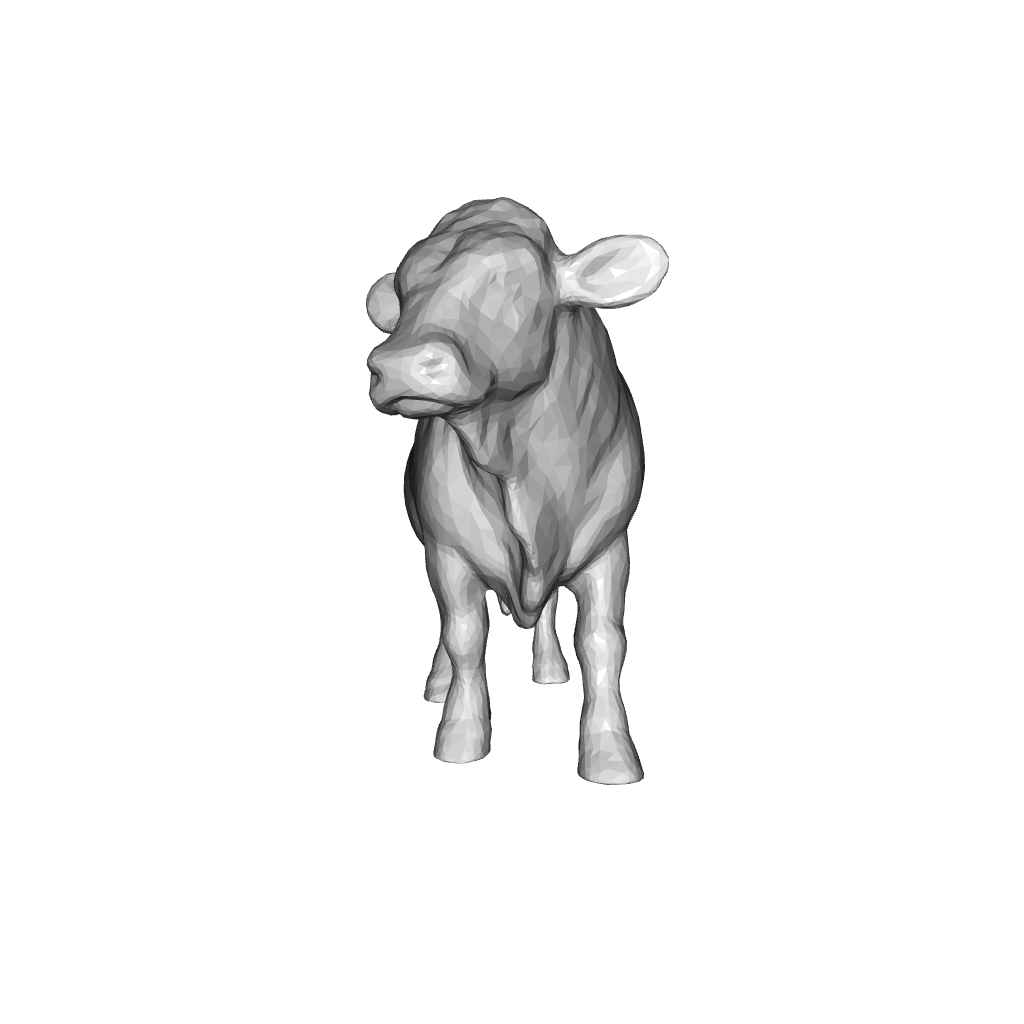} 
        \\
        \includegraphics[width = .2\linewidth,trim=0cm 0cm 0cm 0cm,clip]{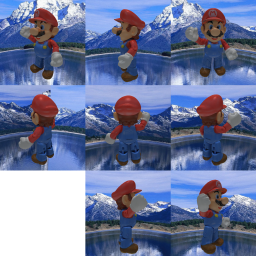}&
        \includegraphics[width = .2\linewidth,trim=0cm 0cm 0cm 0cm,clip]{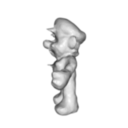}&
		\includegraphics[width = .2\linewidth,trim=0cm 0cm 0cm 0cm,clip]{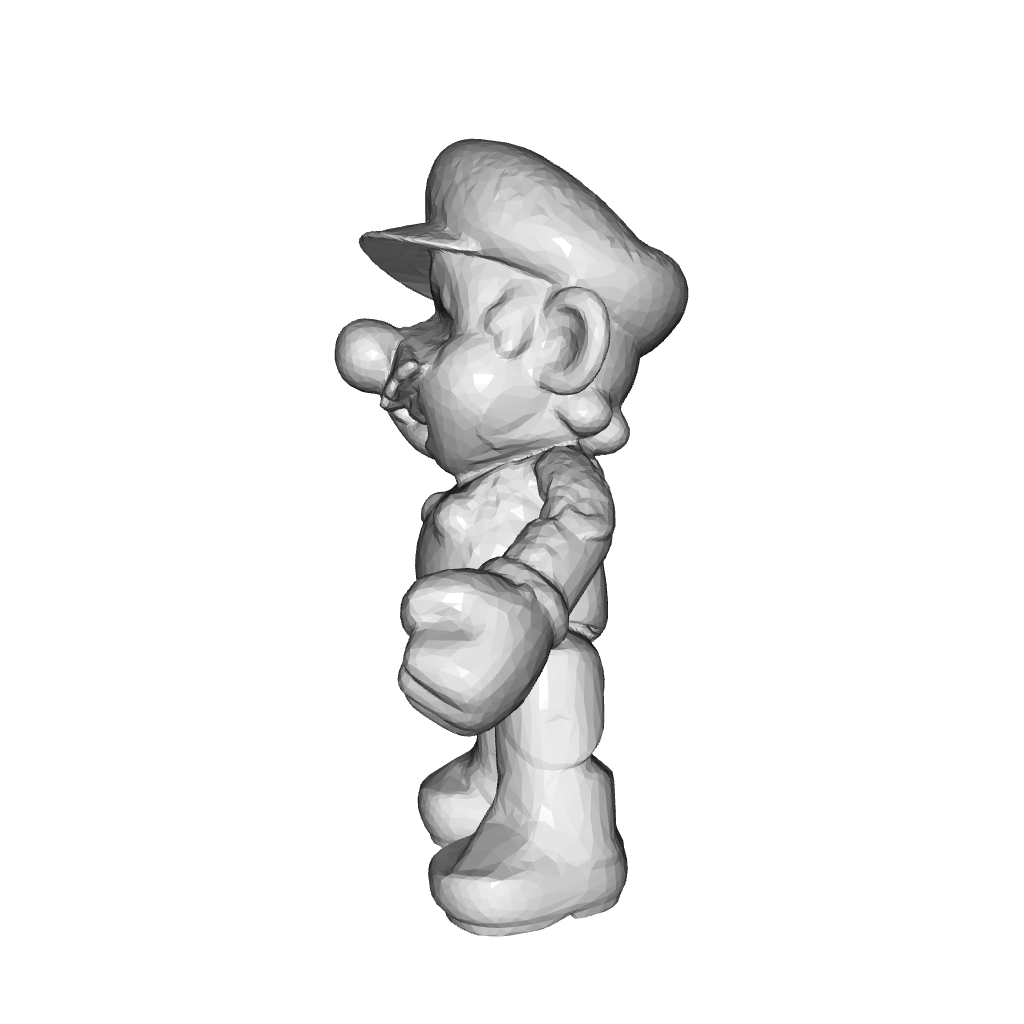} &
        \includegraphics[width = .2\linewidth,trim=0cm 0cm 0cm 0cm,clip]{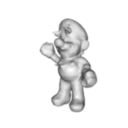}&
		\includegraphics[width = .2\linewidth,trim=0cm 0cm 0cm 0cm,clip]{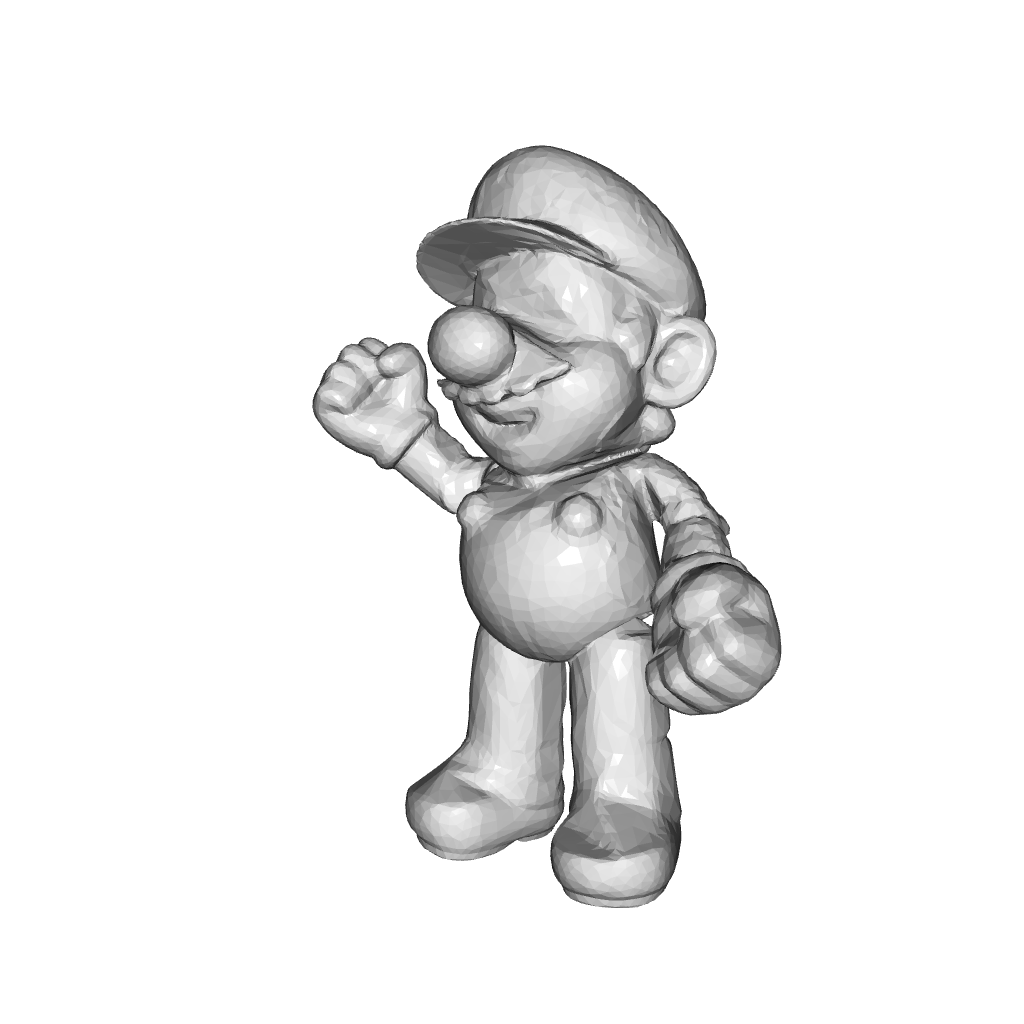} 
        \\
        \includegraphics[width = .2\linewidth,trim=0cm 0cm 0cm 0cm,clip]{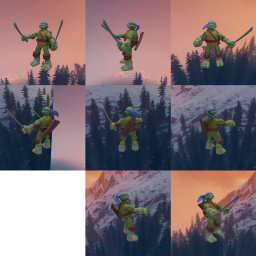}&
        \includegraphics[width = .2\linewidth,trim=0cm 0cm 0cm 0cm,clip]{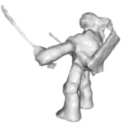}&
		\includegraphics[width = .2\linewidth,trim=0cm 0cm 0cm 0cm,clip]{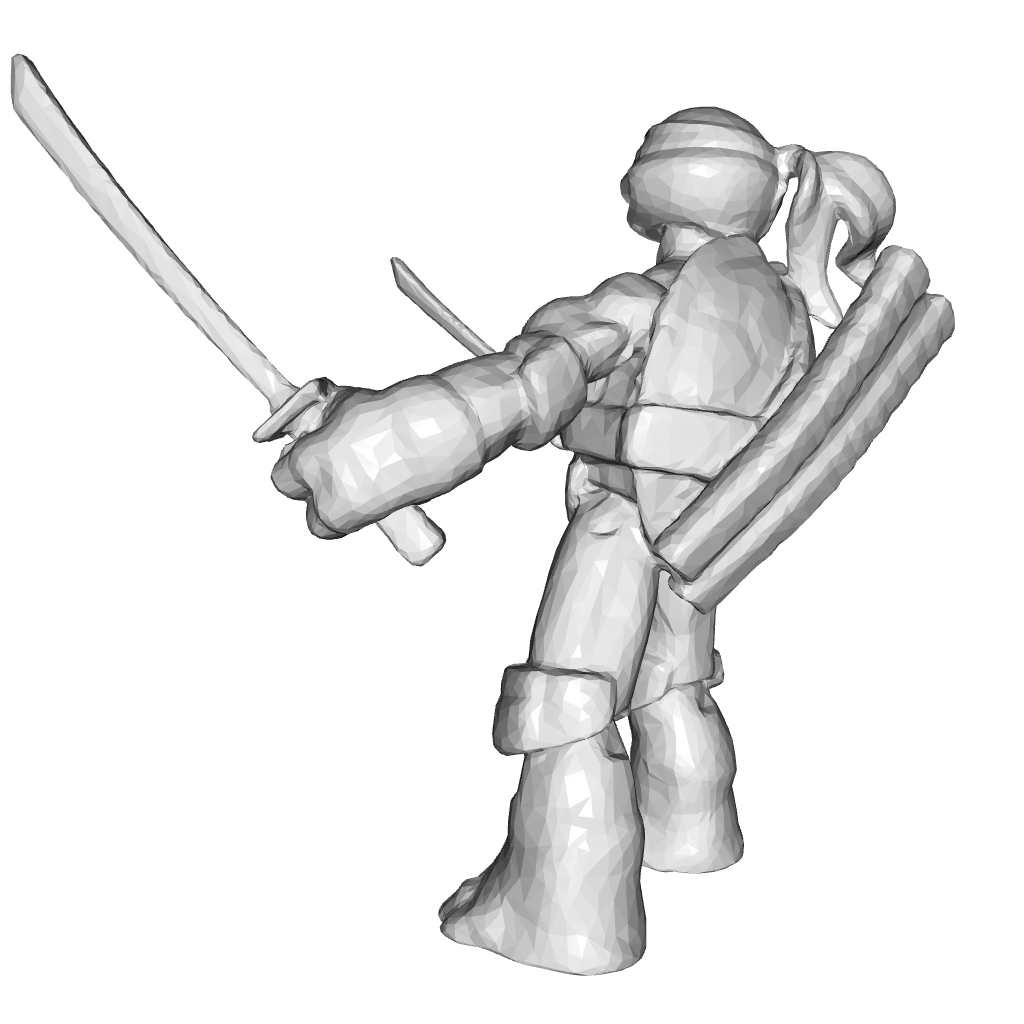} &
        \includegraphics[width = .2\linewidth,trim=0cm 0cm 0cm 0cm,clip]{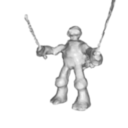}&
		\includegraphics[width = .2\linewidth,trim=0cm 0cm 0cm 0cm,clip]{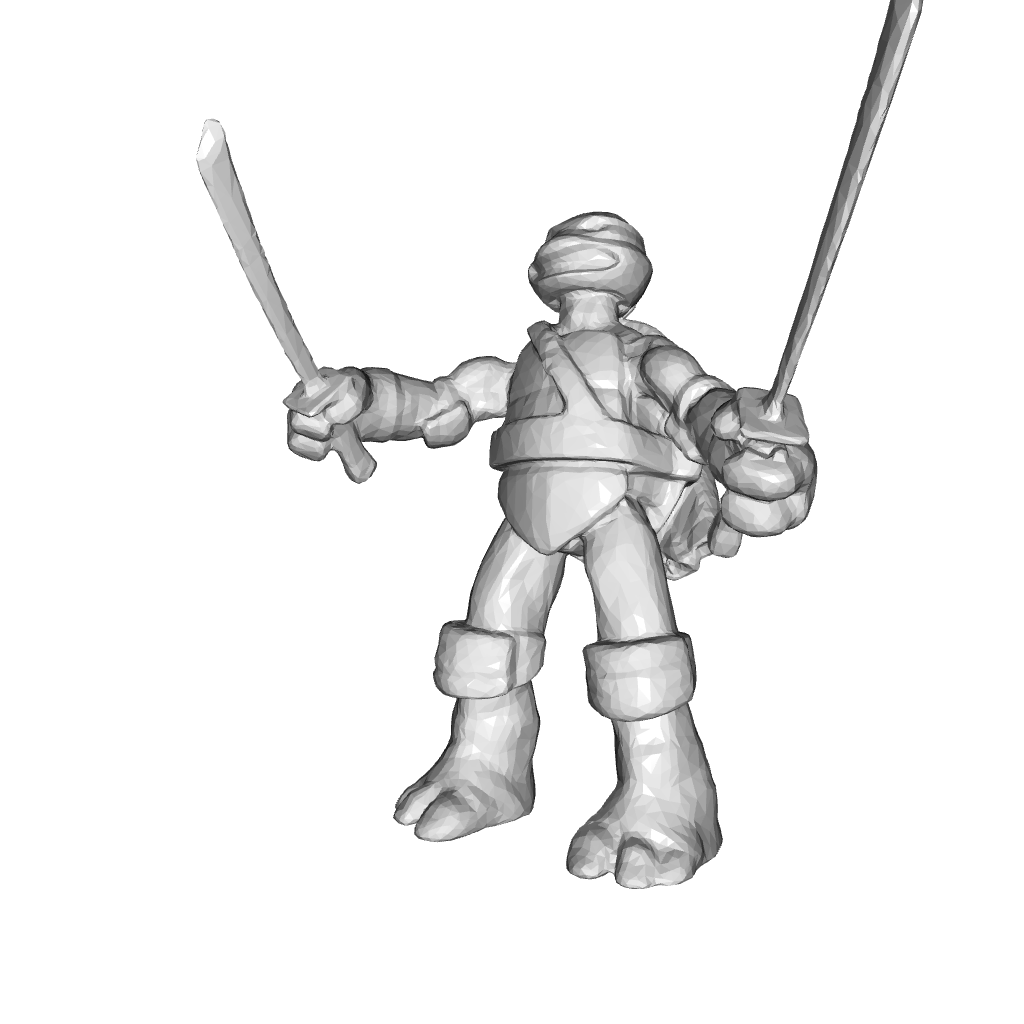} 
        \\
        \includegraphics[width = .2\linewidth,trim=0cm 0cm 0cm 0cm,clip]{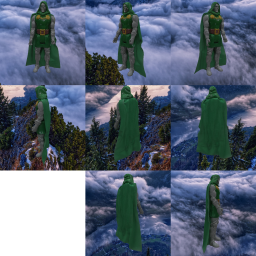}&
        \includegraphics[width = .2\linewidth,trim=0cm 0cm 0cm 0cm,clip]{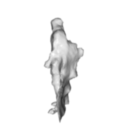}&
		\includegraphics[width = .2\linewidth,trim=0cm 0cm 0cm 0cm,clip]{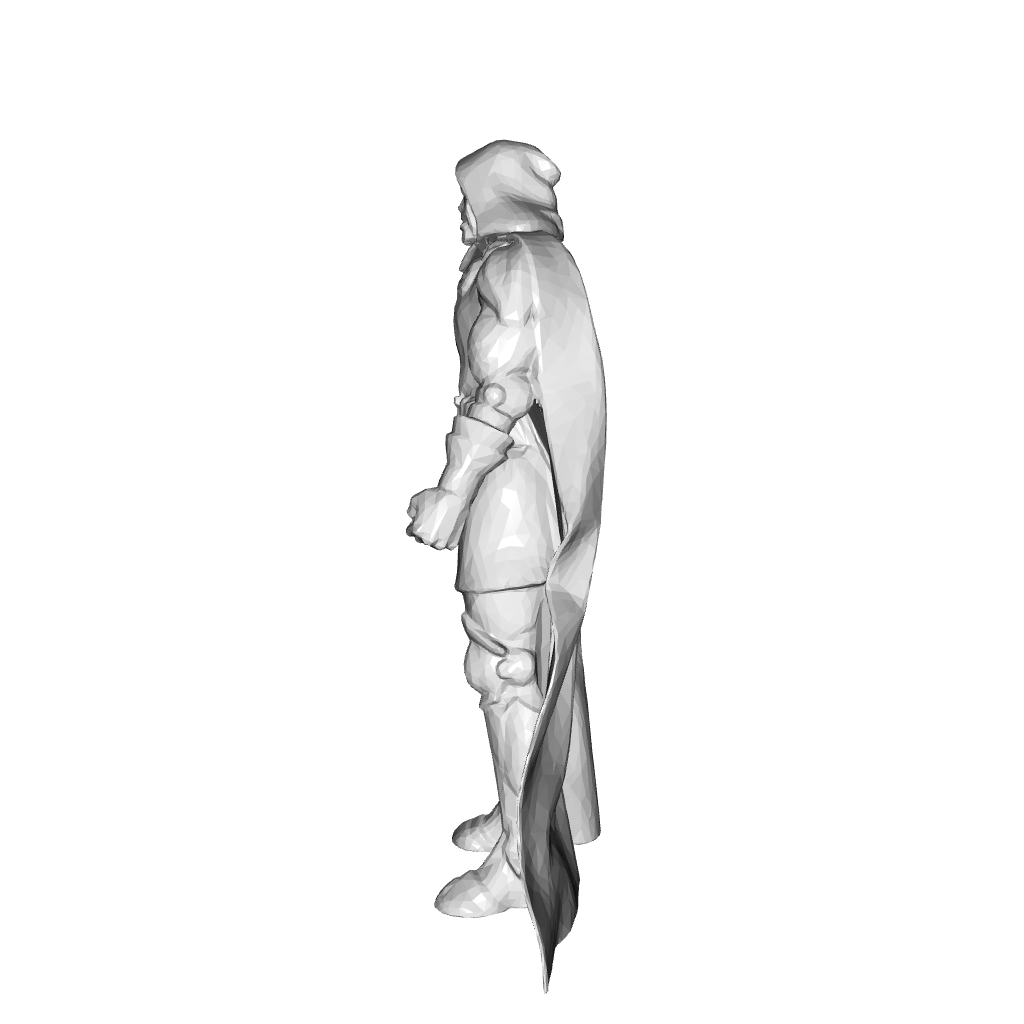} &
        \includegraphics[width = .2\linewidth,trim=0cm 0cm 0cm 0cm,clip]{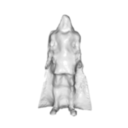}&
		\includegraphics[width = .2\linewidth,trim=0cm 0cm 0cm 0cm,clip]{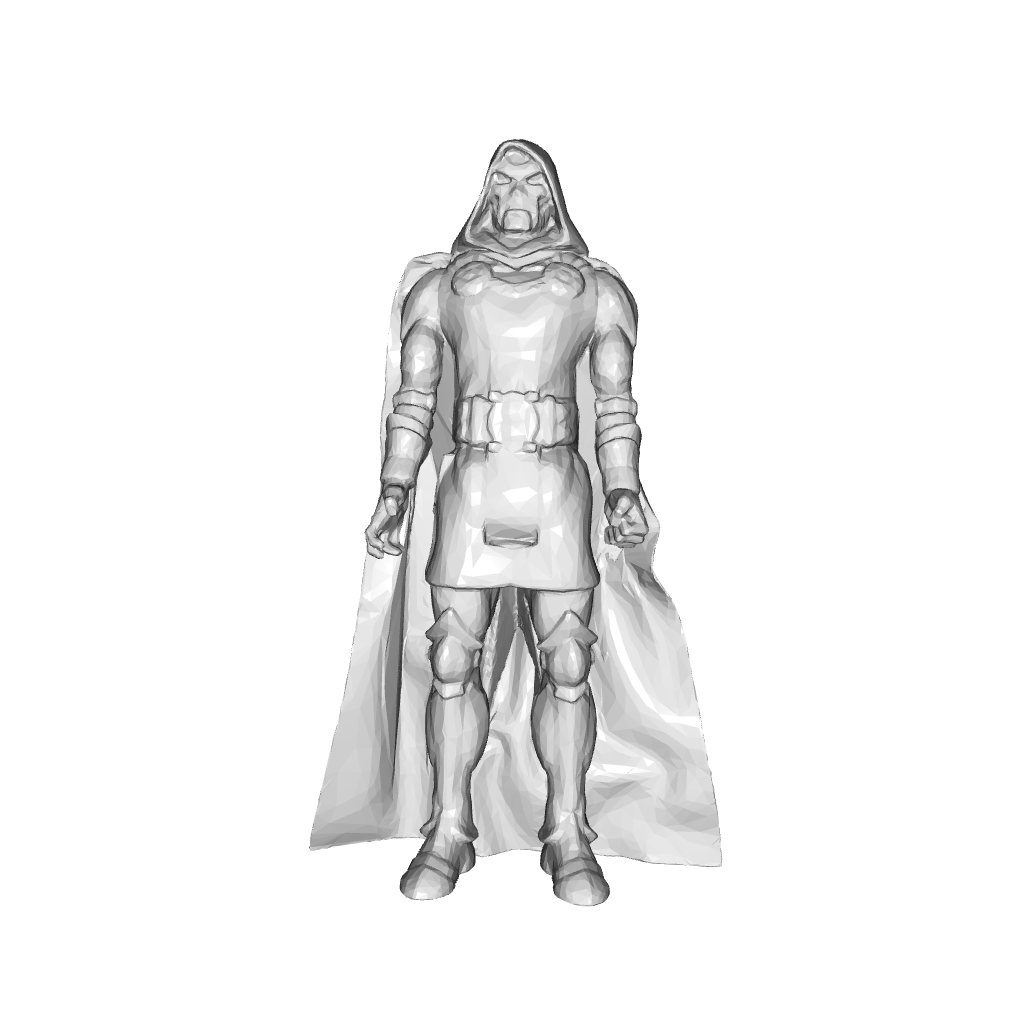} 
        \\
        \scriptsize{Input Images} &\scriptsize{Our - view  1} & \scriptsize{GT - view  1} & \scriptsize{Our - view  2} & \scriptsize{GT - view  2}
    \end{tabular}		

    \end{center}
	\caption{Reconstructed meshes of some objects in the GSO \cite{google_dataset} dataset by starting from a sphere (\ie, we skip the Coarse Model Reconstruction stage). Recovered meshes are shown from two different views (columns 1 and 3) with their corresponding GT view on their right hand side (columns 2 and 4). The reconstructions in the first and the third rows are satisfactory. However, the reconstructions on the second and the fourth rows have some artifacts, \eg, spiky surfaces, which do not appear with the coarse model initialization. 
	}
	\label{fig:Reconstruction_From_Sphere}
\end{figure}

Figure~\ref{fig:Reconstruction_From_Sphere} shows reconstructed meshes, where the initial template mesh was a sphere.
We observe that the detailed model representation is strong enough to recover the correct shape even from a basic initial shape (a sphere in this case).
However, reconstructions with such initialization show some artifacts. 
More tuning and longer training might help to remove these artifacts. 
To verify this assumption, we run the algorithm for more iterations and observe that the performance noticeably improves. 
On the other hand, starting from the coarse shape that our method produces leads to a more stable training and a faster convergence.

\begin{table}[t]
\begin{center}
\caption{3D metrics on the GSO dataset where we skip the first stage.  *All scores have been multiplied by $10^{4}$. **We run for more iterations, \ie, 15K in total. } 
\label{table:from_sphere}
\footnotesize
\begin{tabular}{@{}l@{}  c c c c@{}}
\toprule
 Method  & CH-L2* $\downarrow$  &CH-L1 $\downarrow$  & Normal $\uparrow$ & F@10 $\uparrow$  \\  \hline
 Our w\slash BCG & 13.75 & 0.041 & 0.75 & 98.79 \\
 Our w\slash BCG** & 11.44 & 0.040 & 0.76 & 98.99 \\
 Our w\slash BCG-Full  & 11.08 &  0.038& 0.75 & 98.85 \\
 \hline
\end{tabular}
\end{center}
\end{table}

\renewcommand{\tabcolsep}{0pt}

\begin{figure}[t]
	\begin{center}
	\begin{tabular}{cccc}

		\includegraphics[width = .24\linewidth,trim=0cm 0cm 0cm 0cm,clip]{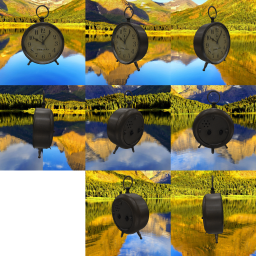} &
        \includegraphics[width = .24\linewidth,trim=0cm 0cm 0cm 0cm,clip]{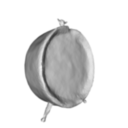}&
		\includegraphics[width = .24\linewidth,trim=3cm 2cm 3cm 2cm,clip]{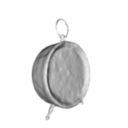} &
		\includegraphics[width = .24\linewidth, trim=3cm 2cm 3cm 2cm,clip]{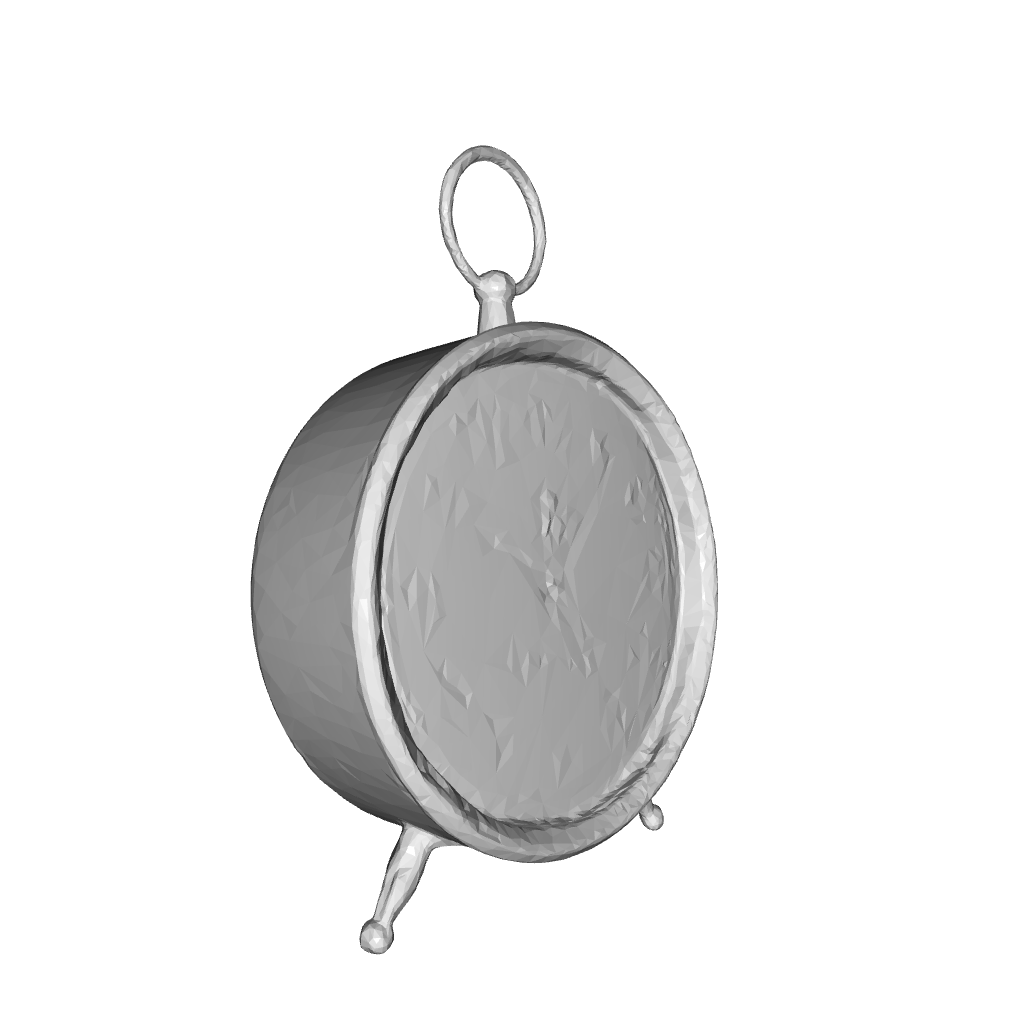}\\

         \scriptsize{Input Images} & \scriptsize{10K} & \scriptsize{20K} & \scriptsize{GT}
    \end{tabular}		

    \end{center}
	\caption{We show our reconstructions using 10K and 20K vertices for the watch object.  
	Using 10K vertices is insufficient to recover the circle on top of the watch. Increasing the mesh resolution, e.g., to 20K vertices, can fix this easily.}
	\label{fig:mesh_resolution}
\end{figure}

\noindent\textbf{The role of the mesh resolution.} In Figure~\ref{fig:mesh_resolution} we show the effect of the mesh resolution on the reconstruction of objects with thin parts. Recovering these parts requires a higher mesh resolution, \ie, more triangles. Note that for this case the training is longer.

\begin{figure}[t]
	\begin{center}
	\begin{tabular}{cccc}

		\includegraphics[width = .2\linewidth,trim=0cm 0cm 0cm 0cm,clip]{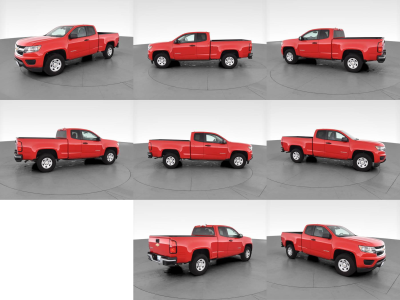} &
        \includegraphics[width = .2\linewidth,trim=0cm 0cm 0cm 0cm,clip]{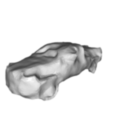}&
		\includegraphics[width = .2\linewidth,trim=0cm 0cm 0cm 0cm,clip]{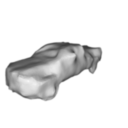} &
		\includegraphics[width = .2\linewidth, trim=0cm 0cm 0cm 0cm,clip]{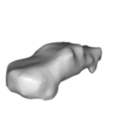}\\
        &
        \includegraphics[width = .2\linewidth,trim=0cm 0cm 0cm 0cm,clip]{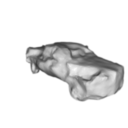}&
        \includegraphics[width = .2\linewidth,trim=0cm 0cm 0cm 0cm,clip]{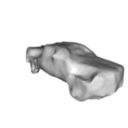} &
        \includegraphics[width = .2\linewidth, trim=0cm 0cm 0cm 0cm,clip]{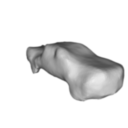}\\
         Input Images & $\lambda = 2$ & $\lambda = 5$ & $\lambda = 20$
    \end{tabular}		

    \end{center}
	\caption{Reconstructed cars for different values $\lambda$ of the Laplacian regularization. The two rows show the mesh of the same object from two different views.
	}
	\label{fig:laplacian_car}
 
\end{figure}

\noindent\textbf{The role of the Laplacian.}
In Figure~\ref{fig:laplacian_car}, we show the effect of the Laplacian regularizer on an object from the MVMC Car \cite{NERS_2021_Neurips} dataset.
To recover a smooth shape of objects with non-Lambertian surfaces, one needs to use more regularization.
Note that for the lower regularization the $F_\text{shape}$ network overfits and generates spiky surfaces around the windows of the car.

\renewcommand{\tabcolsep}{0pt}

\renewcommand{\tabcolsep}{0pt}

\renewcommand{\tabcolsep}{0pt}

\renewcommand{\tabcolsep}{0pt}

\begin{figure}[t]
	\begin{center}
	\begin{tabular}{cccc}

		\includegraphics[width = .25\linewidth,trim=0cm 0cm 0cm 0cm,clip]{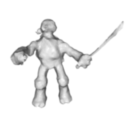} &
        \includegraphics[width = .25\linewidth,trim=0cm 0cm 0cm 0cm,clip]{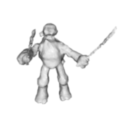}&
		\includegraphics[width = .25\linewidth,trim=0cm 0cm 0cm 0cm,clip]{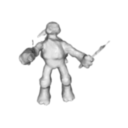} &
		\includegraphics[width = .25\linewidth, trim=0cm 0cm 0cm 0cm,clip]{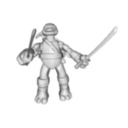}\\
		\includegraphics[width = .25\linewidth,trim=0cm 0cm 0cm 0cm,clip]{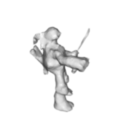} &
        \includegraphics[width = .25\linewidth,trim=0cm 0cm 0cm 0cm,clip]{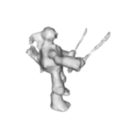}&
		\includegraphics[width = .25\linewidth,trim=0cm 0cm 0cm 0cm,clip]{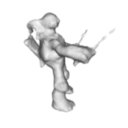} &
		\includegraphics[width = .25\linewidth, trim=0cm 0cm 0cm 1cm,clip]{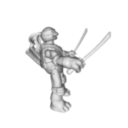}\\

         \scriptsize{First Order} & \scriptsize{Second Order} & \scriptsize{Third Order} & \scriptsize{GT}
    \end{tabular}		

    \end{center}
	\caption{Reconstructed meshes for the \textit{ninja} object using averaging of the face normals in the first, second and third order neighbourhood around a vertex.}
	\label{fig:vertex_normal}
\end{figure}

\noindent\textbf{Calculation of the surface normal.}
In the paper, we calculate the normal to the vertex $V_i$ in the detailed model representation by averaging the normals of the faces within the second order neighborhood around $V_i$.
In Figure~\ref{fig:vertex_normal}, we show  reconstructions obtained by calculating the normal to a vertex by averaging the normals of the faces within the \textit{first, second} and \textit{third} order neighbourhoods around $V_i$. 
We see that using the third order neighbourhood to calculate the normals does not allow the optimization to fully recover thinner parts of the shape. 
This is expected since the normal field is smoother in this case (there is more averaging).
One can handle this by increasing the mesh resolution. 
A first order neighbourhood does not allow to recover the thinner parts of the original shape too. On the other hand, the second order neighbourhood allows to  recover thin object parts.

\renewcommand{\tabcolsep}{0pt}

\section{Implementation Details}

For remeshing we use the pseudocode~\ref{algo:remesh} described here.\footnote{\url{https://github.com/PyMesh/PyMesh/blob/384ba882b7558ba6e8653ed263c419226c22bddf/scripts/fix_mesh.py\#L14}} 
The algorithm receives the mesh and the target edge length as input. 
It refines/fixes the mesh by applying the following sequence of operations: 1) removing degenerated triangles, 2) splitting long edges, 3) collapsing short edges, 4) removing obtuse triangles, and 5) computing the outer hull of the mesh. 
Notice that computing the outer hull allows us to change the topology of the mesh.
In our experiments, our meshes have around 10K vertices. 
In order to match this resolution, we start from the initial target edge length and call the pseudocode. 
If the new mesh has more than 10K vertices we increase the target edge length threshold and run it until we have the mesh that has around 10K vertices. 
For the objects in the GSO dataset, we observe that the remeshing and the classification of inside/outside of the 3D points take around $15\%$ of the total time in the detailed model reconstruction stage. 
The method takes about 2 hours for training on NVIDIA RTX A5000. Specifically, the coarse shape reconstruction accounts for roughly $15\%$ of the total time, while the fine reconstruction takes up about $35\%$, and the remaining $50\%$ is for texture refinement.
\section{Additional Results}

\begin{figure*}[t]
	\begin{center}
	\begin{tabular}{cccccccc}

        \includegraphics[width = .10\textwidth, trim=0.cm 0cm 0cm 0cm,clip]{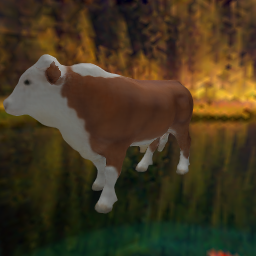}&
        \includegraphics[width = .10\textwidth, trim=0.cm 0cm 0cm 0cm,clip]{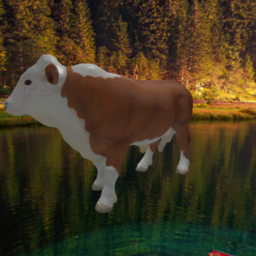} &
        \includegraphics[width = .10\textwidth, trim=0.cm 0cm 0cm 0cm,clip]{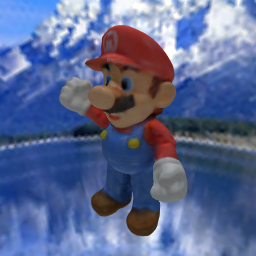}&
        \includegraphics[width = .10\textwidth, trim=0.cm 0cm 0cm 0cm,clip] {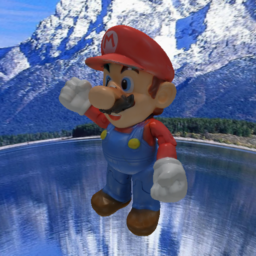} &
        \includegraphics[width = .10\textwidth, trim=0.cm 0cm 0cm 0cm,clip]{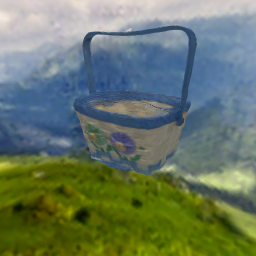}&
        \includegraphics[width = .10\textwidth, trim=0.cm 0cm 0cm 0cm,clip] {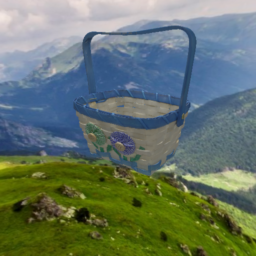}  &
        \includegraphics[width = .10\textwidth, trim=0.cm 0cm 0cm 0cm,clip]{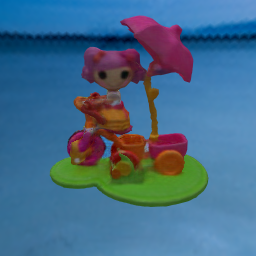}&
        \includegraphics[width = .10\textwidth, trim=0.cm 0cm 0cm 0cm,clip] {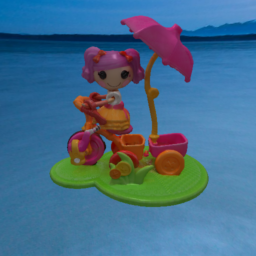} \\
        \includegraphics[width = .10\textwidth, trim=0.cm 0cm 0cm 0cm,clip]{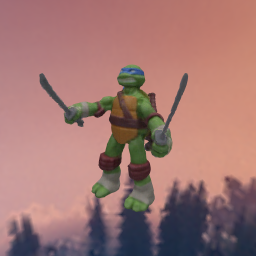}&
        \includegraphics[width = .10\textwidth, trim=0.cm 0cm 0cm 0cm,clip]{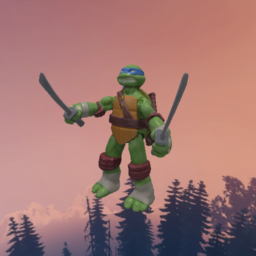} &
        \includegraphics[width = .10\textwidth, trim=0.cm 0cm 0cm 0cm,clip]{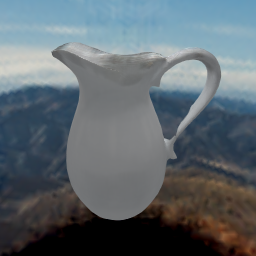}&
        \includegraphics[width = .10\textwidth, trim=0.cm 0cm 0cm 0cm,clip] {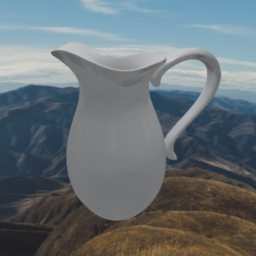} &
        \includegraphics[width = .10\textwidth, trim=0.cm 0cm 0cm 0cm,clip]{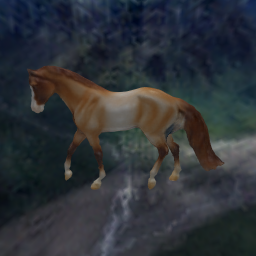}&
        \includegraphics[width = .10\textwidth, trim=0.cm 0cm 0cm 0cm,clip] {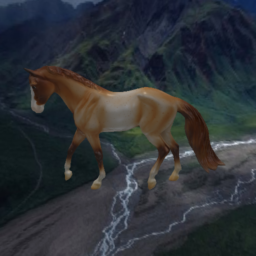}  &
        \includegraphics[width = .10\textwidth, trim=0.cm 0cm 0cm 0cm,clip]{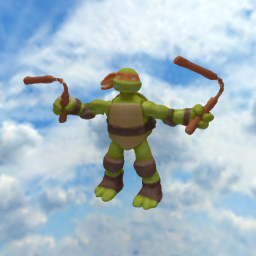}&
        \includegraphics[width = .10\textwidth, trim=0.cm 0cm 0cm 0cm,clip] {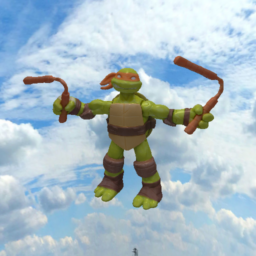} \\
		our w/BCG & GT & our w/BCG & GT & our w/BCG & GT & our w/BCG & GT

    \end{tabular}		

    \end{center}
	\caption{Additional qualitative results on the GSO Dataset. \label{fig:qualitative_results_google_texture}}
\end{figure*}

\begin{figure*}[t]
	\begin{center}
	\begin{tabular}{ccccccc}

        \includegraphics[width = .14\linewidth,trim=0cm 0cm 0cm 0cm,clip]{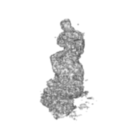} &
        \includegraphics[width = .14\linewidth,trim=0cm 0cm 0cm 0cm,clip]{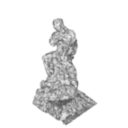}&
        \includegraphics[width = .14\linewidth,trim=0cm 0cm 0cm 0cm,clip]{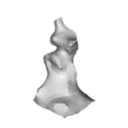}&
        \includegraphics[width = .14\linewidth,trim=0cm 0cm 0cm 0cm,clip]{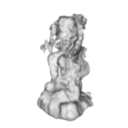} &
        \includegraphics[width = .14\linewidth,trim=0cm 0cm 0cm 0cm,clip]{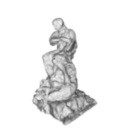} &
		\includegraphics[width = .14\linewidth,trim=0cm 0cm 0cm 0cm,clip]{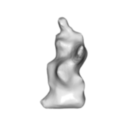} &
		\includegraphics[width = .14\linewidth, trim=0cm 0cm 0cm 0cm,clip]{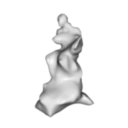}\\
        \includegraphics[width = .14\linewidth,trim=0cm 0cm 0cm 0cm,clip]{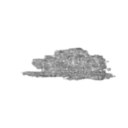} &
        \includegraphics[width = .14\linewidth,trim=0cm 0cm 0cm 0cm,clip]{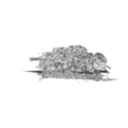}&
        \includegraphics[width = .14\linewidth,trim=0cm 0cm 0cm 0cm,clip]{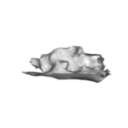}& 
        \includegraphics[width = .14\linewidth,trim=0cm 0cm 0cm 0cm,clip]{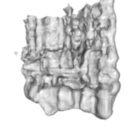} &
        \includegraphics[width = .14\linewidth,trim=0cm 0cm 0cm 0cm,clip]{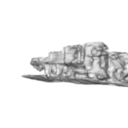} &
		\includegraphics[width = .14\linewidth, trim=0cm 0cm 0cm 0cm,clip]{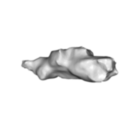} &
  		\includegraphics[width = .14\linewidth,trim=0cm 0cm 0cm 0cm,clip]{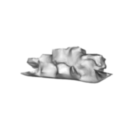} \\
         \scriptsize{RegNeRF} & \scriptsize{Munkberg et al. \cite{munkberg2021nvdiffrec}} & \scriptsize{DS} & \scriptsize{NeuS w/BCG}& \scriptsize{NeuS wo/BCG} & \scriptsize{Our w/BCG} & \scriptsize{Our wo/BCG} 
         \end{tabular}		

    \end{center}
	\caption{Reconstructed meshes for the \textit{truck} and \textit{ignatius}. Note that our meshes are smoother due to the vertex resolution.}
	\label{fig:Tank_Temple_meshes}
\end{figure*}

\begin{figure*}[t]

	\begin{center}
	\begin{tabular}{ccccccc}

        \includegraphics[width = .14\textwidth, trim=0.cm 0cm 0cm 0cm,clip]{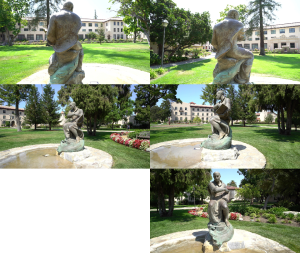}&
        \includegraphics[width = .14\textwidth, trim=3.cm 0cm 3cm 0cm,clip]{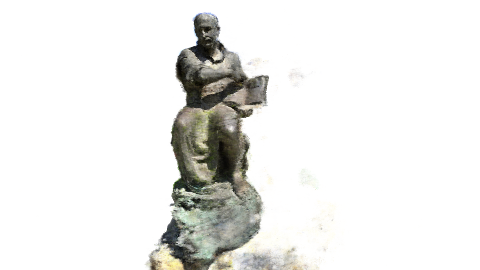}&
		\includegraphics[width = .14\textwidth, trim=3.cm 0cm 3cm 0cm,clip]{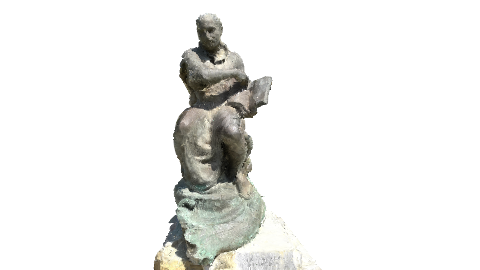}&
        \includegraphics[width = .14\textwidth, trim=3.cm 0cm 3cm 0cm,clip]{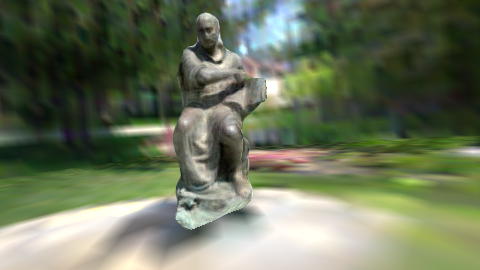}&
        \includegraphics[width = .14\textwidth, trim=3.cm 0cm 3cm 0cm,clip]{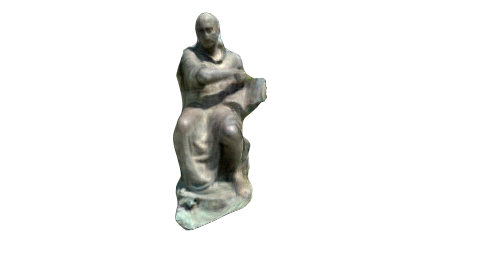}&
		\includegraphics[width = .14\textwidth, trim=3.cm 0cm 3cm 0cm,clip]{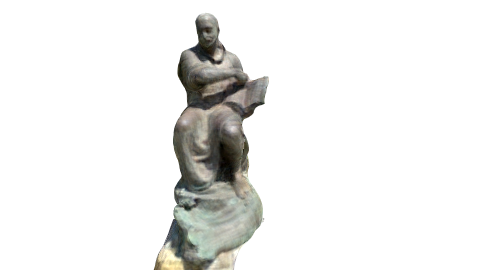}&
		\includegraphics[width = .14\textwidth, trim=3.cm 0cm 3cm 0cm,clip]{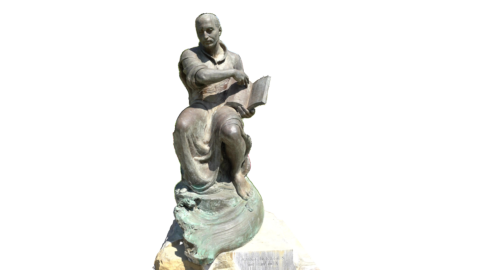} \\
        \includegraphics[width = .14\textwidth, trim=0.cm 0cm 0cm 0cm,clip]{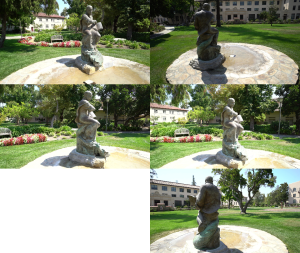}&
        \includegraphics[width = .14\textwidth, trim=3.cm 0cm 3cm 0cm,clip]{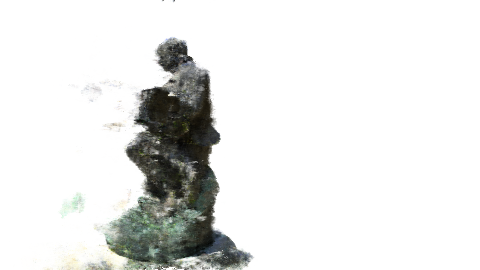}&
		\includegraphics[width = .14\textwidth, trim=3.cm 0cm 3cm 0cm,clip]{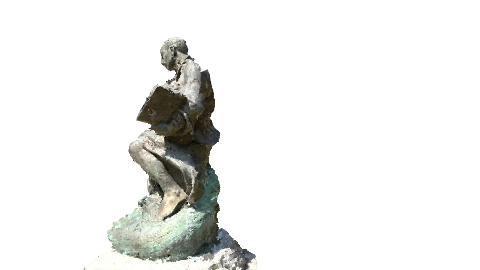}&
        \includegraphics[width = .14\textwidth, trim=3.cm 0cm 3cm 0cm,clip]{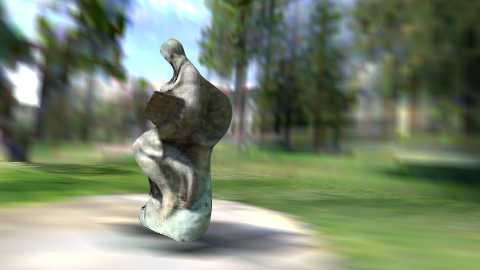}&
        \includegraphics[width = .14\textwidth, trim=3.cm 0cm 3cm 0cm,clip]{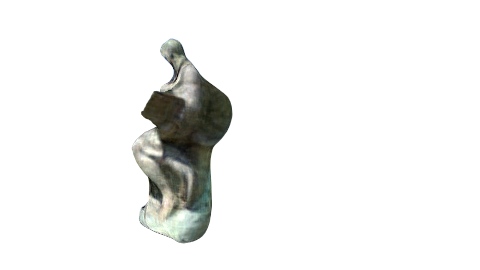}&
		\includegraphics[width = .14\textwidth, trim=3.cm 0cm 3cm 0cm,clip]{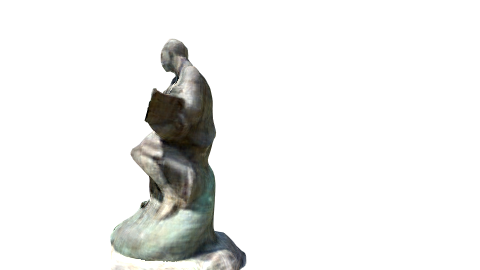}&
		\includegraphics[width = .14\textwidth, trim=3.cm 0cm 3cm 0cm,clip]{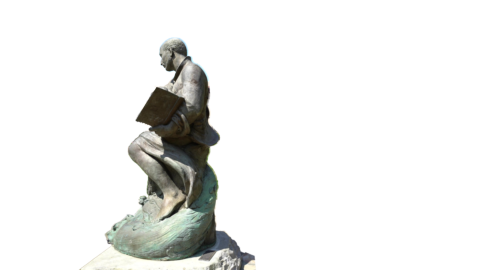} \\
        \includegraphics[width = .14\textwidth, trim=0.cm 0cm 0cm 0cm,clip]{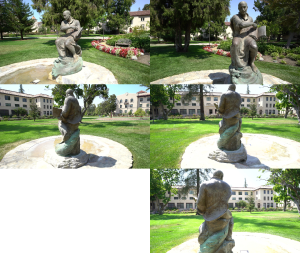}&
        \includegraphics[width = .14\textwidth, trim=3.cm 0cm 3cm 0cm,clip]{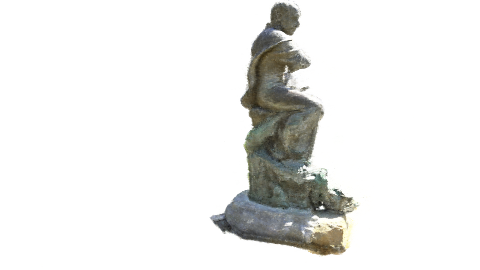}&
		\includegraphics[width = .14\textwidth, trim=3.cm 0cm 3cm 0cm,clip]{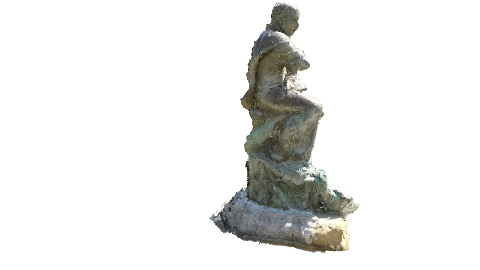}&
        \includegraphics[width = .14\textwidth, trim=3.cm 0cm 3cm 0cm,clip]{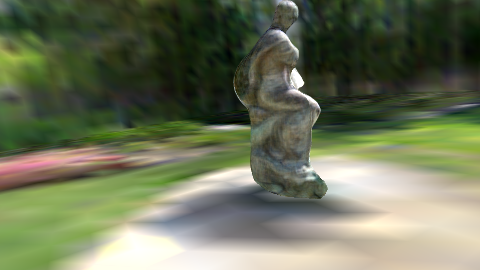}&
        \includegraphics[width = .14\textwidth, trim=3.cm 0cm 3cm 0cm,clip]{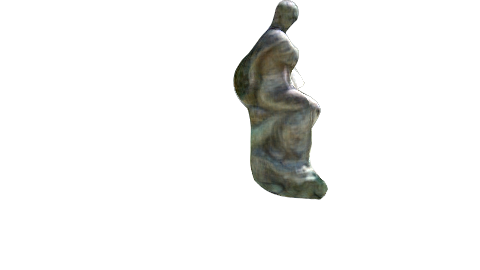}&
		\includegraphics[width = .14\textwidth, trim=3.cm 0cm 3cm 0cm,clip]{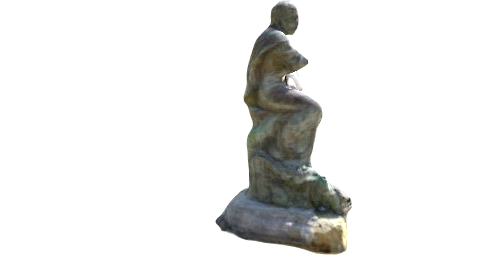}&
		\includegraphics[width = .14\textwidth, trim=3.cm 0cm 3cm 0cm,clip]{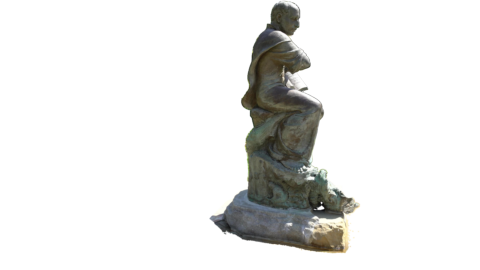} \\
        \includegraphics[width = .14\textwidth, trim=0.cm 0cm 0cm 0cm,clip]{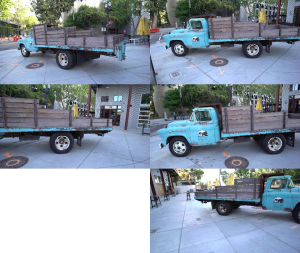}&
        \includegraphics[width = .14\textwidth, trim=0.cm 0cm 2cm 0cm,clip]{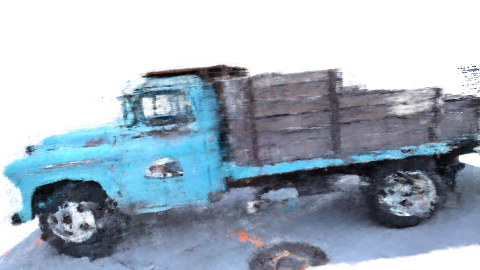}&
		\includegraphics[width = .14\textwidth, trim=0.cm 0cm 2cm 0cm,clip]{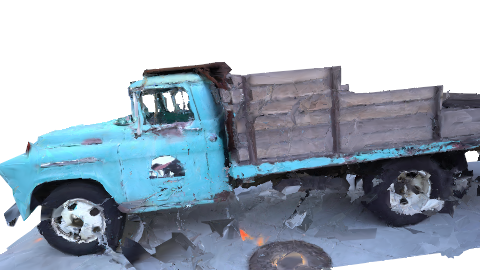}&
        \includegraphics[width = .14\textwidth, trim=0.cm 0cm 2cm 0cm,clip]{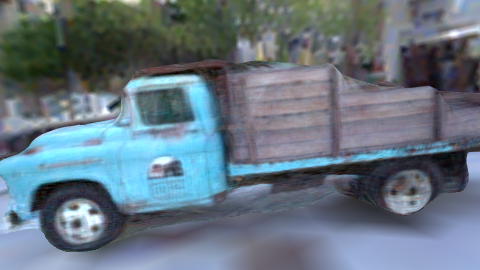}&
        \includegraphics[width = .14\textwidth,trim=0.cm 0cm 2cm 0cm,clip]{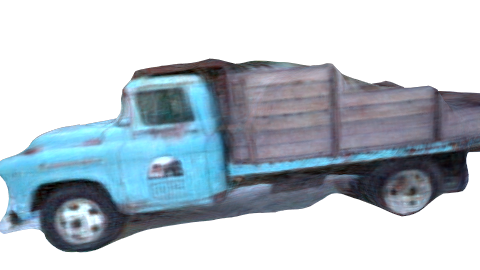}&
		\includegraphics[width = .14\textwidth,trim=0.cm 0cm 2cm 0cm,clip]{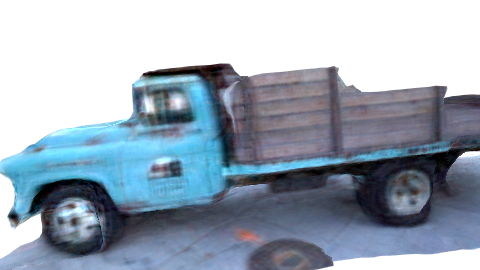}&
		\includegraphics[width = .14\textwidth, trim=0.cm 0cm 2cm 0cm,clip]{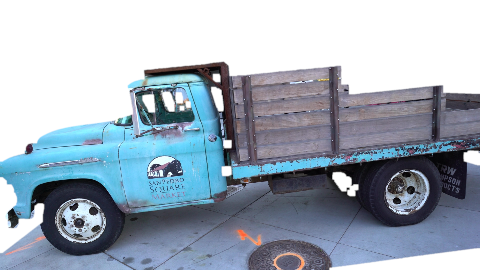} \\
        \includegraphics[width = .14\textwidth, trim=0.cm 0cm 0cm 0cm,clip]{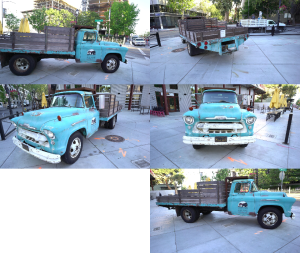}&
        \includegraphics[width = .14\textwidth, trim=0.cm 0cm 2cm 0cm,clip]{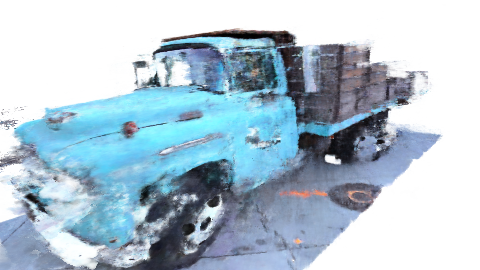}&
		\includegraphics[width = .14\textwidth, trim=0.cm 0cm 2cm 0cm,clip]{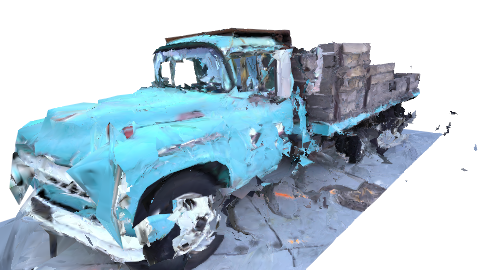}&
        \includegraphics[width = .14\textwidth, trim=0.cm 0cm 2cm 0cm,clip]{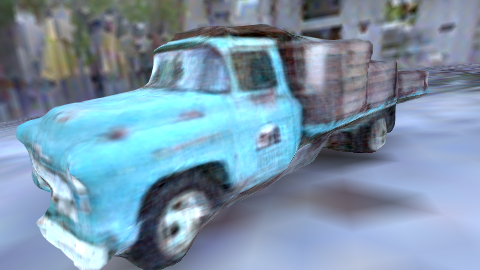}&
        \includegraphics[width = .14\textwidth,trim=0.cm 0cm 2cm 0cm,clip]{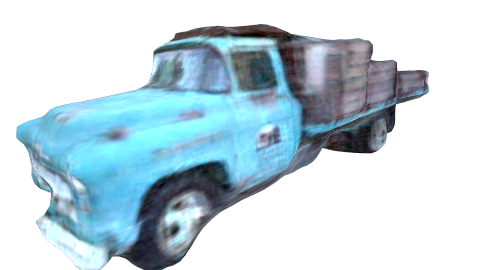}&
		\includegraphics[width = .14\textwidth,trim=0.cm 0cm 2cm 0cm,clip]{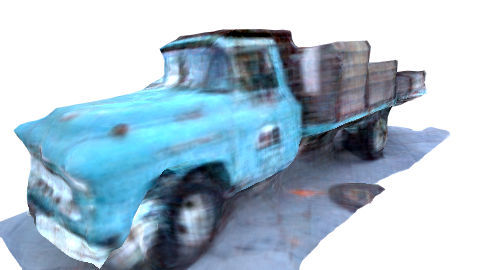}&
		\includegraphics[width = .14\textwidth, trim=0.cm 0cm 2cm 0cm,clip]{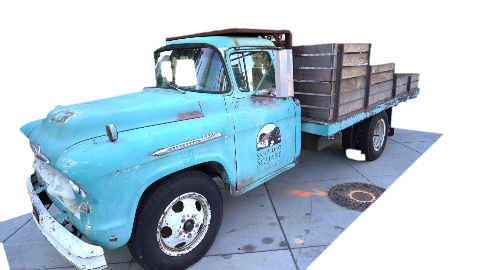} \\
        \includegraphics[width = .14\textwidth, trim=0.cm 0cm 0cm 0cm,clip]{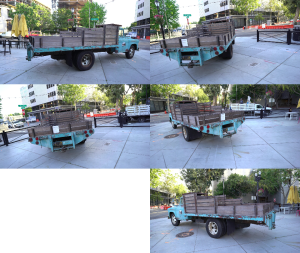}&
        \includegraphics[width = .14\textwidth, trim=0.cm 0cm 0cm 0cm,clip]{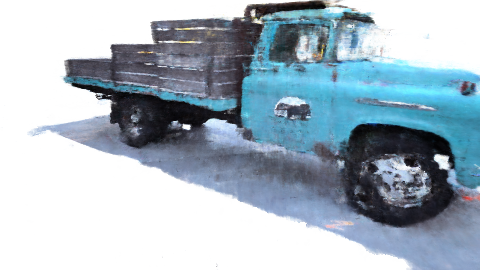}&
		\includegraphics[width = .14\textwidth, trim=0.cm 0cm 0cm 0cm,clip]{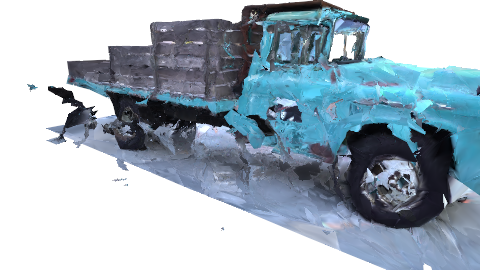}&
        \includegraphics[width = .14\textwidth, trim=0.cm 0cm 0cm 0cm,clip]{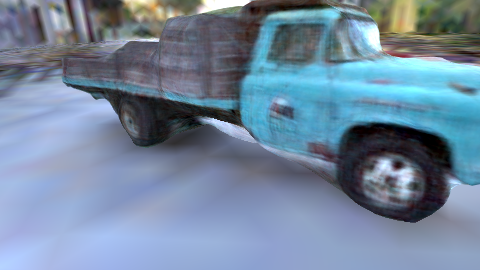}&
        \includegraphics[width = .14\textwidth,trim=0.cm 0cm 0cm 0cm,clip]{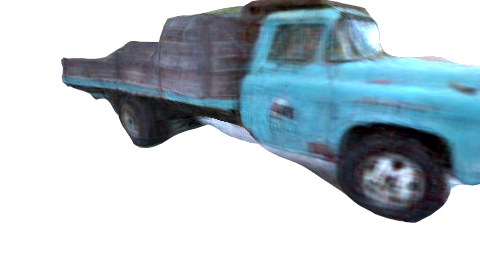}&
		\includegraphics[width = .14\textwidth,trim=0.cm 0cm 0cm 0cm,clip]{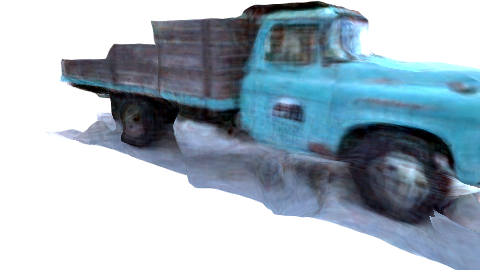}&
		\includegraphics[width = .14\textwidth, trim=0.cm 0cm 0cm 0cm,clip]{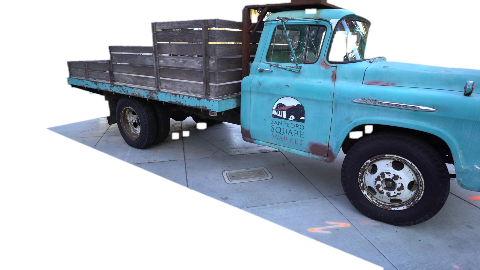} \\
		Input Images&RegNeRF  & Munkberg et al.\cite{munkberg2021nvdiffrec} & our w/BCG & our w/BCG* & our wo/BCG & GT

    \end{tabular}		

    \end{center}
	\caption{Qualitative results on the Tank and Temple Dataset. The reconstructed objects are rendered from three views. Note that in our w/BCG* column we remove the background from our w/BCG for better visualization.\label{fig:Tank_Temple_textures}}
\end{figure*}

\begin{figure*}[t]
	\begin{tabular}{cccccccc}
        \includegraphics[width = .12\linewidth,trim=0cm 0cm 0cm 0cm,clip]{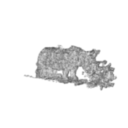}&
		\includegraphics[width = .12\linewidth,trim=0cm 0cm 0cm 0cm,clip]{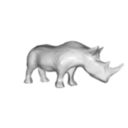} &
		\includegraphics[width = .12\linewidth,trim=0cm 0cm 0cm 0cm,clip]{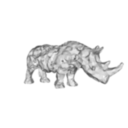} &
		\includegraphics[width = .12\linewidth,trim=0cm 0cm 0cm 0cm,clip]{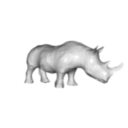} & \includegraphics[width = .12\linewidth,trim=0cm 0cm 0cm 0cm,clip]{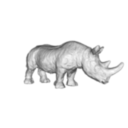} &
        \includegraphics[width = .12\linewidth,trim=0cm 0cm 0cm 0cm,clip]{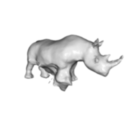}&
		\includegraphics[width = .12\linewidth,trim=0cm 0cm 0cm 0cm,clip]{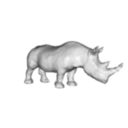} &
		\includegraphics[width = .12\linewidth, trim=0cm 0cm 0cm 0cm,clip]{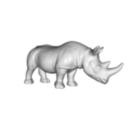}\\
        \includegraphics[width = .12\linewidth,trim=0cm 0cm 0cm 0cm,clip]{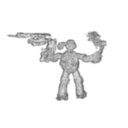}&
		\includegraphics[width = .12\linewidth,trim=0cm 0cm 0cm 0cm,clip]{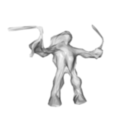} &
		\includegraphics[width = .12\linewidth,trim=0cm 0cm 0cm 0cm,clip]{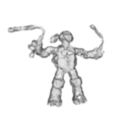} &
		\includegraphics[width = .12\linewidth,trim=0cm 0cm 0cm 0cm,clip]{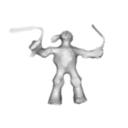} & \includegraphics[width = .12\linewidth,trim=0cm 0cm 0cm 0cm,clip]{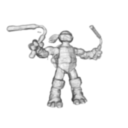} &
        \includegraphics[width = .12\linewidth,trim=0cm 0cm 0cm 0cm,clip]{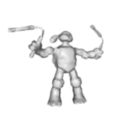}&
		\includegraphics[width = .12\linewidth,trim=0cm 0cm 0cm 0cm,clip]{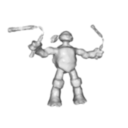} &
		\includegraphics[width = .12\linewidth, trim=0cm 0cm 0cm 0cm,clip]{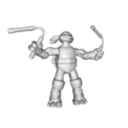}\\
        \includegraphics[width = .12\linewidth,trim=0cm 0cm 0cm 0cm,clip]{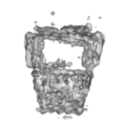}&
		\includegraphics[width = .12\linewidth,trim=0cm 0cm 0cm 0cm,clip]{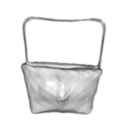} &
		\includegraphics[width = .12\linewidth, trim=0cm 0cm 0cm 0cm,clip]{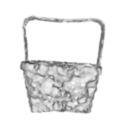} &
		\includegraphics[width = .12\linewidth,trim=0cm 0cm 0cm 0cm,clip]{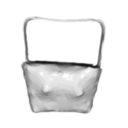} &
        \includegraphics[width = .12\linewidth,trim=0cm 0cm 0cm 0cm,clip]{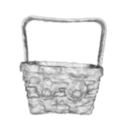} &
        \includegraphics[width = .12\linewidth,trim=0cm 0cm 0cm 0cm,clip]{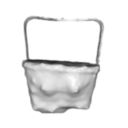}&
		\includegraphics[width = .12\linewidth,trim=0cm 0cm 0cm 0cm,clip]{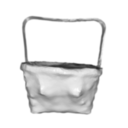} &
		\includegraphics[width = .12\linewidth, trim=0cm 0cm 0cm 0cm,clip]{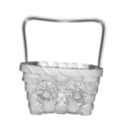}\\
        \includegraphics[width = .12\linewidth,trim=0cm 0cm 0cm 0cm,clip]{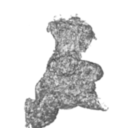}&
		\includegraphics[width = .12\linewidth,trim=0cm 0cm 0cm 0cm,clip]{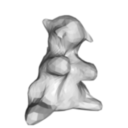} &
		\includegraphics[width = .12\linewidth,trim=0cm 0cm 0cm 0cm,clip]{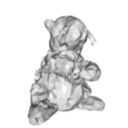} &
		\includegraphics[width = .12\linewidth,trim=0cm 0cm 0cm 0cm,clip]{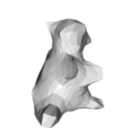} &
		\includegraphics[width = .12\linewidth,trim=0cm 0cm 0cm 0cm,clip]{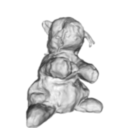} &
        \includegraphics[width = .12\linewidth,trim=0cm 0cm 0cm 0cm,clip]{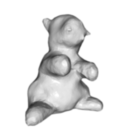}&
		\includegraphics[width = .12\linewidth,trim=0cm 0cm 0cm 0cm,clip]{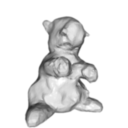} &
		\includegraphics[width = .12\linewidth,trim=0cm 0cm 0cm 0cm,clip]{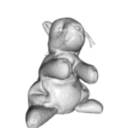}\\
        \includegraphics[width = .12\linewidth,trim=0cm 0cm 0cm 0cm,clip]{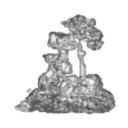}&
		\includegraphics[width = .12\linewidth,trim=0cm 0cm 0cm 0cm,clip]{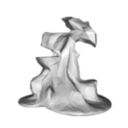} &
		\includegraphics[width = .12\linewidth, trim=0cm 0cm 0cm 0cm,clip]{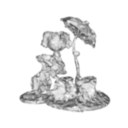} &
		\includegraphics[width = .12\linewidth,trim=0cm 0cm 0cm 0cm,clip]{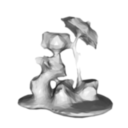} &
		\includegraphics[width = .12\linewidth,trim=0cm 0cm 0cm 0cm,clip]{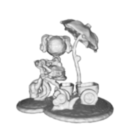} &
        \includegraphics[width = .12\linewidth,trim=0cm 0cm 0cm 0cm,clip]{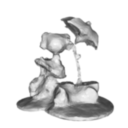}&
		\includegraphics[width = .12\linewidth,trim=0cm 0cm 0cm 0cm,clip]{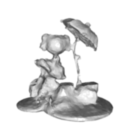} &
		\includegraphics[width = .12\linewidth, trim=0cm 0cm 0cm 0cm,clip]{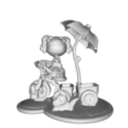}\\
        \includegraphics[width = .12\linewidth,trim=0cm 0cm 0cm 0cm,clip]{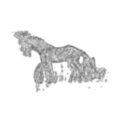}&
		\includegraphics[width = .12\linewidth,trim=0cm 0cm 0cm 0cm,clip]{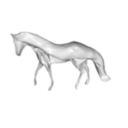} &
		\includegraphics[width = .12\linewidth, trim=0cm 0cm 0cm 0cm,clip]{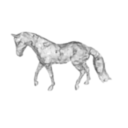} &
  	\includegraphics[width = .12\linewidth,trim=0cm 0cm 0cm 0cm,clip]
        {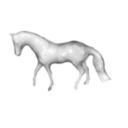} &
        \includegraphics[width = .12\linewidth,trim=0cm 0cm 0cm 0cm,clip]
        {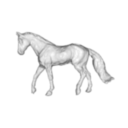} &
        \includegraphics[width = .12\linewidth,trim=0cm 0cm 0cm 0cm,clip]{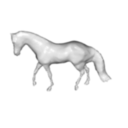}&
		\includegraphics[width = .12\linewidth,trim=0cm 0cm 0cm 0cm,clip]{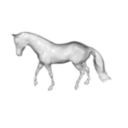} &
		\includegraphics[width = .12\linewidth, trim=0cm 0cm 0cm 0cm,clip]{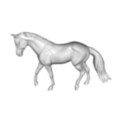}\\
       \includegraphics[width = .12\linewidth,trim=0cm 0cm 0cm 0cm,clip]{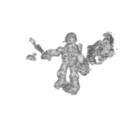}&
		\includegraphics[width = .12\linewidth,trim=0cm 0cm 0cm 0cm,clip]{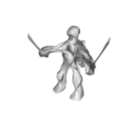} &
		\includegraphics[width = .12\linewidth, trim=0cm 0cm 0cm 0cm,clip]{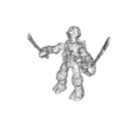} &
		\includegraphics[width = .12\linewidth,trim=0cm 0cm 0cm 0cm,clip]{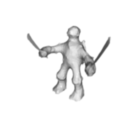} &
		\includegraphics[width = .12\linewidth,trim=0cm 0cm 0cm 0cm,clip]{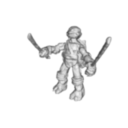} &
        \includegraphics[width = .12\linewidth,trim=0cm 0cm 0cm 0cm,clip]{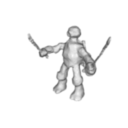}&
		\includegraphics[width = .12\linewidth,trim=0cm 0cm 0cm 0cm,clip]{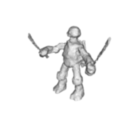} &
		\includegraphics[width = .12\linewidth, trim=0cm 0cm 0cm 0cm,clip]{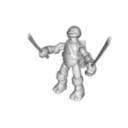}\\
         \scriptsize{RegNeRF} & \scriptsize{NeRS} & \scriptsize{Munkberg et al.~\cite{munkberg2021nvdiffrec}} & \scriptsize{DS} & \scriptsize{NeuS wo/BCG}& \scriptsize{Our w/BCG} & \scriptsize{Our wo/BCG} & \scriptsize{GT}
    \end{tabular}		

	\caption{Reconstructed meshes for more objects in the GSO Dataset.
	}
	\label{fig:GSO_supplementary_meshes}
\end{figure*}

\begin{figure*}[t]
	\begin{tabular}{ccccccc}

       \includegraphics[width = .14\linewidth,trim=0cm 0cm 0cm 0cm,clip]{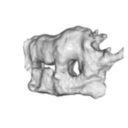}&
        \includegraphics[width = .14\linewidth,trim=0cm 0cm 0cm 0cm,clip]{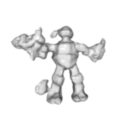} &
        \includegraphics[width = .14\linewidth,trim=0cm 0cm 0cm 0cm,clip]{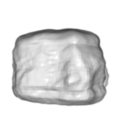} &
		\includegraphics[width = .14\linewidth,trim=0cm 0cm 0cm 0cm,clip]{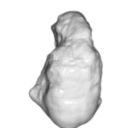} &
       \includegraphics[width = .14\linewidth,trim=0cm 0cm 0cm 0cm,clip]{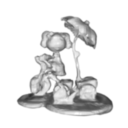} &
       \includegraphics[width = .14\linewidth,trim=0cm 0cm 0cm 0cm,clip]{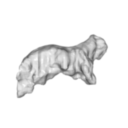} &
      \includegraphics[width = .14\linewidth,trim=0cm 0cm 0cm 0cm,clip]{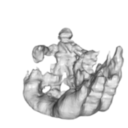}
    \end{tabular}		
    
	\caption{Reconstructed meshes for NeuS with background. The corresponding GT meshes are shown in Figure~\ref{fig:GSO_supplementary_meshes} (from top to bottom) } 
    \label{fig:Neus_wBCG_GSO}	
\end{figure*}

In Figure~\ref{fig:Tank_Temple_meshes} and Figure~\ref{fig:Tank_Temple_textures} we show qualitative results for the Tank and Temple datasets. 
Note that, for the truck object \textit{our w/BCG} is able to reconstruct the object, but not the bottom part around it as this was part of the background mesh during training.
For the ignatius object, \textit{our w/BCG} could not recover the bottom (seddle) part and \textit{our wo/BCG} could recover it partially.
The part that was not recovered has a high brightness (almost white color) and there is not enough signal (feedback) from the RGB loss to recover that part. 
Observe that the performance of NeUS shows a significant decrease when the background is present (without mask supervision).
We have smoother meshes for both objects than the baselines. 
One can increase the vertex resolution to recover more details. 
In Figures~\ref{fig:GSO_supplementary_meshes},~\ref{fig:qualitative_results_google_texture} and \ref{fig:qualitative_results_mvmc_car} we show reconstructions for more objects in the GSO dataset and MVMC car dataset respectively. 
Figure~\ref{fig:Neus_wBCG_GSO} shows the reconstructed mashes for NeuS w/BCG. Notice that the method cannot accurately estimate the mesh for most of the objects.

\renewcommand{\tabcolsep}{0.5pt}
\begin{figure*}[t]
	\begin{center}
	\begin{tabular}{cccccccc}

        \includegraphics[width = .12\textwidth, trim=0.cm 0cm 0cm 0cm,clip]{Figures/Google_Texture/cow/cow_merged_400.png}&
        \includegraphics[width = .12\textwidth, trim=0.cm 1.5cm 1.5cm 1.5cm,clip]{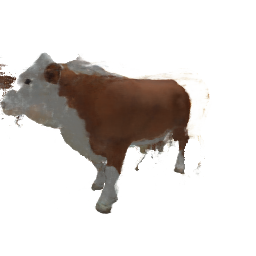}&
		\includegraphics[width = .12\textwidth, trim=0.cm 1.5cm 1.5cm 1.5cm,clip]{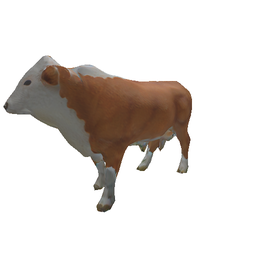} &
		\includegraphics[width = .12\textwidth, trim=0.cm 1.5cm 1.5cm 1.5cm,clip]{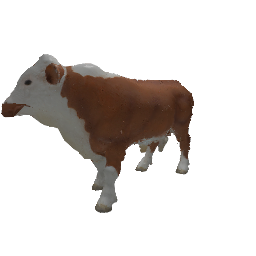}&
        \includegraphics[width = .12\textwidth, trim=0.cm 1.5cm 1.5cm 1.5cm,clip]{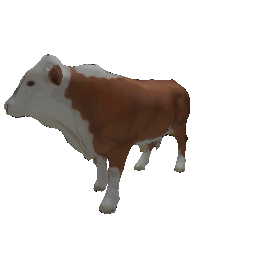}&
        \includegraphics[width = .12\textwidth, trim=0.cm 1.5cm 1.5cm 1.5cm,clip]{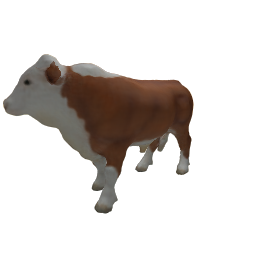}&
		\includegraphics[width = .12\textwidth, trim=0.cm 1.5cm 1.5cm 1.5cm,clip]{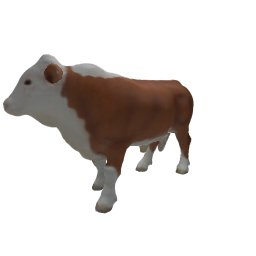}&
		\includegraphics[width = .12\textwidth, trim=0.cm 1.5cm 1.5cm 1.5cm,clip]{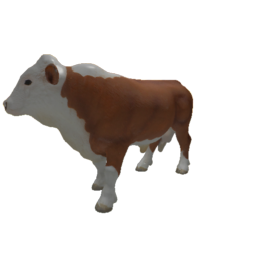} \\
        \includegraphics[width = .12\textwidth, trim=0.cm 0cm 0cm 0cm,clip]{Figures/Google_Texture/mario/mario_merged_400.png}&
		\includegraphics[width = .12\textwidth,trim=1.5cm 0.5cm 1.5cm 0.5cm,clip]{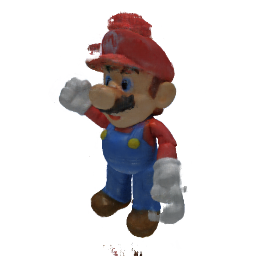}&
		\includegraphics[width = .12\textwidth,trim=1.5cm 0.5cm 1.5cm 0.5cm,clip]{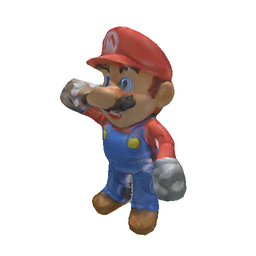}&
		\includegraphics[width = .12\textwidth,trim=1.5cm 0.5cm 1.5cm 0.5cm,clip]{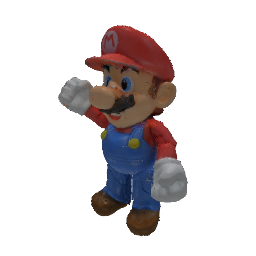}&
        \includegraphics[width = .12\textwidth,trim=1.5cm 0.5cm 1.5cm 0.5cm,clip]{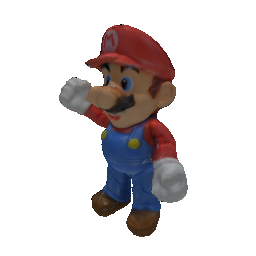}&
        \includegraphics[width = .12\textwidth,trim=1.5cm 0.5cm 1.5cm 0.5cm,clip]{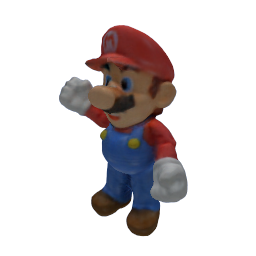}&
        \includegraphics[width = .12\textwidth,trim=1.5cm 0.5cm 1.5cm 0.5cm,clip]{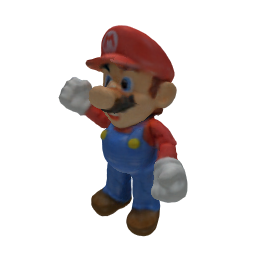}&
		\includegraphics[width = .12\textwidth,trim=1.5cm 0.5cm 1.5cm 0.5cm,clip]{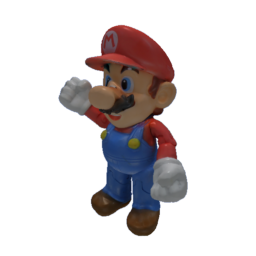}\\
        \includegraphics[width = .12\textwidth, trim=0.cm 0cm 0cm 0cm,clip]{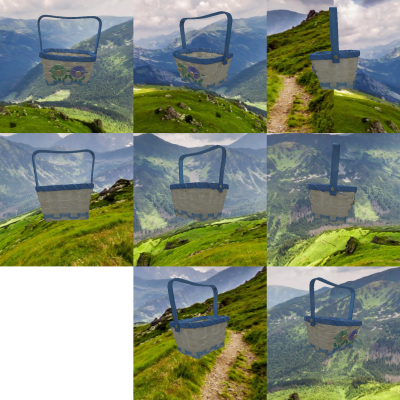}&
		\includegraphics[width = .12\textwidth, trim=2.2cm 2.5cm 1.8cm 0.5cm,clip]{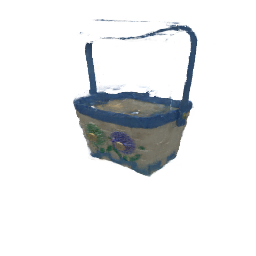}&
		\includegraphics[width = .12\textwidth,trim=1.3cm 1.cm 1.3cm 0.8cm,clip]{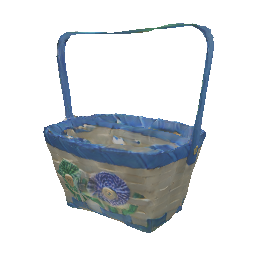}&
		\includegraphics[width = .12\textwidth,trim=2.2cm 2.5cm 1.8cm 0.5cm,clip]{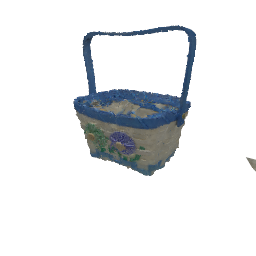}&
        \includegraphics[width = .12\textwidth,trim=2.2cm 2.5cm 1.8cm 0.5cm,clip]{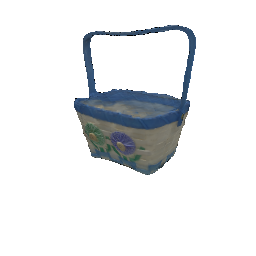}&
        \includegraphics[width = .12\textwidth,trim=2.2cm 2.5cm 1.8cm 0.5cm,clip]{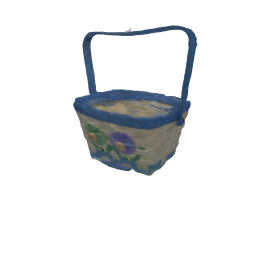}&
		\includegraphics[width = .12\textwidth,trim=2.2cm 2.5cm 1.8cm 0.5cm,clip]{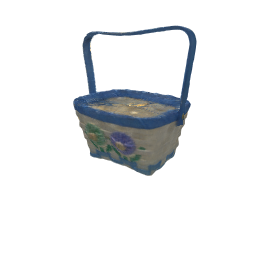}&
		\includegraphics[width = .12\textwidth,trim=2.2cm 2.5cm 1.8cm 0.5cm,clip]{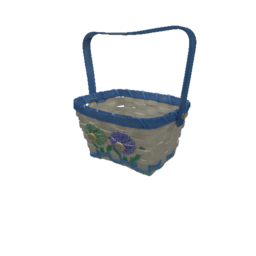}\\

		Input Images&RegNeRF & NeRS & Munkberg et al.~\cite{munkberg2021nvdiffrec} & DS & our w/BCG* & our wo/BCG & GT

    \end{tabular}		

    \end{center}
	\caption{Qualitative Results on the GSO Dataset. Note that in our w/BCG* column we remove the background from our w/BCG for better visualization. Differences can be better appreciated by zooming in.\label{fig:qualitative_results_google}}
\end{figure*}

\renewcommand{\tabcolsep}{0.5pt}

\begin{figure*}[t]
	\begin{tabular}{cccccccc}
        \multicolumn{2}{c}{\textit{Hereford Bull}} &  \multicolumn{2}{c}{\textit{Nintendo Mario}} &  \multicolumn{2}{c}{\textit{Ninja Turtles Michelangelo}}  & \multicolumn{2}{c}{\textit{Lalaloopsy Peanut}} \\
        \includegraphics[width = .12\textwidth,trim=1.25cm 0.75cm 0.75cm 0.5cm,clip]{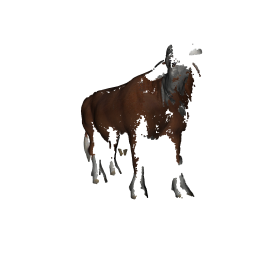}&
		\includegraphics[width = .125\textwidth,trim=0.cm 0cm 0cm 0cm,clip]{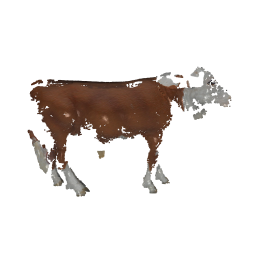} &
		\includegraphics[width = .125\textwidth, trim=0.75cm 0cm 0.75cm 0cm,clip]{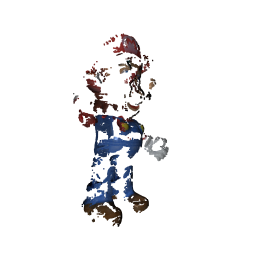}&
        \includegraphics[width = .125\textwidth, trim=0.75cm 0.cm 0.75cm 0.cm,clip]{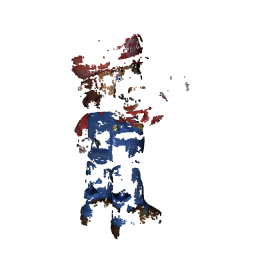}&
		\includegraphics[width = .125\textwidth, trim=0.25cm 0cm 0.25cm 0cm,clip]{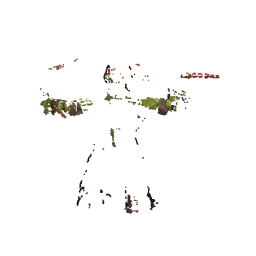}&
		\includegraphics[width = .125\textwidth,trim=0.cm 0cm 0.5cm 0cm,clip]{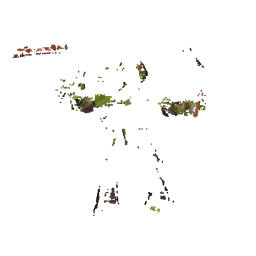} &
        \includegraphics[width = .125\textwidth, trim=0.cm 0cm 0cm 0cm,clip]{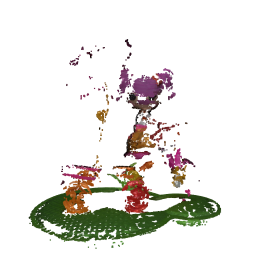}  &
        \includegraphics[width = .125\textwidth, trim=0.cm 0cm 0cm 0cm,clip]{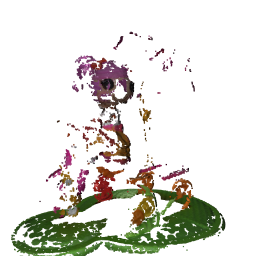}\\
          view 1 & view 2 & view 1 & view 2 & view 1 & view 2  & view 1 & view 2 \\
         &  &  & &  &  &  & \\
        \multicolumn{2}{c}{\textit{Ninja Turtles Leonardo}} & \multicolumn{2}{c}{\textit{Fire Engine}}  & \multicolumn{2}{c}{\textit{Breyer Horse}}  & \multicolumn{2}{c}{\textit{Porcelain White Pitcher}}   \\
        \includegraphics[width = .12\textwidth,trim=0cm 0cm 0cm 0cm,clip]{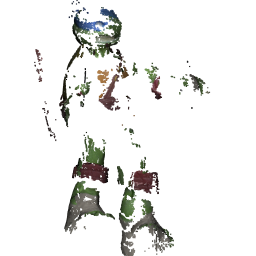}&
		\includegraphics[width = .125\textwidth,trim=0.cm 0cm 0cm 0cm,clip]{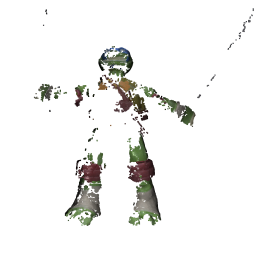} &
		\includegraphics[width = .125\textwidth, trim=0.cm 0cm 0cm 0cm,clip]{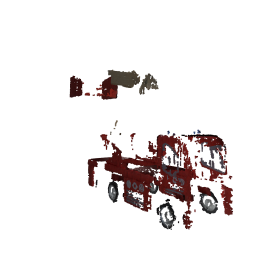}&
        \includegraphics[width = .125\textwidth, trim=0.cm 0cm 0cm 0cm,clip]{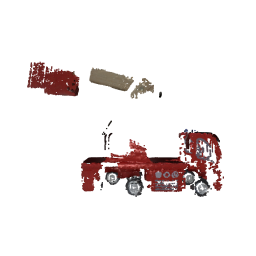}&
		\includegraphics[width = .125\textwidth, trim=0cm 0cm 0cm 0cm,clip]{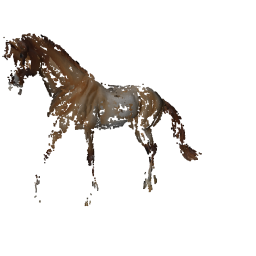}&
		\includegraphics[width = .125\textwidth,trim=0.75cm 0cm 0cm 0cm,clip]{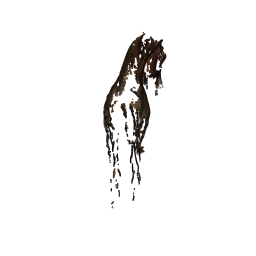} &
        \includegraphics[width = .125\textwidth, trim=0cm 0cm 0cm 0cm,clip]{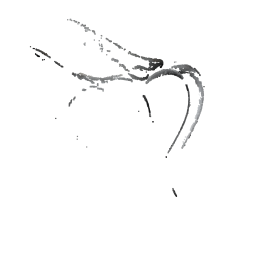}  &
        \includegraphics[width = .125\textwidth, trim=0cm 0cm 0cm 0cm,clip]{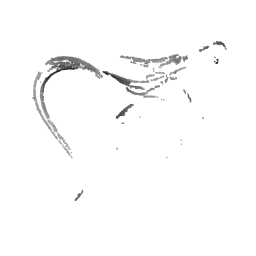}\\
       
        view 1 & view 2 & view 1 & view 2 & view 1 & view 2  & view 1 & view 2 \\
    \end{tabular}		

	\caption{. COLMAP-reconstructed dense point clouds with 8 input views from known ground-truth cameras for objects in the GSO Dataset. Better viewed by zooming in.
	}
	\label{fig:colmap_8views_dence}
\end{figure*}

\begin{figure*}[t]
	\begin{tabular}{ccccc}

		\includegraphics[width = .19\textwidth,trim=1.75cm 1.5cm 0.75cm 0.5cm,clip]{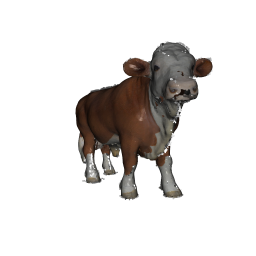} &
		\includegraphics[width = .19\textwidth, trim=1.5cm 0.5cm 1.6cm 0.5cm,clip]{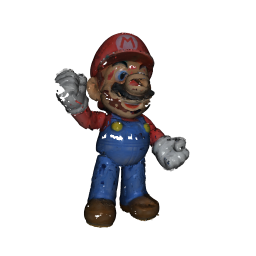}&
		\includegraphics[width = .19\textwidth,trim=0.5cm 0.5cm 1.25cm 0.25cm,clip]{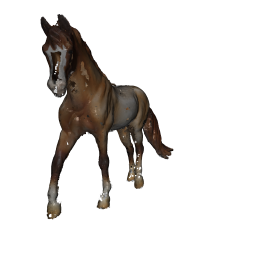} &
		\includegraphics[width = .19\textwidth, trim=1.25cm 0.25cm 1.25cm 0.25cm,clip]{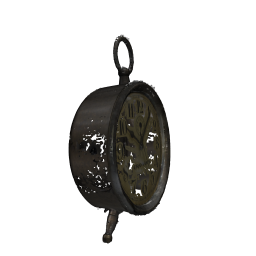} & 
  \includegraphics[width = .19\textwidth, trim=2.cm 0cm 0.5cm 1cm,clip]{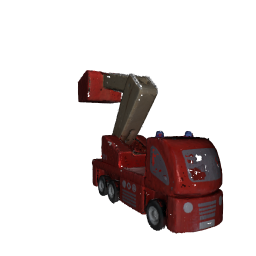}\\

	\includegraphics[width = .19\textwidth,trim=1.75cm 1.5cm 0.75cm 0.5cm,clip]             
        {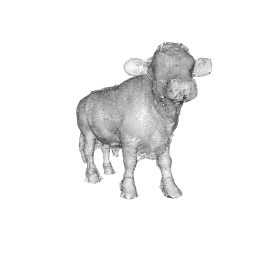} &
	  \includegraphics[width = .19\textwidth, trim=1.5cm 0.5cm 1.5cm 0.5cm,clip]{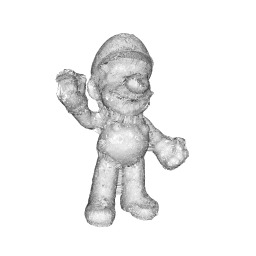}&
		
  \includegraphics[width = .19\textwidth,trim=0.5cm 0.5cm 1.25cm 0.25cm,clip]{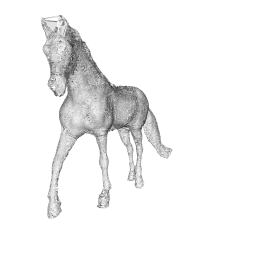} &
		\includegraphics[width = .19\textwidth, trim=1.25cm 0.25cm 1.25cm 0.25cm,clip]{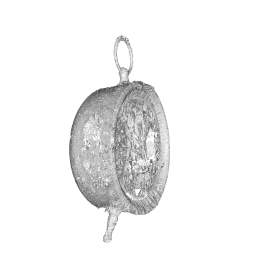} & \includegraphics[width = .19\textwidth, trim=2.cm 0cm 0.5cm 1.cm,clip]{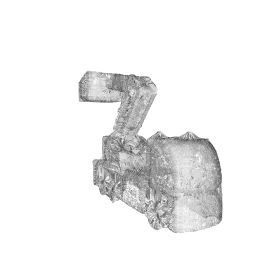} \\

  \includegraphics[width = .19\textwidth,trim=1.cm 1.5cm 1cm 1.5cm,clip]{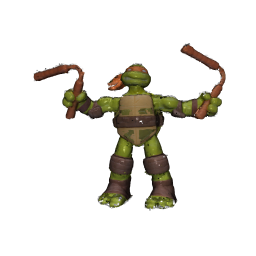} &
        \includegraphics[width = .19\textwidth, trim=0.25cm 0.75cm 0.25cm 0cm,clip]{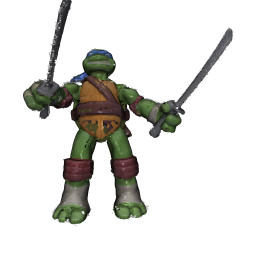}  &
		\includegraphics[width = .19\textwidth, trim=0.cm 0cm 0cm 0.75cm,clip]{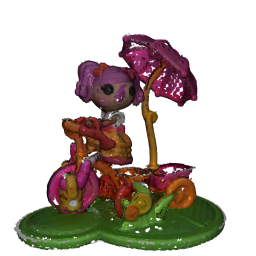}&
          \includegraphics[width = .19\textwidth, trim=0.cm 0cm 0cm 0cm,clip]{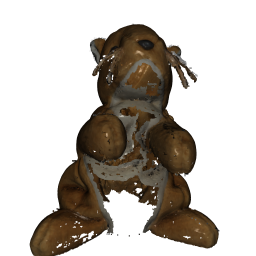} &
        \includegraphics[width = .19\textwidth, trim=0.cm 0cm 0cm 0cm,clip]{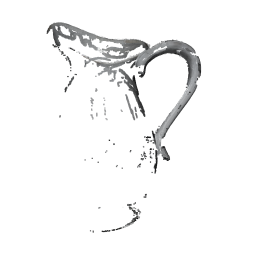} 
  \\
       
  \includegraphics[width = .19\textwidth,trim=1.cm 1.5cm 1cm 1.5cm,clip]{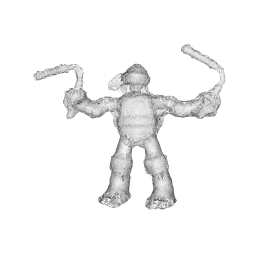} &
        \includegraphics[width = .19\textwidth, trim=0.25cm 0.75cm 0.25cm 0cm,clip]{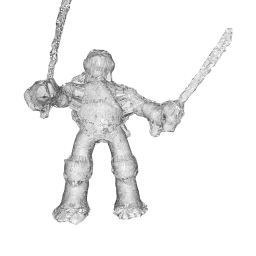}		&
		\includegraphics[width = .19\textwidth, trim=0.cm 0cm 0cm 0.75cm,clip]{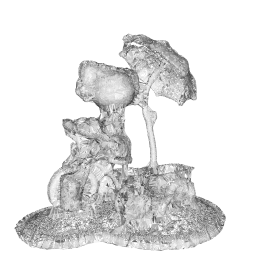}&
                \includegraphics[width = .19\textwidth, trim=0.cm 0cm 0cm 0cm,clip]{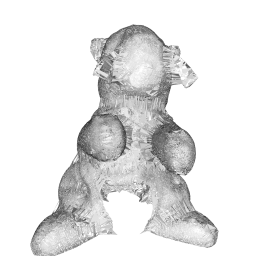}  &
        \includegraphics[width = .19\textwidth, trim=0.cm 0cm 0cm 0cm,clip]{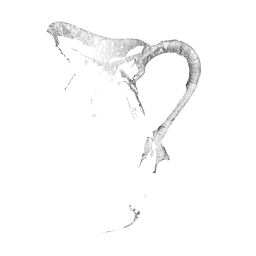}  \\
    \end{tabular}		

	\caption{COLMAP-reconstructed dense point clouds with 50 input views from known ground-truth cameras for objects in the GSO Dataset. Better viewed by zooming in.
	}
	\label{fig:colmap_50views_dense}
\end{figure*}

\begin{figure*}[t]
	\begin{tabular}{ccccc}       
  \includegraphics[width = .33\textwidth,trim=4.cm 10cm 4cm 10cm,clip]{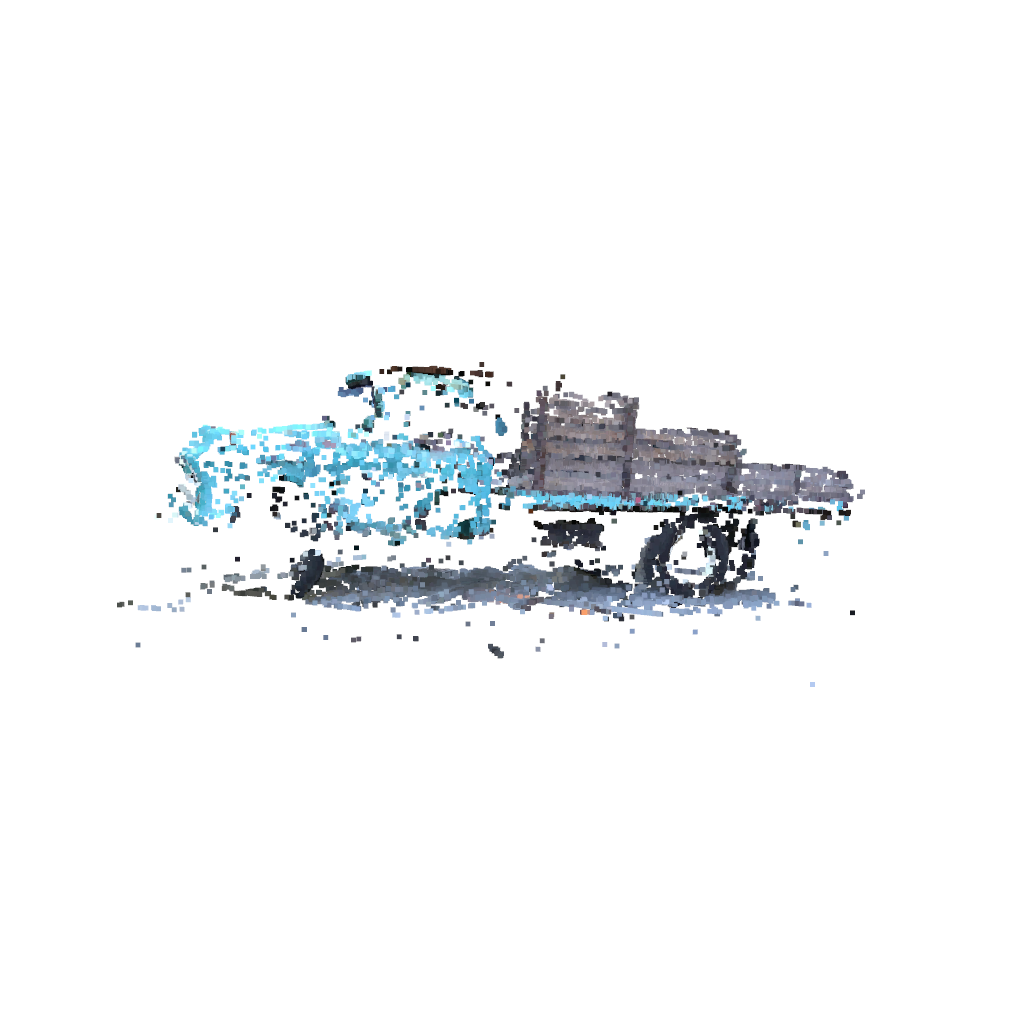} &
  \includegraphics[width = .33\textwidth, trim=4.cm 10cm 3cm 10cm,clip]{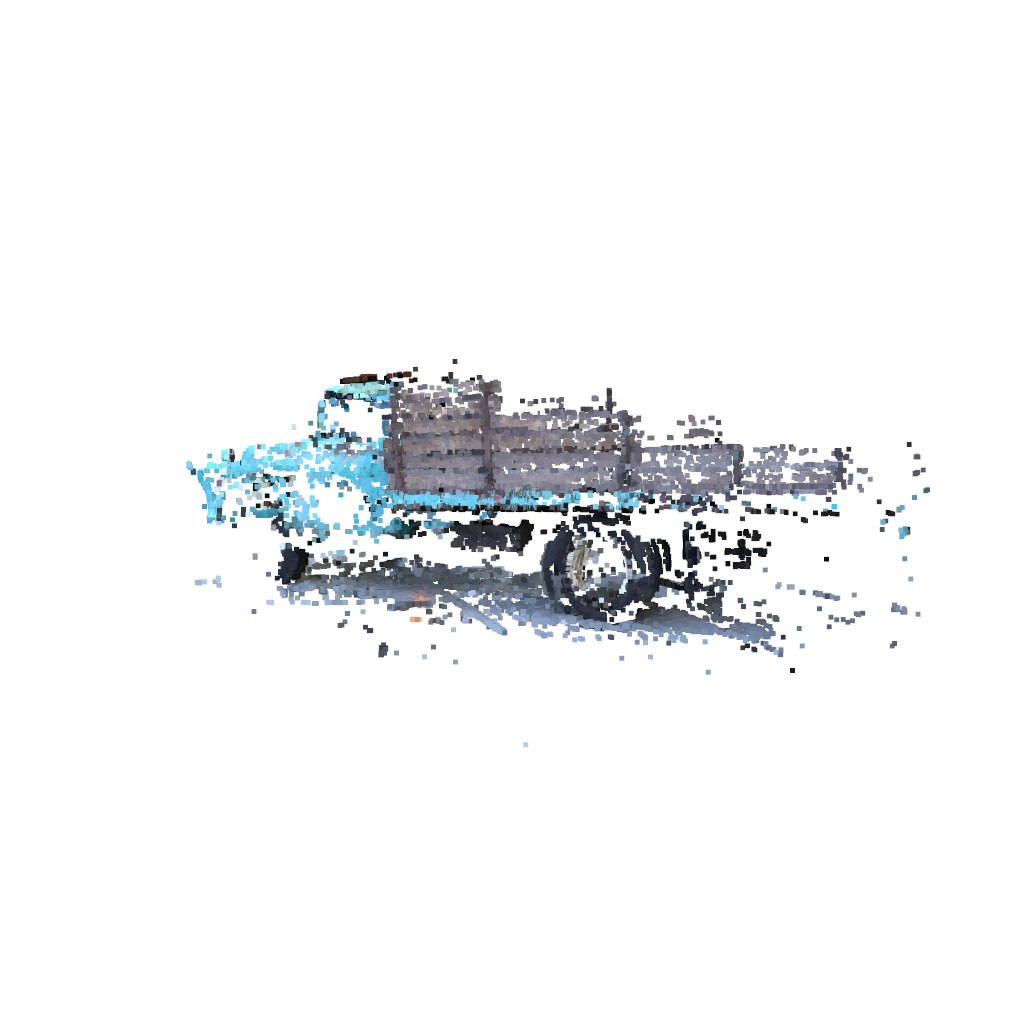}		&
  \includegraphics[width = .15\textwidth,trim=5.cm 0cm 5cm 0cm,clip]{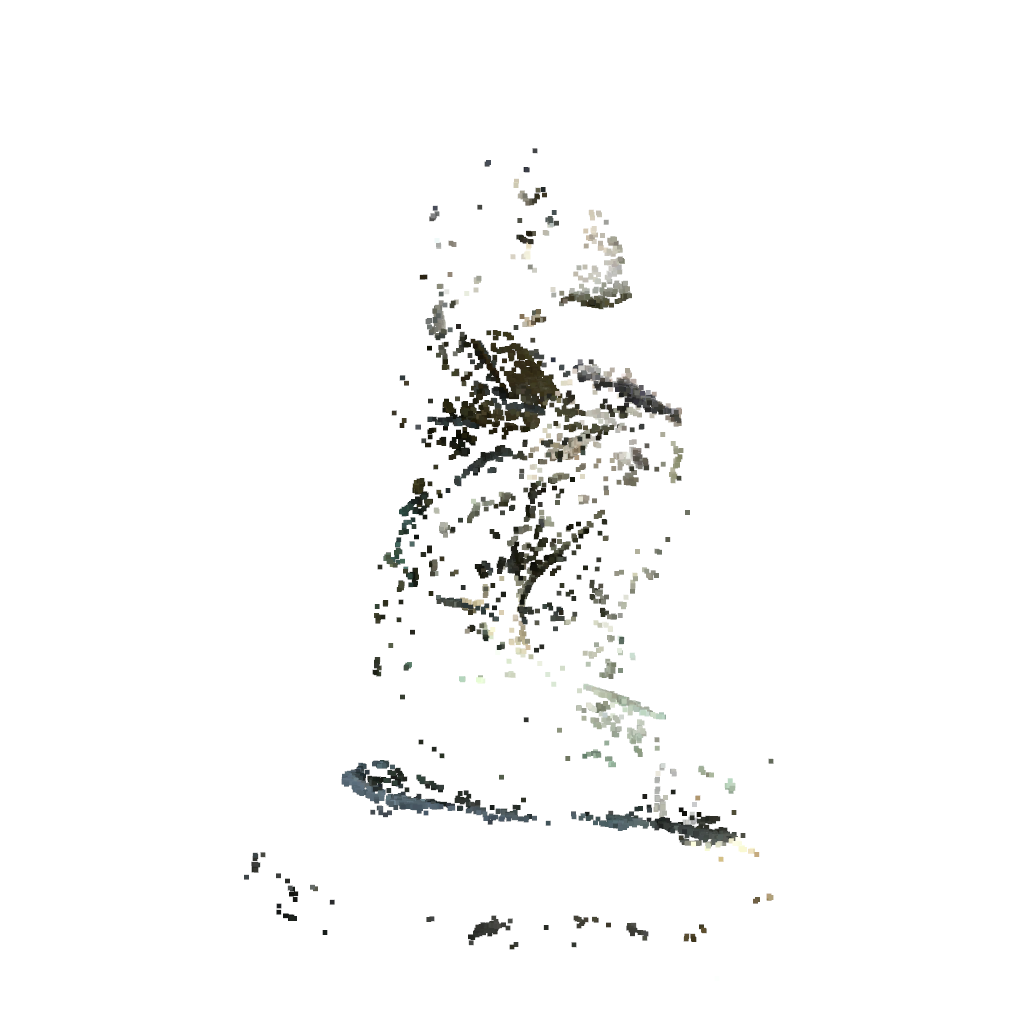} &
  \includegraphics[width = .15\textwidth,trim=5.cm 0cm 5cm 0cm,clip]{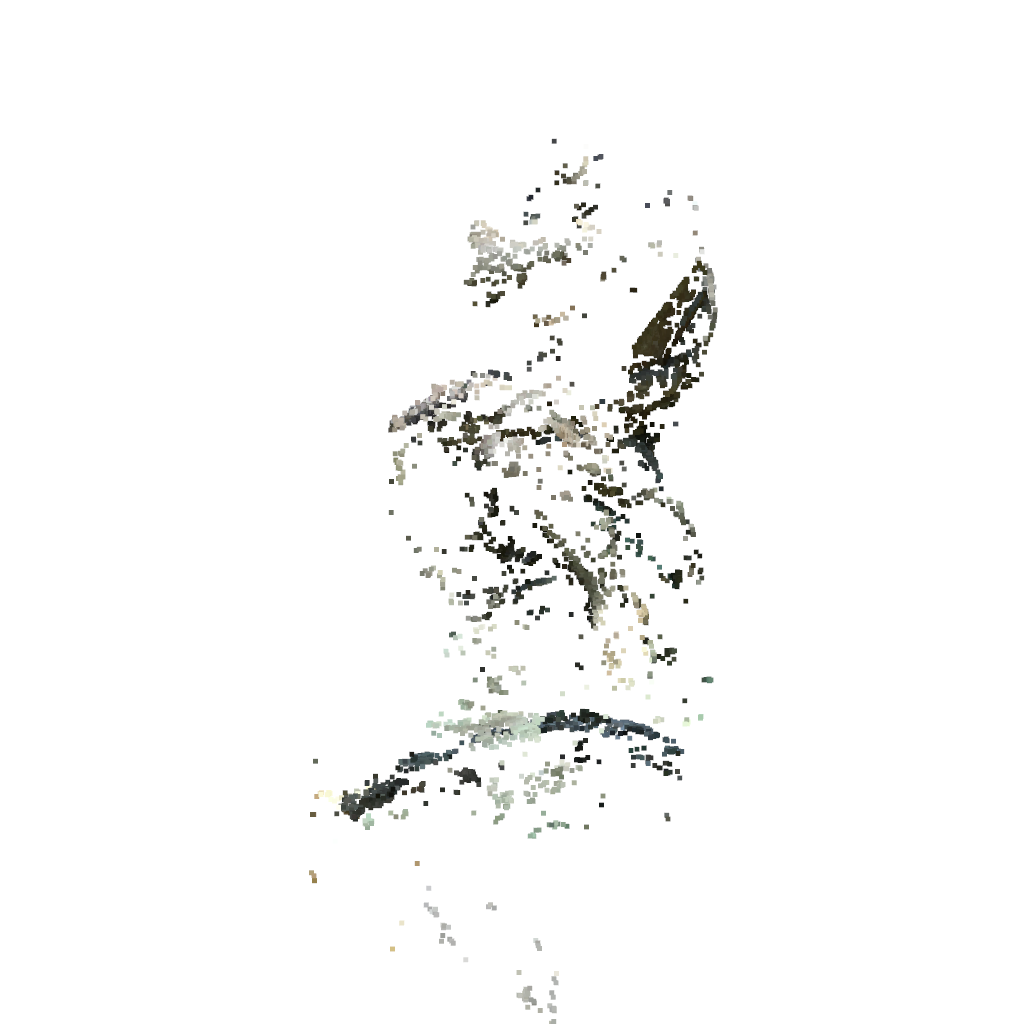}  \\
    \end{tabular}		

	\caption{COLMAP-reconstructed dense point clouds with 50 input views from known ground-truth cameras for \textit{Truck} and \textit{Ignatius} objects. Better viewed by zooming in.
	}
	\label{fig:colmap_50views_dense_truck}
\end{figure*}

\noindent\textbf{COLMAP results.} In order to increase the quality of the reconstructions we eliminate the background from the raw images and run the COLMAP with the white background. Figure~\ref{fig:colmap_8views_dence} shows dense point clouds reconstructed via COLMAP  from 8 views in the GSO dataset. We observe that the reconstructed point clouds are either indistinguishable or miss many parts from the object. This is due to the sparse view setting since any part of the object is seen from at most 3 views and as well as the poor texture of the objects.
In Figure~\ref{fig:colmap_50views_dense} and \ref{fig:colmap_50views_dense_truck} we show reconstructed dense point clouds and meshes with 50 views in the GSO dataset and Tank and Temple dataset respectively. 
The reconstructed point clouds are noticeably better than the ones with 8 views.
However, they still might have artifacts, \ie, holes and also it might still not work for objects with uniform texture, \ie, the \textit{porcelain white pitcher} (third row-last column).
Note that also the reconstructed meshes are not smooth and not watertight. One needs to post-process them via remeshing to remove many of these artifacts.

\begin{algorithm}[h]
\scriptsize
\SetAlgoLined
    \PyCode{def remesh(mesh, target\_len):} \\
    \Indp   
        \PyComment{target\_len: target edge length} \\
        \PyComment{} \\
        \PyComment{It removes triangles having collinear vertices i.e. zero areas} \\
        \PyCode{mesh = pymesh.remove\_degenerated\_triangles(mesh)}  \\
        \PyComment{It split long edges into 2 or more shorter edges} \\
        \PyCode{mesh = pymesh.split\_long\_edges(mesh, target\_len)}  \\
        \PyComment{It collapses short edges} \\
        \PyCode{mesh = pymesh.collapse\_short\_edges(mesh, 1e-6)}  \\
        \PyCode{mesh = pymesh.collapse\_short\_edges(mesh, target\_len, preserve\_feature = True}  \\
        \PyComment{It removes obtuse triangles i.e. triangles having one of the interior angles more than $90$ degrees} \\
        \PyCode{mesh = pymesh.remove\_obtuse\_triangles(mesh, 150, 100)}  \\
        \PyCode{mesh = pymesh.resolve\_self\_intersection(mesh)}  \\
        \PyComment{It removes vertices with nearly same coordinates} \\
        \PyCode{mesh = pymesh.remove\_duplicated\_faces(mesh)}  \\
        \PyComment{It computer the outer hull of the input mesh} \\
        \PyCode{mesh = pymesh.compute\_outer\_hull(mesh, 100)}  \\
        \PyCode{mesh = pymesh.remove\_duplicated\_faces(mesh)}  \\
        \PyCode{mesh = pymesh.remove\_obtuse\_triangles(mesh, 179, 5)}  \\
        \PyComment{It removes vertices not referred by any face} \\
        \PyCode{mesh = pymesh.remove\_isolated\_triangles(mesh)}  \\
        \PyCode{return mesh}\\
    \Indm 

\caption{Python style pseudocode remeshing}
\label{algo:remesh}
\end{algorithm}

\section{Discussion}

Even though our proposed method performs well, it has limitations.
For objects with thinner surfaces, one needs to increase the mesh resolution (see Figure~\ref{fig:mesh_resolution}) and this affects the computational load of the method. Although in our experiments  we found that 10K vertices works fine for most of the objects, one needs to adjust the mesh resolution for objects having more details.

The method can easily update its topology via remeshing during the learning. This allows reconstructing objects  that are not homeomorphic to spheres. However if there is a hole at the center of the ground truth object our model would not handle that as our initial shape is either a sphere or a coarse shape that is homeomorphic to a sphere. %

We handle the background by fixing its geometry to a simple shape, \eg, a cuboid, and by updating only its texture. Thus, the background is multiview consistent as well. 
Note that, in this study our goal is to model only the main object in the scene, not the whole scene.
We observed that the method is working sufficiently well if the background geometry is locally correct around the main object in the scene.
One can consider different options, \ie, generating separate background images for each training image like Monnier et al.~\cite{monnier2022unicorn} or model the background like NeRF++~\cite{nerf_plusplus}. In both cases, the background texture would not be multiview consistent and this would break the training. 
The current model might have an issue when the background is close to the foreground object, as, \eg, in the MVMC Car dataset. Handling this case is quite challenging and still an open problem. To our knowledge, we are the first among the explicit mesh-based representation methods to propose including the background in the pipeline.
We leave handling more complex background models to future work.

\renewcommand{\tabcolsep}{0pt}

\renewcommand{\tabcolsep}{0pt}

\renewcommand{\tabcolsep}{0pt}

\renewcommand{\tabcolsep}{0pt}

\begin{figure*}[t]
	\begin{center}
	\begin{tabular}{cccccccc}
        \includegraphics[width = .12\textwidth, trim=0cm 0cm 0cm 0cm,clip]{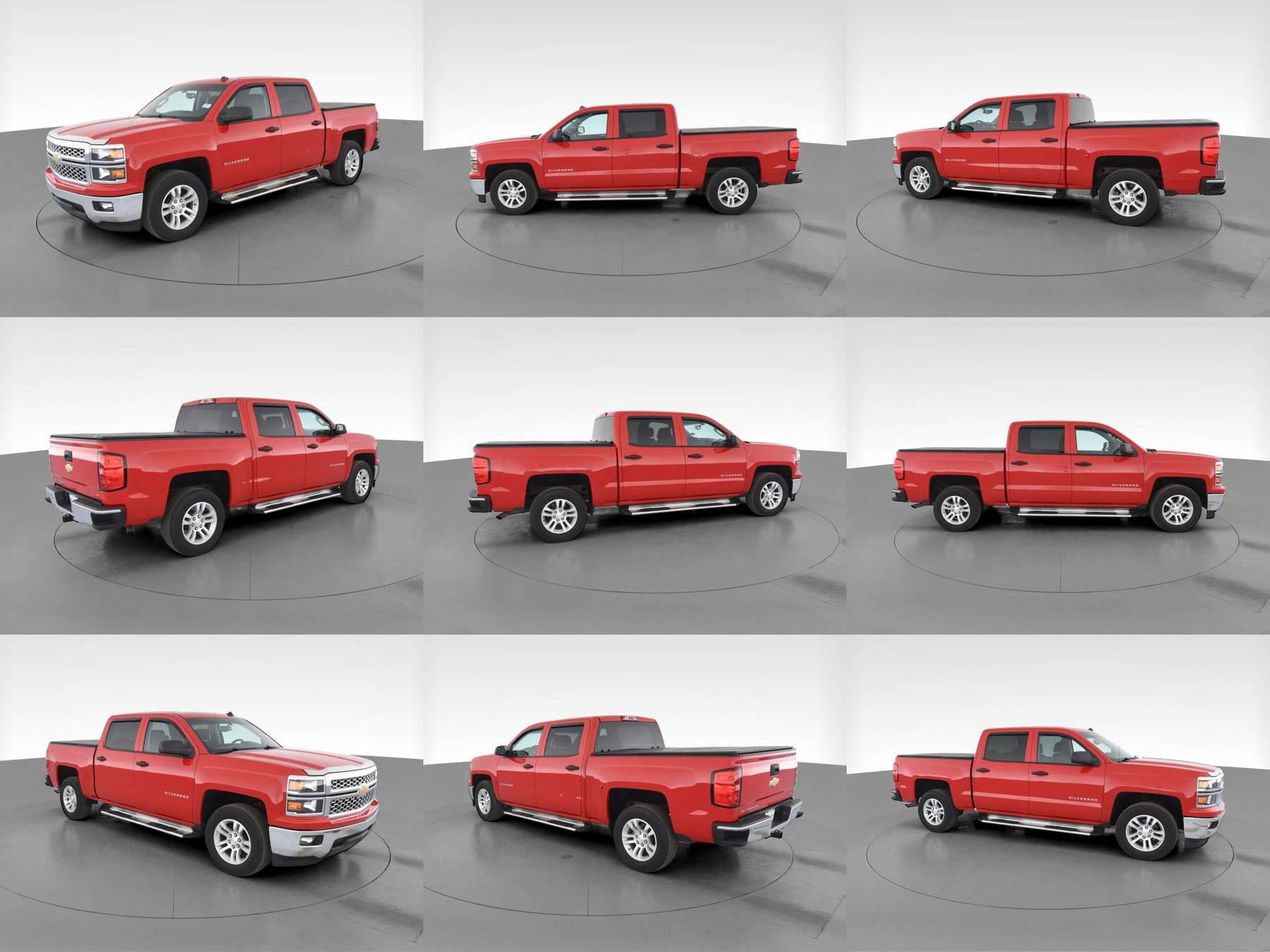} &
		\includegraphics[width = .12\textwidth, trim=0.9cm 1.5cm 0.9cm 1.5cm,clip]{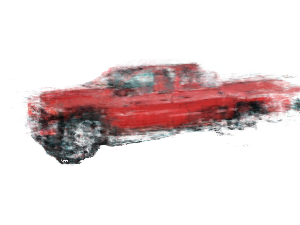}&
		\includegraphics[width = .12\textwidth, trim=0.5cm 3cm 0.5cm 3.cm,clip]{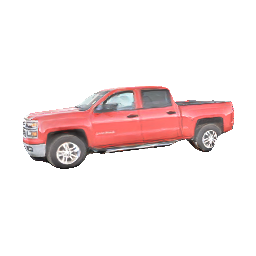}&
		\includegraphics[width = .12\textwidth, trim=0.9cm 1.5cm 0.9cm 1.5cm,clip]{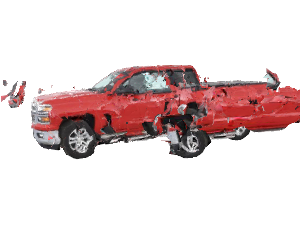}&
        \includegraphics[width = .12\textwidth, trim=0.9cm 1.5cm 0.9cm 1.5cm,clip]{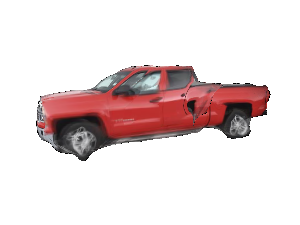}&
        \includegraphics[width = .12\textwidth, trim=0.9cm 1.5cm 0.9cm 1.5cm,clip]{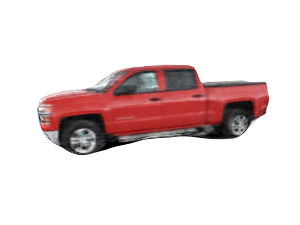}&
		\includegraphics[width = .12\textwidth, trim=0.9cm 1.5cm 0.9cm 1.5cm,clip]{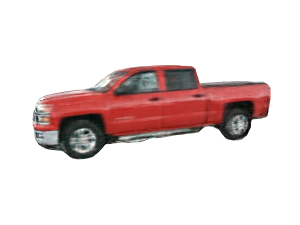} &
		\includegraphics[width = .12\textwidth, trim=0.9cm 1.5cm 0.9cm 1.5cm,clip]{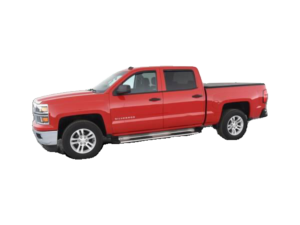} \\
        \includegraphics[width = .12\textwidth, trim=0cm 0cm 0cm 0cm,clip]{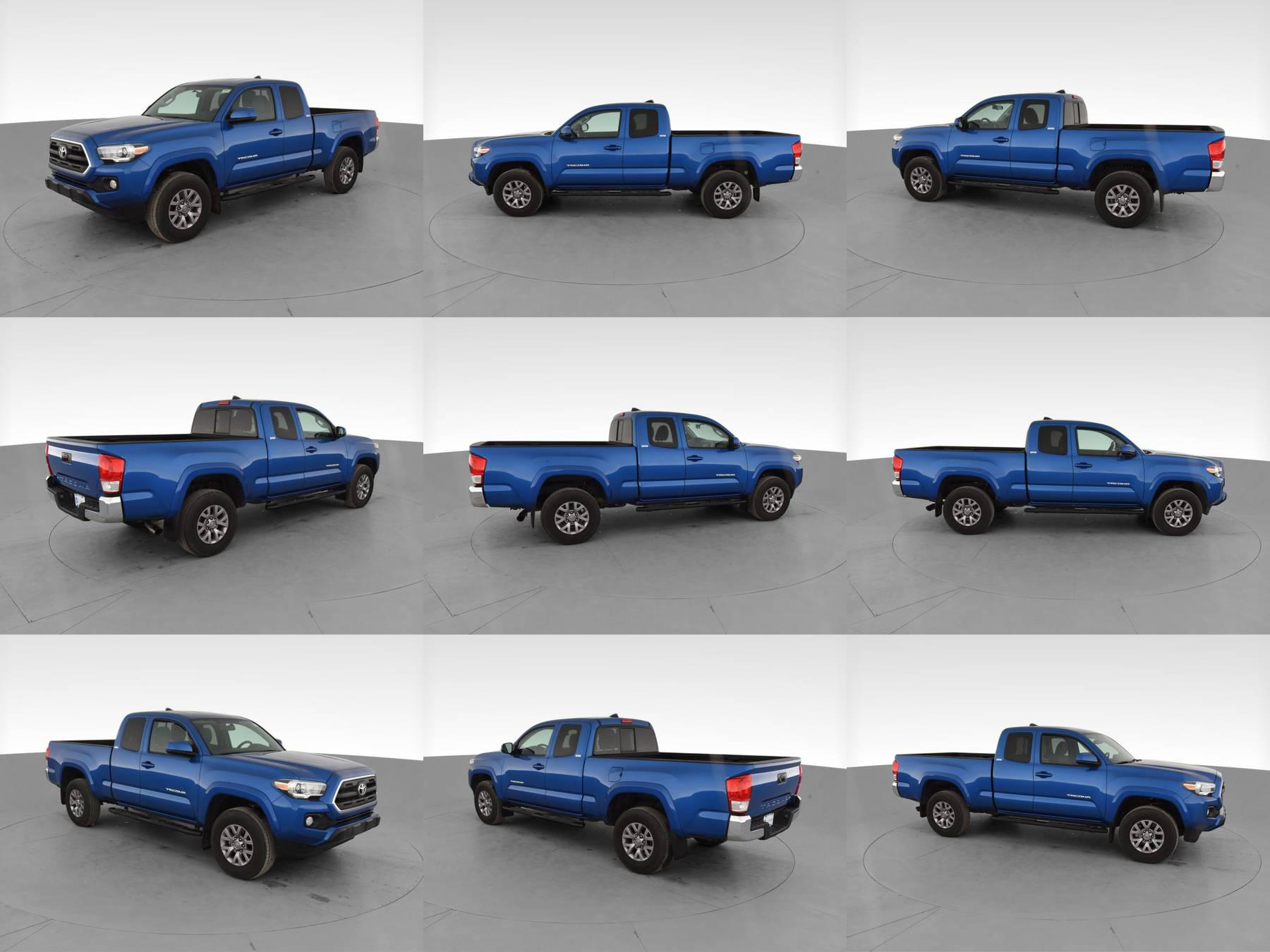} &
		\includegraphics[width = .12\textwidth, trim=0.9cm 1.5cm 0.9cm 1.5cm,clip]{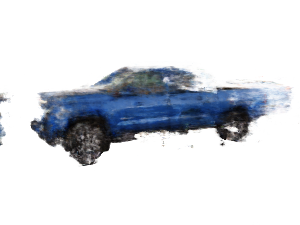}&
		\includegraphics[width = .12\textwidth, trim=0.5cm 3cm 0.5cm 3.cm,clip]{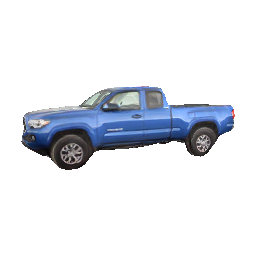}&
		\includegraphics[width = .12\textwidth, trim=0.9cm 1.5cm 0.9cm 1.5cm,clip]{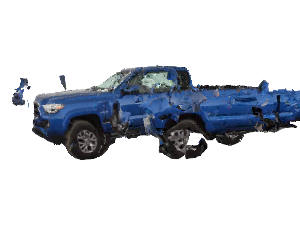}&
        \includegraphics[width = .12\textwidth, trim=0.9cm 1.5cm 0.9cm 1.5cm,clip]{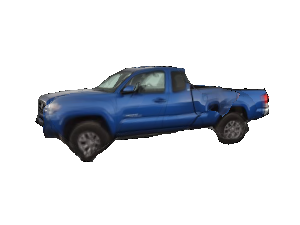}&
        \includegraphics[width = .12\textwidth, trim=0.9cm 1.5cm 0.9cm 1.5cm,clip]{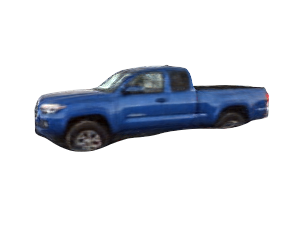}&
		\includegraphics[width = .12\textwidth, trim=0.9cm 1.5cm 0.9cm 1.5cm,clip]{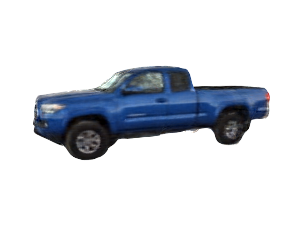} &
		\includegraphics[width = .12\textwidth, trim=0.9cm 1.5cm 0.9cm 1.5cm,clip]{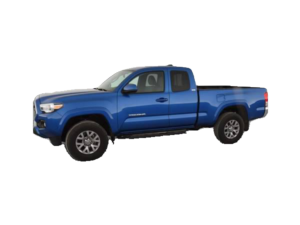} \\

		Input &RegNeRF & NeRS* & Munkberg et al. \cite{munkberg2021nvdiffrec} &DS & our w/BCG*  & our wo/BCG & GT
    \end{tabular}	
    \end{center}

	\caption{Qualitative results on the MVMC Car dataset. NeRS*: the original NeRS implementation crops the images before training and thus changes their aspect ratio during training. Thus, the rendered images have an aspect ratio of 1, while the original ones do not. 
    Note that in our w/BCG* column we remove the background from our w/BCG results for better visualization.  
	}
	\label{fig:qualitative_results_mvmc_car}
    
\end{figure*}

%


\end{document}